\documentclass[12pt]{article}
\usepackage{amsfonts,amssymb}
\usepackage{hyperref}
\usepackage{titlesec}
\usepackage{titletoc}
\usepackage{amsmath}
\numberwithin{equation}{section}
\usepackage{listings}
\usepackage{verbatim}
\usepackage{graphicx}
\usepackage{geometry}
\usepackage{mathrsfs}
\usepackage{float}
\usepackage[mathlines]{lineno}
\usepackage{amsthm}
\usepackage{bbm}
\usepackage{bm}

\usepackage{makecell}

\newtheorem{assumption}{Assumption}
\usepackage[dvipsnames]{xcolor}
\newtheorem{theorem}{Theorem}
\newtheorem{pro}{Proposition}
\newtheorem{lem}{Lemma}
\newtheorem{definition}{Definition}

\newtheorem{rem}{Remark}
\DeclareMathOperator{\Tr}{Tr}

\DeclareMathOperator{\rank}{rank}
\DeclareMathOperator{\sign}{\mathrm{sign}}

\DeclareMathOperator{\ii}{\bm{\mathrm{i}}} 

\usepackage{cite}
\title{Error Bound of Empirical $\ell_2$ Risk Minimization for 
Noisy Standard and Generalized Phase Retrieval Problems}
\author{Junren Chen\thanks{Department of 
		Mathematics, The University of Hong Kong. E-mail: chenjr58@connect.hku.hk.
		Thanks for the support from Hong Kong Research Grant Council HKPF scholarship.
		}\and Michael K. Ng\thanks{Institute of Data Science and Department of 
		Mathematics, The University of Hong Kong. E-mail: mng@maths.hku.hk.
		Research supported in part by Hong Kong Research Grant Council GRF 12300519, 17201020, 17300021, and joint NSFC-RGC N-HKU76921.
		}}
\date{}
\begin{document}
	
	\maketitle

\begin{abstract}
    In a noisy (standard) phase retrieval (NPR) problem, we want to reconstruct $x_0\in \mathbb{K}^d$ ($\mathbb{K}=\mathbb{R}$ or $\mathbb{C}$)
    from the measurements $\{|\alpha_k^*x_0|^2+\eta_k\}_{k=1}^n$ where $\{\alpha_k\}_{k=1}^n \subset \mathbb{K}^d$ are known vectors and $\{\eta_k\}_{k=1}^n\subset \mathbb{R}$ represent noise. Similarly, in a noisy generalized phase retrieval (NGPR) problem, we encounter the same task but the measurements become $\{x_0^*A_kx_0+\eta_k\}_{k=1}^n$ where $\{A_k\}_{k=1}^n$ are symmetric if $\mathbb{K}=\mathbb{R}$, or Hermitian if $\mathbb{K}=\mathbb{C}$. Many algorithms for the above two problems are based on empirical $\ell_2$ risk minimization (ERM), while the estimation performance of ERM in a noisy setting remains unclear, especially in the complex case. The main aim of this paper is to close the gap and establish new error bounds under different noise patterns, and our proofs are valid for both $\mathbb{K}=\mathbb{R}$ and $\mathbb{K}=\mathbb{C}$.
In NPR under arbitrary noise vector $\eta$, we derive a new error bound $O\big(\|\eta\|_\infty\sqrt{\frac{d}{n}} + \frac{|\mathbf{1}^\top\eta|}{n}\big)$, which is tighter than the currently known one $O\big(\frac{\|\eta\|}{\sqrt{n}}\big)$ in many cases. 
In NGPR, we show $O\big(\|\eta\|\frac{\sqrt{d}}{n}\big)$ for arbitrary $\eta$. 
In both problems, the bounds for arbitrary noise immediately give rise to $\tilde{O}(\sqrt{\frac{d}{n}})$ for sub-Gaussian or sub-exponential random noise, with some conventional but inessential assumptions (e.g., independent or zero-mean condition) removed or weakened. 
In addition, we make a first attempt to ERM under heavy-tailed random noise assumed to have bounded $l$-th moment. To achieve a trade-off between bias and variance, we truncate the responses and propose a corresponding robust ERM estimator, which is shown to possess the guarantee $\tilde{O}\big(\big[\sqrt{\frac{d}{n}}\big]^{1-1/l}\big)$ in both NPR, NGPR. All the error bounds straightforwardly extend to the more general problems of rank-$r$ matrix recovery, and these results deliver a conclusion that the full-rank frame $\{A_k\}_{k=1}^n$ in NGPR is more robust to biased noise than the rank-1 frame $\{\alpha_k\alpha_k^*\}_{k=1}^n$ in NPR. Extensive experimental results are presented to illustrate our theoretical findings.
\end{abstract}


\vspace{3mm}

\section{Introduction}
\label{section1}

Let $\mathbb{K} = \mathbb{R}$ or $\mathbb{C}$. Given a set of known measurement vectors $\{\alpha_1,...,\alpha_n\}\subset \mathbb{K}^d$ and the desired signal $x_0 \in \mathbb{K}^d$, the problem of recovering or estimating $x_0$ from the phaseless measurements $\{|\alpha_k^*x_0|^2: k\in [n]\}$ is known as (standard) phase retrieval\footnote{Since $\alpha_k$ is not restricted to Fourier measurement, this is termed as generalized phase retrieval in \cite{candes2015phase,sun2018geometric}. While in this work we adopt the terms used in \cite{wang2019generalized,huang2021almost}, and generalized phase retrieval refers to another model.}. This problem arises in a bunch of real world applications such as X-ray crystallography \cite{drenth2007principles}, speech recognition \cite{rabiner1993fundamentals}, nanostructure \cite{juhas2006ab}, coherent
diffractive imaging \cite{miao1999extending}, where it is expensive or even impossible for the sensors to capture the phase information, and the phase is thus missing. A theoretical line of works \cite{balan2006signal,bandeira2014saving,conca2015algebraic} affirmatively confirm that phase retrieval over $\mathbb{K}^d$ is possible. These works explore the required measurement number $n$ and show $n = 2d-1$ for $\mathbb{K}=\mathbb{R}$, or $n = 4d-4$ for $\mathbb{K}=\mathbb{C}$ is sufficient.

Let $\mathcal{H}_d(\mathbb{R})$ and $\mathcal{H}_d(\mathbb{C})$ be the set of real symmetric matrices, or the set of Hermitian matrices, respectively. Given known measurement matrices $\{A_1,...,A_n\}\subset \mathcal{H}_d(\mathbb{K})$ and an unknown signal $x_0\in\mathbb{K}^d$, generalized phase retrieval refers to the problem of recovering $x_0$ from the quadratic measurements $\{x_0^*A_kx_0:k\in [n]\}$. This model nicely encompasses several phaseless reconstruction problem, e.g., (standard) phase retrieval if $A_k = \alpha_k\alpha_k^*$ \cite{balan2006signal}, fusion frame phase retrieval if $A_k$ is orthogonal projection matrix \cite{edidin2017projections}. Generalized phase retrieval was first proposed and studied by Wang and Xu \cite{wang2019generalized} from the perspective of mimimal measurement number.

Besides the missing phase, a real-world problem should be studied as a noisy setting where measurements are corrupted by noise, for instance, the inevitable inexactness induced by the sensor, any deterministic or random noise with or without specific structure that may arise under different bad circumstances. 
Thus, let $\eta_k$ be the noise of the $k$-th measurement, the problems of noisy (standard) phase retrieval (NPR) and noisy generalized phase retrieval (NGPR) can be introduced as follows:
\begin{equation}
    \begin{cases}
    \label{n1.1}
    ~\mathrm{NPR}~: y_k = |\alpha_k^*x_0|^2 +\eta_k,~k\in [n], \\
    \mathrm{NGPR} : y_k = x_0^*A_kx_0 + \eta_k,~k\in [n].
    \end{cases}
\end{equation}
In such noisy setting, arguably the most natural way to estimate $x_0$ is via an empirical $\ell_2$ risk minimization (ERM), e.g., many algorithms   are developed with the goal of minimizing $\ell_2$ loss, as will be reviewed shortly. More precisely, ERM for NPR is
\begin{equation}
    \label{n1.2}
    \widehat{x} = \mathop{\arg\min}_{x \in \mathbb{K}^d} ~\sum_{k=1}^n \big(y_k - |\alpha_k^*x|^2\big)^2,
\end{equation}
and in NGPR the ERM estimator is given by 
\begin{equation}
    \label{n1.3}
    \widehat{x} = \mathop{\arg\min}_{x\in \mathbb{K}^d} ~\sum_{k=1}^n \big(y_k - x^*A_kx\big)^2.
\end{equation}

With an enlightening formulation, let us further generalize  NPR, NGPR to a more general rank-$r$ matrix recovery problem. Given $A,B \in \mathbb{K}^{d\times d}$, we define the  inner product $\big<A,B\big>=\mathrm{Trace}(A^*B)$ where $A^*$ is the conjugate transpose of $A$. Since $A_k$ in NGPR, $\alpha_k\alpha_k^*$ in NPR are Hermitian, these two models can be reformulated as follows:
\begin{equation}
\begin{cases}
    \nonumber
    ~\text{NPR~:}~y_k = \big<\alpha_k\alpha_k^*,x_0x_0^*\big>+\eta_k,~~k\in [n],\\
    \text{NGPR:}~y_k = \big<A_k,x_0x_0^*\big>+\eta_k,~~k\in [n].
\end{cases}
\end{equation}
Now one can see that NPR, NGPR are rank-1 positive semi-definite (PSD) matrix recovery in essence, and evidently the only difference lies in the measurement matrices. From this viewpoint, we further generalize NPR and NGPR to a rank-$r$ ($r$ is known), PSD matrix recovery problem:
\begin{equation}
\label{1.5}
    \begin{cases}
    y_k = \big<\alpha_k\alpha_k^*,U_0U_0^*\big>+\eta_k,~~k\in [n],\\
    y_k = \big<A_k ,U_0U_0^*\big>+\eta_k,~~k\in [n],
    \end{cases}
\end{equation}
where $U_0 \in \mathbb{K}^{d \times r}$ and $X_0=U_0U_0^*$ is the desired low-rank matrix. Similarly, $X_0=U_0U_0^*$ in (\ref{1.5}) can be estimated via ERM, i.e., minimizing the $\ell_2$ loss
\begin{equation}
    \label{1.6}
    \begin{cases}
       \displaystyle\widehat{U} = \mathop{\arg\min}_{U \in \mathbb{K}^{d\times r}}  
       { \sum_{k=1}^n} \big(y_k - \big<\alpha_k\alpha_k^*, UU^* \big>\big)^2,\\
      \displaystyle \widehat{U} = \mathop{\arg\min}_{U \in \mathbb{K}^{d\times r}} 
       { \sum_{k=1}^n} \big(y_k - \big<A_k , UU^* \big>\big)^2.
    \end{cases}
\end{equation}

Before we give a detailed review of related works, we briefly depict a big picture of the literature to justify our motivation. On one hand, probably starting from Wirtinger flow \cite{candes2015phase}, many guaranteed algorithms based on the non-convex formulation of ERM have been developed for either phase retrieval or low-rank matrix recovery, see  \cite{zhao2015nonconvex,zheng2015convergent,tu2016low,bhojanapalli2016global,wang2017unified,kolte2016phase}. On the other hand, only a few papers \cite{soltanolkotabi2014algorithms,huang2020performance,eldar2014phase,fan2022oracle}
discuss the estimation performance of ERM, and the theoretical bounds for reconstructed error are quite rare. Furthermore, existing error bounds are either highly sub-optimal and not tight for some noise patterns \cite{soltanolkotabi2014algorithms,huang2020performance}, or too restrictive 
(for instance, some are for the real case $\mathbb{K}=\mathbb{R}$ only \cite{eldar2014phase,lecue2015minimax}, some are obtained under a very specific noise (e.g., i.i.d., zero-mean, sub-Gaussian noise) and measurement type (e.g., Gaussian or complex Gaussian) \cite{fan2022oracle,soltanolkotabi2014algorithms,huang2020performance}). Due to the incompleteness of existing results, in general it remains unclear how well we can recover the underlying signal by ERM in a noisy setting, especially in the complex 
case $\mathbb{K}=\mathbb{C}$. 

The main aim of this paper is to close this gap by establishing new and tighter error bounds that are valid for both real and complex cases. We assume   
$\alpha_k$ in NPR or $A_k$ in NGPR has sub-Gaussian, zero-mean entries, and consider several different noise patterns. In the regime of heavy-tailed noise, we propose a robust ERM estimator with recovery guarantee. Moreover, our theoretical results straightforwardly extend to ERM in the more general problems of rank-$r$ PSD matrix recovery (\ref{1.5}).

\subsection{Related Works and Our Contributions}

The most related works that study exactly the same problem should be \cite{lecue2015minimax,eldar2014phase,huang2020performance,fan2022oracle}, while other related works include ERM-based algorithms in phase retrieval or low-rank matrix recovery, or recent developments of generalized phase retrieval after the first study \cite{wang2019generalized}. In this part, we review these related works and state new results obtained in this paper to show the improvement or complement over existing ones. 

\subsubsection{
Standard Phase Retrieval}
Let $\eta = (\eta_1,...,\eta_n)^{\top}\in\mathbb{R}^n$ be the noise vector. Unlike the algebraic arguments used in \cite{balan2006signal,bandeira2014saving,conca2015algebraic} that do not give any stability guarantee, by regarding $x_0\in \mathbb{K}^d$ as $x_0x_0^*\in \mathbb{K}^{d\times d}$ to linearize the problem (known as "lifting"), Phaselift is a tractable convex relaxation with error bounds $O\big(\|\eta\|\big)$ \cite{candes2013Phaselift} and $O\big(\frac{\|\eta\|_1}{n}\big)$ \cite{candes2014solving} in a noisy setting. However, due to essentially larger variables, the memory and computational requirements of Phaselift are prohibitive in large-scale problems.

To avoid the lifting, Wirtinger flow that directly minimizes $\ell_2$ loss in the original vector field was proposed in \cite{candes2015phase}, also see some subsequent developments \cite{chen2017solving,kolte2016phase}. Soltanolkotabi showed Wirtinger flow could still find a global minimizer of $\ell_2$ loss under moderate noise \cite{soltanolkotabi2014algorithms}. Thus, he naturally considered the reconstruction error of ERM and established the error bound $O\big(\frac{\|\eta\|}{\sqrt{n}}\big)$ for arbitrary noise in both real and complex case, see Theorem 12.3.1 in \cite{soltanolkotabi2014algorithms}. More recently, Huang et al. revisited this problem in the preprint \cite{huang2020performance} but obtained the same error bound $O\big(\frac{\|\eta\|}{\sqrt{n}}\big)$ without improvement. We point out that the result in \cite{soltanolkotabi2014algorithms,huang2020performance} is restricted to complex Gaussian measurement. Two other works\footnote{The results in \cite{lecue2015minimax,eldar2014phase} are applicable to recovery of $x \in \mathcal{T}$ for some constraint set $\mathcal{T}$. Here, we only show the result of $\mathcal{T} = \mathbb{R}^d$, which is the focus of our work.} \cite{eldar2014phase,lecue2015minimax} aimed to understand stability of phase retrieval by studying ERM estimator under independent sub-Gaussian noise with zero mean. Specifically, in \cite{eldar2014phase} Eldar et al. proved $O\big( \sqrt{\frac{d(\log d)^2}{n}} + \frac{\sqrt{\log d}}{n^{1/4}}\big)$ for empirical $\ell_q$ risk minimization where $q\in (1,2]$ should be properly specified by other parameters. Considering the specific $\ell_2$ loss minimization (i.e., ERM in this work), Lecu{\'e} et al. \cite{lecue2015minimax} showed the near minimax rate $O\big(\sqrt{\frac{d \log n}{n}}\big)$. However, \cite{eldar2014phase,lecue2015minimax} are for the real case only, and one may see their arguments heavily rely on $\mathbb{K}=\mathbb{R}$, hence do not apply to $\mathbb{K}=\mathbb{C}$. For instance, the   relation $|\alpha_k^*x|^2- |\alpha_k^*x_0| = \big<\alpha_k,x-x_0\big>\big<x+x_0,\alpha_k\big>$ is frequently used in their proofs but fails in the complex case. In addition, some works discuss ERM performance in a different amplitude-based noisy model $y_k = |\alpha_k^* x_0|+\eta_k$, see \cite{huang2020estimation} for instance.

To see existing results are highly incomplete, let us consider the complex NPR where the noise $\eta_k$ are i.i.d. drawn from standard Gaussian distribution. On one hand, many algorithms have been developed based on $\ell_2$ loss minimization, e.g., Wirtinger flow and its variants. But on the other hand, even the algorithms ideally find a global minimizer, since $O\big(\frac{\|\eta\|}{\sqrt{n}}\big) $ provided in \cite{soltanolkotabi2014algorithms,huang2020performance} reduces to $O(1)$, and $O\big({\sqrt{\frac{d\log n}{n}}}\big)$ in \cite{lecue2015minimax} does not apply to $\mathbb{K}=\mathbb{C}$, we do not know whether the obtained solution achieves sufficiently small error when $n\to \infty$ (statistically termed as consistency), and at what rate if yes. The main issue is that $O\big(\frac{\|\eta\|}{\sqrt{n}}\big)$, the only existing bound for complex NPR, is essentially a loose bound for sub-Gaussian random noise.

To close the gap, we establish a new error bound $O\big(\|\eta\|_\infty\sqrt{\frac{d}{n}}+\frac{|\mathbf{1}^{\top}\eta|}{n}\big)$ for arbitrary noise. In some noise patterns, this new bound characterizes ERM performance much more precise than the already known one $O\big(\frac{\|\eta\|}{\sqrt{n}}\big)$ (e.g., see Figure \ref{nfig3} to see in advance). For instance, it directly gives rise to $O\big(\sqrt{\frac{d\log n }{n}}\big)$ for zero-mean, sub-Gaussian $\eta$, thus finding the complex counterpart of \cite{eldar2014phase,lecue2015minimax} and affirmatively answering the question in the last paragraph. Notably, even the independent condition required in \cite{eldar2014phase,lecue2015minimax} is removed.  

\subsubsection{
Generalized Phase Retrieval}
Wang and Xu introduced the generalized phase retrieval problem and investigated the minimal measurement number \cite{wang2019generalized}. Under a general class of $A_k$, they showed that $n = 2d-1$ for $\mathbb{K}=\mathbb{R}$ and $n = 4d-4$ for $\mathbb{K}=\mathbb{C}$ suffice to reconstruct all $x \in \mathbb{K}^d$, thereby extending known results for (standard) phase retrieval \cite{balan2006signal,bandeira2014saving,conca2015algebraic}. The measurement number can be further lowered if we only pursue  almost everywhere phase retrieval  \cite{huang2021almost}. Algorithms and optimization have been considered in subsequent works. In \cite{huang2020solving}, Huang et al. proposed to use Wirtinger flow to solve a complex random quadratic system $y_k = x^*A_kx,~k\in [n]$ via minimizing $\ell_2$ loss. Thaker et al. showed the non-convex $\ell_2$ loss enjoys a benign optimization landscape that enables gradient algorithm with arbitrary initialization \cite{thaker2020sample}.

While \cite{huang2020solving,thaker2020sample} are restricted to noiseless regime, a recent manuscript \cite{fan2022oracle} considered noisy setting and proved the error bound $\| \widehat{x}\widehat{x}^{\top}-x_0x_0\|_F = O\big(\sqrt{\frac{d \log n}{n}}\big)$\footnote{The bound in their work is for $d^{'}(\widehat{x},x_0) = \min_{\theta}\|\widehat{x}-\exp(\ii\theta)x_0\| $. For the reason stated at the beginning of Section \ref{section3}, here we transfer it to a bound for $\|\widehat{x}\widehat{x}^*-x_0x_0^*\|_F$. Unless otherwise specified, the error bounds in this work are for $\|\widehat{x}\widehat{x}^*-x_0x_0^*\|_F$.} for ERM, but there are several issues: The proof is for the real case only, and their proof cannot handle the complex case\footnote{The authors of \cite{fan2022oracle} mentioned that their proof could be applied to $\mathbb{K}=\mathbb{C}$ by reducing complex matrix to the real one, but the claim does not make sense since the obtained real measurement matrices never satisfy Assumption 2.1 in their work.
}; The bound itself contains inessential logarithmic factor; The noise is restricted to be i.i.d., zero-mean, sub-Gaussian noise.

In this work, we do not directly deal with sub-Gaussian noise. Instead, we first show the bound $O\big(\|\eta\|\frac{\sqrt{d}}{n}\big)$
that is valid for both $\mathbb{R}$ and $\mathbb{C}$ under arbitrary noise. This bound immediately translates to the guarantee $O\big(\sqrt{\frac{d}{n}}\big)$ for independent, sub-Gaussian noise (Thus removing $\log n$ in \cite{fan2022oracle}), or $O\big(\sqrt{\frac{d\log n}{n}}\big)$ for sub-Gaussian noise without independent assumption. Note that zero-mean assumption is completely removed.

\subsubsection{
Heavy-tailed Noise Regime
} Heavy-tailed behaviour is ubiquitous in many real data including economic data \cite{ibragimov2015heavy}, biomedical data \cite{woolson2011statistical}, noise in signal processing \cite{fan2021shrinkage,kruczek2020detect} and machine learning \cite{wei2019statistical}. Statistically, it is widely adopted formulation to capture heavy-tailed random variable via a bounded $l$-th moment condition. Compared with sub-Gaussian assumption, heavy-tailed data contain outliers that bring challenges to the robustness. \textcolor{black}{To address the issue, many techniques in statistical estimation have been developed}, such as median of means \cite{hsu2016loss}, robust empirical risk \cite{catoni2012challenging}. Based on the intuition of making outliers less influential, the recent paper \cite{fan2021shrinkage} applied a simple shrinkage (or truncation) step to high-dimensional regression problems with heavy-tailed noise and established (near) minimax rate, also see some subsequent works \cite{zhu2021taming,chen2022high,wang2021robust}.

We also study NPR, NGPR under heavy-tailed random noise drawn from a distribution with bounded $l$-th moment. Inspired by previous success of truncation in high-dimensional regression via convex programs \cite{fan2021shrinkage}, we propose to truncate $y_k$ to $\widetilde{y}_k= \mathrm{sign}(y_k)\min\{|y_k|,\tau\}$ with threshold $\tau>0$, and then recover $x_0$ by minimizing the robust $\ell_2$ loss constructed by the truncated responses. With suitable $\tau$, theoretically we prove the proposed robust ERM estimators achieve $\tilde{O}\big(\big[\sqrt{\frac{d}{n}} \big]^{1-1/l}\big)$ in both NPR, NGPR. Thus, the sample complexity (for $o(1)$ error) is near optimal, but the rate is slower compared with $\tilde{O}\big(\sqrt{\frac{d}{n}}\big)$ for sub-Gaussian noise. To the best of our knowledge, this is the first study of phase retrieval problem under heavy-tailed noise with a bounded moment assumption. And we leave it an open problem whether faster rate can be achieved by sharper analysis or different techniques. 

\subsubsection{
Low Rank Matrix Recovery
} NPR, NGPR could be naturally generalized to be a rank-$r$ PSD matrix sensing problem (\ref{1.5}), with sensing matrix being a general Hermitian matrix $A_k$, or a specific rank-1 PSD matrix $\alpha_k\alpha_k^*$, respectively.

When entries of $A_k \in \mathbb{K}^{d\times d}$ are drawn from sub-Gaussian distribution, many algorithms based on $\ell_2$ loss minimization has been developed \cite{zhao2015nonconvex,chen2015fast,zheng2015convergent,tu2016low,bhojanapalli2016global,wang2017unified}. We emphasize that these works and ours are of radically different purposes. In particular, the main aim of these works is to propose guaranteed non-convex algorithms that is cheaper and more scalable than convex relaxation. Thus, most of them do not consider the presence of noise \cite{zhao2015nonconvex,tu2016low,zheng2015convergent}, or at most i.i.d., sub-Gaussian noise by statistical convention \cite{bhojanapalli2016global,wang2017unified}. By contrast, this work focuses on the theoretical aspect and answers a more fundamental problem, i.e., how well ERM recovers $x_0$ under different noise patterns. Moreover, all these papers are for real case only, and rely on quite stringent RIP condition which greatly restricts the random measurement ensemble one may use. By contrast, our theoretical error bounds are for both real and complex cases, and also RIP condition is in a relaxed version that allows random sub-Gaussian matrix with correlated entries (see Assumption \ref{assumption2}). \textcolor{black}{When it comes to rank-1 measurement matrix $A_k = \alpha_k\alpha_k^*$}, non-convex algorithms aiming to minimize $\ell_2$ loss have been studied in \cite{sanghavi2017local,li2019nonconvex}, but restricted to real and noiseless case.

In subsection \ref{rank-r}, all the obtained results for NPR, NGPR are extended to the ERM performance in low-rank matrix sensing (recovery) problems. It is not possible to precisely compare our results with these more algorithmic works, but we believe this work justifies the algorithms designed for minimizing $\ell_2$ loss in a noisy setting.

\subsubsection{
Rank-1 Frame v.s. Full-Rank Frame
} 

Note that NPR and NGPR only differ in the measurement frame. As widely adopted by related works (\textcolor{black}{e.g.}, \cite{candes2015phase,soltanolkotabi2014algorithms,huang2020performance,fan2022oracle,candes2011tight}), entries of $\alpha_k$ in NPR, or $A_k$ in NGPR are assumed to be zero-mean and sub-Gaussian in this work, and $\{\alpha_k\alpha_k^*\}_{k=1}^n$, $\{A_k\}_{k=1}^n$ are respectively called "rank-1 frame" or "full-rank frame" for convenience. An obvious advantage of rank-1 frame are essentially lower storage and computational costs. On the other hand, our results reveal an advantage of full-rank frame, that is, full-rank frame is more robust to biased noise than the rank-1 frame. We also present numerical results to confirm the conclusion (One shall see Figure \ref{nfig6} to see in advance). 
\subsection{Summary}

In the last subsection, we have mentioned 
our new error bounds. The related works \cite{huang2020performance,lecue2015minimax,fan2022oracle,soltanolkotabi2014algorithms} contain several comparable results, so we summarize these known results and our obtained error bounds in Table \ref{table2} for NPR,
and Table \ref{table1} for NGPR\footnote{In the real case of rank-$r$ matrix sensing (or regression), \cite{wang2017unified,chen2015fast} show the proposed gradient descent algorithms linearly shrink the error to $O\big(\sqrt{\frac{rd}{n}}\big)$ for i.i.d. sub-Gaussian noise. These results are of slightly different nature and require more stringent RIP condition, so we do not include them in table \ref{table1}.} to ease comparison.

\begin{table}[ht]
	\centering
\begin{tabular}{c|c |c| c |c } 
  \hline
 ~& $\mathbb{K}$ & Measurement $A_k$ & Noise $\eta_k$ & Error Bound \\ 
\hline
 \cite{lecue2015minimax} & $\mathbb{R}$ & {\footnotesize sub-Gaussian} & \makecell{{\footnotesize i.i.d. zero-mean} \\ {\footnotesize sub-Gaussian}} & $O\big(\sqrt{\frac{d\log n}{n}}\big)$\\ 
 \hline
 \cite{soltanolkotabi2014algorithms,huang2020performance} &  $\mathbb{R},\mathbb{C}$& {\footnotesize complex Gaussian} & {\footnotesize arbitrary noise}& $O\big(\frac{\|\eta\|}{\sqrt{n}}\big)$  
 \\
 \hline
 {\footnotesize this paper} & $\mathbb{R},\mathbb{C}$&  {\footnotesize sub-Gaussian} & {\footnotesize arbitrary noise} &$O\big(\frac{\|\eta\|}{\sqrt{n}}\big)$ \\ 
 \hline
 {\footnotesize this paper} & $\mathbb{R},\mathbb{C}$ &  {\footnotesize sub-Gaussian} & {\footnotesize arbitrary noise} & $O\big(\|\eta\|_\infty\sqrt{\frac{d}{n}}+\frac{|\mathbf{1}^{\top}\eta|}{n}\big)$\\ 
 \hline 
 {\footnotesize this paper} & $\mathbb{R},\mathbb{C}$ & {\footnotesize sub-Gaussian} & \makecell{ {\footnotesize   zero-mean} \\ {\footnotesize sub-Gaussian $\eta$}}& $O\big(\sqrt{\frac{d \log n}{n}}\big)$ \\
 \hline 
 {\footnotesize this paper} & $\mathbb{R},\mathbb{C}$ & {\footnotesize sub-Gaussian}   &
 \makecell{ {\footnotesize i.i.d. zero-mean} \\ {\footnotesize sub-exponential $\eta_k$}}& $O\big(\sqrt{\frac{d (\log n)^2}{n}}\big)$
 \\
 \hline 
 {\footnotesize this paper} & $\mathbb{R},\mathbb{C}$ & {\footnotesize sub-Gaussian} &  
 \makecell{ {\footnotesize i.i.d. symmetric} \\ {\footnotesize heavy-tailed}} &$O\big(\big[\sqrt{\frac{d}{n}}\big]^{1-\frac{1}{l}}(\log n)^2\big)$
 \\
 \hline
\end{tabular}
\caption{Error Bounds for ERM in NPR}
\label{table2}
\end{table}

	\begin{table}[ht]
	\centering
\begin{tabular}{c|c|c|c|c } 
 \hline
 ~& $\mathbb{K}$ & Measurement $A_k$ & Noise $\eta_k$ & Error Bound \\ 
 \hline
 \cite{fan2022oracle} & $\mathbb{R}$ &  {\footnotesize sub-Gaussian} & \makecell{ {\footnotesize i.i.d. zero-mean} \\ {\footnotesize sub-Gaussian}}   &$O\big(\sqrt{\frac{d\log n }{n}}\big) $\\ 
 \hline
 {\footnotesize this paper} & $\mathbb{R},\mathbb{C}$ & {\footnotesize sub-Gaussian} & {\footnotesize arbitrary noise} &  $O\big(\|\eta\|\frac{\sqrt{d}}{n}\big)$\\
 \hline
 {\footnotesize this paper} & $\mathbb{R},\mathbb{C}$ & {\footnotesize sub-Gaussian } &  \makecell{ {\footnotesize sub-Gaussian $\eta-\mathbbm{E}\eta$} \\ {\footnotesize bounded $\|\mathbbm{E}\eta\|_\infty$}} & $O\big(\sqrt{\frac{d}{n}}\big)$ \\ 
 \hline
  {\footnotesize this paper} & $\mathbb{R},\mathbb{C}$ & {\footnotesize sub-Gaussian } & {\footnotesize sub-Gaussian $\eta_k$} & $O\big(\sqrt{\frac{d\log n}{n}}\big)$ \\ 
 \hline
 {\footnotesize this paper} & $\mathbb{R},\mathbb{C}$ & {\footnotesize sub-Gaussian} & {\footnotesize sub-exponential $\eta_k$} & $O\big(\sqrt{\frac{d(\log n)^2}{n}}\big)$
  \\ 
 \hline
 {\footnotesize this paper} & $\mathbb{R},\mathbb{C}$ & {\footnotesize sub-Gaussian} &{\footnotesize i.i.d. heavy-tailed}  &$O\big(\big[\sqrt{\frac{d}{n}}\big]^{1-\frac{1}{l}}\log n\big)$ 
 \\
\hline
\end{tabular}
\caption{Error Bounds for ERM in NGPR}
\label{table1}
\end{table}



\subsubsection{Outline}

	The rest of the paper is organized as follows. In Section \ref{section2}, we introduce notations and necessary preliminaries. In Section \ref{section3}, we first show the obtained error bounds for NPR and NGPR with different noise patterns in $\mathbb{K}=\mathbb{C}$ (but the proofs apply to $\mathbb{K}=\mathbb{R}$),
and then provide the extension to low-rank matrix recovery. 
In Section \ref{section4}, we report experimental results to corroborate the theoretical findings. Some concluding remarks are given in Section \ref{section5} to end this paper.  
	
	\section{Notations and Preliminaries}
	\label{section2}
	\subsection{Notations}
	For $v\in \mathbb{C}^d$, we let $\|v\| = (\sum_i |v_i|^2)^{1/2}$ be the $\ell_2$ norm, $\|v\|_{\infty} = \max |v_i|$ be the max norm, $\|v\|_1 = \sum_{i}|v_i|$ be the $\ell_1$ norm. For $A\in \mathbb{C}^{d\times d}$ with singular values $\sigma_1,...,\sigma_d$, the Frobenius norm, operator norm, and nuclear norm are given by $\|A\|_F = (\sum_{i,j}|a_{i,j}|^2)^{1/2}$, $\|A\| = \max_i \sigma_i$, $\|A\|_{nu} = \sum_i \sigma_i$, respectively. Let $\big<.,.\big>$ be the standard inner product in $\mathbb{C}^d$ or $\mathbb{C}^{d\times d}$, that is, $\big<x,y\big> = x^*y$ for $x,y\in \mathbb{C}^d$, or $\big<A,B\big>= \Tr(A^*B)$ for $A,B\in \mathbb{C}^{d\times d}$, where $.^* = \bar{.}^{\top}$ is the conjugate transport. A matrix $B = [b_1,...,b_d] \in \mathbb{C}^{d\times d}$ can be vectorized by $\mathrm{vec}(.)$, i.e., $\mathrm{vec}(B) = [b_1^{\top},...,b_d^{\top}]^{\top}\in \mathbb{C}^{d^2}$. For a real random variable $X$, we define the $L^p$ norm as $\|X\|_{L_p} = \big(\mathbbm{E}|X|^p\big)^{1/p}$ for $p\geq 1$. The sub-Gaussian norm $\|X\|_{\psi_2}$ and sub-exponential norm $\|X\|_{\psi_1}$ will be defined later.

	We use $\mathcal{H}_d$ to denote the set of Hermitian matrices in $\mathbb{C}^{d\times d}$, note that we simplify the previous notation $\mathcal{H}_d(\mathbb{C})$ as we focus on the complex case in the following. Since we frequently work with the set of low-rank Hermitian matrices, we further define $\mathcal{H}_{d,r} = \mathcal{H}_d \cap \{A \in \mathbb{C}^{d\times d}:rank(A)\leq r\}$, and the set of those with normalized Frobenius norm $\mathcal{H}_{d,r}^s = \{A\in\mathcal{H}_{d,r}:\|A\|_F=1\}$.

 We use $a+b\ii$ to denote complex numbers, while $i$ still serves for other purposes. For $a \in \mathbb{C}$, we use $a^\mathcal{R}= \mathrm{Re}(a)$, $a^\mathcal{I} = \mathrm{Im}(a)$ to represent the real part, complex part, respectively, and $\mathrm{Re}(.),~.^\mathcal{R},~\mathrm{Im}(.),~.^\mathcal{I}$ operate on complex vectors or matrices element-wisely. The phase of a non-zero $a\in\mathbb{C}$ is $\sign(a) = a/|a|$, and we let $\sign(0)=1.$

    We use $\mathbbm{P}(.),~\mathbbm{E}(.)$ to denote the probability or expectation, respectively. For an event $\mathscr{X}$, let $\mathbbm{1}(\mathscr{X})$ be its indicator function, i.e., $\mathbbm{1}(\mathscr{X}) = 1$ if it happens, $0$ otherwise. We denote absolute constants by $C_i, i = 1,2,...$, whose values may vary from line to line. 
    
    Standard complexity notations are adopted. For two terms $T_1$, $T_2$, the fact that $T_1 \leq C_1 T_2$ for some $C_1 >0$ is expressed as $T_1 = O(T_2)$. If $T_1\geq C_2T_2$ for some $C_2>0$, we write $T_1 = \Omega(T_2)$. It is denoted by $T_1 = \Theta(T_2)$ if both $T_1 = O(T_2)$ and $T_1 = \Omega(T_2)$ hold. Moreover, $\tilde{O}(.)$, $\tilde{\Omega}(.)$, $\tilde{\Theta}(.)$ are of similar meaning but ignore logarithmic factors, e.g., we can write $A = \tilde{O}(n)$ if $A = O\big(n(\log n )^2\big)$.

    We let $[k] = \{1,2,...,k\}$. For a set $\mathcal{K}$ we define $|\mathcal{K}|$ to be its size. The all-ones vector $(1,1,...,1)^{\top}$ with self-evident dimension is denoted by $\mathbf{1}$. The simple fact \begin{equation}
	    \label{nunormbound}
	    \sup_{X\in \mathcal{H}^s_{d,r}}\|X\|_{nu} \leq \sup_{X\in \mathcal{H}^s_{d,r}} \sqrt{\rank(X)}\|X\|_F \leq \sqrt{r}
	\end{equation}
    	will be repeatedly invoked in our proof. 
    \subsection{Sub-Gaussian and sub-exponential random variable}
     For random variable $X$, the sub-Gaussian norm $\|X\|_{\psi_2}$ and sub-exponential norm $\|X\|_{\psi_1}$ are defined to be 
    \begin{equation}
        \|X\|_{\psi_2} = \inf \Big\{t>0:\mathbbm{E} \exp \big(\frac{X^2}{t^2}\big) \leq 2\Big\}~~,~~\|X\|_{\psi_1} = \inf \Big\{t>0:\mathbbm{E} \exp \big(\frac{|X|}{t}\big) \leq 2\Big\},
        \label{2.2}
    \end{equation}
    Note that the two norms have the following relation (e.g., Lemma 2.7.7 in \cite{vershynin2018high})
     \begin{equation}
         \label{2.6}
         \|X_1X_2\|_{\psi_1} \leq \|X\|_{\psi_2}\|Y\|_{\psi_2}.
     \end{equation}
    Let us define sub-Gaussian and sub-exponential random variable (or vector) without requiring zero mean. This deviates from some textbooks but is more suitable for presenting our results that allow noise bias.
    
    \begin{definition}
    \label{def1}
    A real random variable $X$ is sub-Gaussian if $\|X\|_{\psi_2}<\infty$. A $d$-dimensional real random vector $X $ is sub-Gaussian if $w^{\top}X$ is sup-Gaussian for any $w\in \mathbb{R}^d$, or equivalently, $\sup_{\|w\|=1}\| w^{\top}X\|_{\psi_2}<\infty.$ We further let $\|X\|_{\psi_2} =\sup_{\|w\|=1}\| w^{\top}X\|_{\psi_2} $.
    \end{definition}
     
     \begin{definition}
     A real random variable $X$ is sub-exponential if $\|X\|_{\psi_1}<\infty$.
     \end{definition}
    
    Sub-Gaussian random variable shares many similar properties with Gaussian distribution. We list some of them here, and one may check more in Proposition 2.5.2, 2.6.1 in \cite{vershynin2018high}.
    \begin{pro}
    \label{pro1}
    Assume random variable $X$ is sub-Gaussian, then:
    
    \vspace{0.5mm}
    
    \noindent{\rm (a)} For some $C_1>0$, we have $\mathbbm{P}\big(|X|\geq t\big)\leq 2\exp\big( -\frac{t^2}{C_1\|X\|_{\psi_2}^2}\big)$ for any $t>0$.
    
    
    \noindent{\rm (b)} For some $C_2>0$, we have $\big(\mathbbm{E}|X|^p\big)^{1/p}\leq C_2\|X\|_{\psi_2}\sqrt{p}$ for any positive integer $p$. 
    \end{pro}
    
    \begin{pro}
    \label{pro2}
    Assume zero-mean, independent random variables $X_1,...,X_N$ are sub-Gaussian, then for some $C_1>0$, $\| \sum_{i=1}^NX_i\|^2_{\psi_2}\leq C_1\sum_{i=1}^N \|X_i\|_{\psi_2}^2$.
    \end{pro}
    
    As Proposition \ref{pro1}, sub-exponential $X$ also possesses a slightly heavier exponentially-decaying tail and weaker moment constraint, see Proposition 2.7.1 in \cite{vershynin2018high} for instance.
    
    \begin{pro}
    \label{pro3}
    Assume random variable $X$ is sub-exponential, then:
    
    \vspace{0.5mm}
    
    \noindent{\rm (a)} For some $C_1>0$, we have $\mathbbm{P}\big(|X|\geq t\big) \leq 2\exp \big(-\frac{t}{C_1\|X\|_{\psi_1}}\big)$ for any $t>0$.
    
    \noindent{\rm (b)} For some $C_2>0$, we have $\big(\mathbbm{E}|X|^p\big)^{1/p}\leq C_2\|X\|_{\psi_1}p$ for any positive integer $p$. 
    \end{pro}

     We close this part by a standard estimation on maximum of sub-Gaussian or sub-exponential random variables.

     \begin{pro}
     \label{pro4} Let $X = (X_1,...,X_N)^{\top}$ be a random vector. 
     
     \vspace{0.5mm}

     \noindent
     {\rm (a)} If $\max_{k\in [N]} \|X_k\|_{\psi_2} \leq R_0$, then for some $C_1>0$ we have $\|X\|_\infty \leq C_1R_0\sqrt{\log N}$ with probability at least $1-2n^{-24}$.
     
     \vspace{0.5mm}
     
     \noindent
     {\rm (b)} If $\max_{k\in [N]}\|X_k\|_{\psi_1}\leq R_0$, then for some $C_2>0$ we have $\|X\|_\infty \leq C_2R_0\log N$ with probability at least $1-2n^{-9}$.
     \end{pro}
     \noindent{\it Proof.} (a) By (a) of Proposition \ref{pro1}, for any $1\leq k\leq N$ we have $\mathbbm{P}\big(|X_k|\geq t\big) \leq 2\exp\big(-\frac{t^2}{C_1R_0^2}\big)$. Thus, a union bound gives $\mathbbm{P}\big(\|X\|_\infty \geq t\big)\leq 2N\exp\big(-\frac{t^2}{C_1R_0^2}\big)$. Setting $t = 5R_0\sqrt{C_1\log N}$ concludes the proof. (b) We apply (a) of Proposition \ref{pro3} instead, followed by a union bound. The proof is similar and hence omitted.   \hfill $\square$

    Note that Proposition \ref{pro4} does not require $\{X_k:k\in[N]\}$ to be independent.

     In addition, we need Mendelson's small ball method to establish a lower bound for an empirical process, and also several technical Lemmas to support the proofs of our main results. For clarity, these materials are postponed to Appendix \ref{appendixa} and Appendix \ref{appendixb}.
     
     \section{Main Results}
	\label{section3}
	
Before presenting the main results, we first state the metric used to measure reconstruction error. Note that in the complex case of phase retrieval problems, the measurements never distinguish $\{\exp(\ii \theta)x_0:\theta\in \mathbb{R}\}$, so even exact reconstruction should be up to a global phase factor. Thus, in the noisy case, one reasonable and frequently adopted metric for error is $d^{'}(\widehat{x},x_0) = \min_{\theta}~\|\widehat{x}-\exp(\ii\theta)x_0\|=\|\widehat{x}-\hat{z}x_0 \|$ where $\hat{z} = \mathrm{sign}(x_0^*\widehat{x})$. Besides, we can also identify the set $\{zx:|z|=1,z\in \mathbb{C}\}$ with the rank-1 PSD matrix $xx^*$ (lifting), which suggests the metric $d(\widehat{x},x_0) = \|\widehat{x}\widehat{x}^* - x_0x_0^* \|_F$.

We point out that an error bound for $d(\widehat{x},x_0)$ always produces a corresponding one for $d^{'}(\widehat{x},x_0)$. To see this, we assume the bound for $d(\widehat{x},x_0)$ as $d(\widehat{x},x_0) \leq \mathscr{B}/\sqrt{2}$, then we can directly verify the relation $\|\widehat{x}-\hat{z}x_0\|\cdot\| \widehat{x}+\hat{z}x_0\| \leq \sqrt{2}d(\widehat{x},x_0) \leq \mathscr{B},$ which further gives
	\begin{equation}
	    \begin{aligned}
	         \|\widehat{x}- \hat{z} x_0\| \leq &\min \Big\{\frac{\mathscr{B}}{\max\{\|\widehat{x}\|,\|x_0\|\}},2\max\{\|\widehat{x}\|,\|x_0\|\}\Big\}
	         \leq &\min \Big\{\frac{\mathscr{B}}{\|x_0\|}, \sqrt{2\mathscr{B}}\Big\}.
	    \end{aligned}
	    \nonumber
	\end{equation}
	Here, in the first inequality we use two simple facts $\|\widehat{x}+\hat{z}x_0\| \geq \max\{\|\widehat{x}\|,\|x_0\|\}$, $\|\widehat{x}-e^{\ii \alpha_0}x_0\| \leq 2 \max\{\|\widehat{x}\|,\|x_0\|\}$, and in the second inequality we get rid of $\|\widehat{x}\|$ by comparing $\max\{\|\widehat{x}\|,\|x_0\|\}$ and $\sqrt{\mathscr{B}/2}$. Therefore, we only present error bounds with regard to $d(\widehat{x},x_0)$, which is more conducive for the technical proof.


	\subsection{NPR with Arbitrary Noise}


    The only known error bound for complex (i.e., $\mathbb{K}=\mathbb{C}$) NPR, to our best knowledge, is $O\big(\frac{\|\eta\|}{\sqrt{n}}\big)$ obtained in \cite{soltanolkotabi2014algorithms,huang2020performance}. On one hand, this error bound applies to arbitrary corruption, and has been shown to be optimal for some noise patterns (e.g., $\eta_k = \mathbf{1}$, $\eta_k\sim \mathcal{N}(1,1)$); On the other hand, this is in fact far from a tight bound for many other noise patterns (e.g., $\eta_k \sim \mathcal{N}(0,1)$). The main aim of this part is to show a new error bound for arbitrary corruption. It is sometimes essentially tighter than $O\big(\frac{\|\eta\|}{\sqrt{n}}\big)$.

	We start from the random measurement ensemble in NPR. Since $\alpha_k\in \mathbb{C}^d$ contains $2d$ real variables, we first draw random vector $\breve{\alpha}_0 = [\breve{a}_{0,i}] \in \mathbb{R}^{2d}$ satisfying Assumption \ref{assumption2}, and construct $\breve{\alpha} = [\breve{a}_{i}]$ by letting $\breve{a}_{i} =\breve{a}_{0,i} + \breve{a}_{0,i+d}\ii$. Then, $\{\alpha_1,...,\alpha_n\}$ are assumed to be independent copies of $\breve{\alpha}$. We use the notation  $D = [\alpha_1,\alpha_2,...,\alpha_n]^* \in \mathbb{C}^{n\times d}$ and its associated operator $\mathcal{D}(X) = (\big<\alpha_1\alpha_1^*,X\big>,...,\big<\alpha_n\alpha_n^*,X\big>)^{\top}$ for any $X\in \mathbb{C}^{d\times d}$.
	
	\begin{assumption}
	\label{assumption2}
	The $2d$-dimensional random vector $\breve{\alpha}_0$ has i.i.d. entries drawn from a zero-mean, sub-Gaussian distribution $\mathcal{M}$ satisfying $\|\mathcal{M}\|_{\psi_2} = O(1)$, $\mathbbm{E}(\mathcal{M}^4)>(\mathbbm{E}\mathcal{M}^2)^2$. We refer to the latter one as a fourth moment condition.
	\end{assumption}
	
	In a nutshell, for each $k$, $[\alpha_k^\mathcal{R},\alpha_k^\mathcal{I}]$ has i.i.d., zero-mean and sub-Gaussian coordinates. Probably the assumption can be slightly weakened to $\alpha_k \in \mathbb{C}^d$ with i.i.d., zero-mean and sub-Gaussian coordinates, see \cite{krahmer2020complex}.

	Let us now justify $\mathbbm{E}(\mathcal{M}^4)>(\mathbbm{E}\mathcal{M}^2)^2$ as a necessary condition. Evidently, this fourth moment condition fails if and only if $\mathbbm{P}(|\mathcal{M}| = v_1) =1$ for some non-negative $v_1$. In this situation, it is not hard to see the measurements never distinguish $e_1 = (1,0,...,0)^{\top}$ and $e_2=(0,1,...,0)^{\top}$ \cite{krahmer2017phase}. Since the results in this paper are for any $x_0\in \mathbb{C}^d$ and indicate $o(1)$ error under sufficiently large sample size, such fourth moment condition is needed to guarantee uniqueness of solution\footnote{Nevertheless, one can still anticipate stable recovery result of ERM by considering the set of not overly spiky $x_0$, see \cite{krahmer2017phase,krahmer2020complex}. We do not make this effort in this work.}.

	In the next Theorem, we show a property of the rank-1 frame which implies identifiability of two low-rank Hermitian matrices (specifically two rank-1 Hermitian matrices for NPR).

\begin{theorem}
    \label{theorem6}
    Under Assumption \ref{assumption2}, assume $\{A_k = \alpha_k\alpha_k^*:k\in [n]\}$ are drawn as we stated before. Then for a specific positive integer $1\leq r\leq d $, there exist $C_i$ such that when $n\geq C_1rd$, with probability at least $1-2\exp(-C_2n)$, we have $C_3{n} \leq \|\mathcal{D}(X)\|_1\leq C_4\sqrt{r}{n}$, $\forall X \in \mathcal{H}_{d,r}^s.$
\end{theorem}

\noindent{\it Proof.} We first show the lower bound of $\inf_{X\in \mathcal{H}_{d,r}^s}\|\mathcal{D}(X)\|_1$ via Mendelson's small ball method and the associated Lemmas introduced in Appendix \ref{appendixa}. Lemma \ref{lemma1} yields that for any $\xi,t>0$, with probability at least $1-\exp(-\frac{t^2}{2})$, we have 
\begin{equation}
    \label{3.24}
    \inf_{X\in \mathcal{H}_{d,r}^s} \frac{1}{\sqrt{n}}\|\mathcal{D}(X)\|_1 \geq \xi \sqrt{n} Q_{2\xi} (\mathcal{H}_{d,r}^s;D) -2W_n(\mathcal{H}_{d,r}^s;D)-\xi t.
\end{equation}
where $Q_{2\xi}(\mathcal{H}_{d,r}^s;D)= \inf_{X\in \mathcal{H}_{d,r}^s}\mathbbm{P}(|\big<\alpha\alpha^*,X\big>| \geq 2\xi)$,
$W_n(\mathcal{H}_{d,r}^s;D) = \mathbbm{E} ~\sup_{X\in \mathcal{H}_{d,r}^s}\big<h,u\big>.$ In $W_n(\mathcal{H}_{d,r}^s;D)$, $h = \sum_{k=1}^n \frac{\varepsilon_k}{\sqrt{n}}\alpha_k\alpha_k^*$ with i.i.d. rademacher variables $\varepsilon_k$. Apply Lemma \ref{lemma2} to nonnegative random variable $\big<\alpha \alpha^*,X\big>^2$ and set $ t = \frac{1}{2}\mathbbm{E}\big<\alpha \alpha^*,X\big>^2$, we obtain
\begin{equation}
    \label{519add}
    \mathbb{P} \big(\big<\alpha \alpha^*,X\big>^2 \geq \frac{1}{2}\mathbbm{E}\big<\alpha \alpha^*,X\big>^2\big)\geq \frac{\big(\mathbbm{E}\big<\alpha\alpha^*,X\big>^2\big)^2}{4\mathbbm{E}\big(\big<\alpha\alpha^*,X\big>^4\big)}.
\end{equation}
Let $\alpha = \beta + \ii \gamma$, then we have 
$$\big<\alpha\alpha^*,X\big> = \begin{bmatrix} \beta^{\top}&\gamma^{\top}\end{bmatrix}\begin{bmatrix} X^\mathcal{R}&-X^{\mathcal{I}} \\X^{\mathcal{I}} &X^{\mathcal{R}} \end{bmatrix}\begin{bmatrix} \beta \\\gamma\end{bmatrix}: = \widetilde{\alpha}^{\top}\widetilde{X}\widetilde{\alpha}= \sum_{i=1}^{2d}\sum_{j=1}^{2d}\tilde{\alpha}_i \tilde{\alpha}_j \tilde{X}_{ij}.$$
Evidently, $\| \widetilde{X}\|_F = \sqrt{2}$. Recall that $\tilde{\alpha}_1,...,\tilde{\alpha}_{2d}$ are i.i.d. copies of $\mathcal{M}$. Combining with the forth moment condition, we apply some algebra to bound $\mathbbm{E}\big<\alpha\alpha^*,X\big>^2$ from below  
\begin{equation}
    \begin{aligned}
    \label{519add1}
   & \mathbbm{E}\big<\alpha\alpha^*,X\big>^2 = \mathbbm{E} \sum_{i,j}\sum_{p,q} \tilde{\alpha}_i\tilde{\alpha}_j\tilde{\alpha}_p\tilde{\alpha}_q\tilde{X}_{ij}\tilde{X}_{pq}  \\
    &= \mathbbm{E}(\mathcal{M}^4)\sum_{i} \tilde{X}_{ii}^2 + \mathbbm{E}(\mathcal{M}^2)^2\sum_{i\neq j}\big(\tilde{X}_{ii}\tilde{X}_{jj}+2\tilde{X}_{ij}^2\big)   \\
    &=\big[ \mathbbm{E}\mathcal{M}^4-\big(\mathbbm{E}\mathcal{M}^2\big)^2\big] \sum_i \tilde{X}_{ii}^2 + 2\big(\mathbbm{E}\mathcal{M}^2\big)^2\sum_{i\neq j}\tilde{X}_{ij}^2 +  \big(\mathbbm{E}\mathcal{M}^2\big)^2 \big(\sum_{i}\tilde{X}_{ii}\big)^2 \\ 
    &\geq  \min\big\{ \mathbbm{E}\mathcal{M}^4-\big(\mathbbm{E}\mathcal{M}^2\big)^2,2\big(\mathbbm{E}\mathcal{M}^2\big)^2\big\}\|\tilde{X}\|_F^2 + \big(\mathbbm{E}\mathcal{M}^2\big)^2 \big(\sum_{i}\tilde{X}_{ii}\big)^2 \\ 
    & \geq 2\min\big\{ \mathbbm{E}\mathcal{M}^4-\big(\mathbbm{E}\mathcal{M}^2\big)^2,2\big(\mathbbm{E}\mathcal{M}^2\big)^2\big\}:= C_1.
    \end{aligned}
\end{equation}
	To control $\mathbbm{E}\big<\alpha\alpha^*,X\big>^4$ we let $G = \big<\alpha\alpha^*,X\big> - \mathbbm{E}\big<\alpha\alpha^*,X\big>$, and note that 
	\begin{equation}
	    \nonumber
	    \begin{aligned}
	        \mathbbm{E}\big<\alpha\alpha^*,X\big>^4 = \mathbbm{E}\big(G+\mathbbm{E}\big<\alpha\alpha^*,X\big>\big)^4 \leq 8\mathbbm{E}G^4 + 8\big(\mathbbm{E}\mathcal{M}^2\big)^4\big(\sum_i \tilde{X}_{ii}\big)^4.
	    \end{aligned}
	\end{equation}
Recall that $\|\mathcal{M}\|_{\psi_2}=O(1)$, $\|\widetilde{X}\| \leq \|\widetilde{X}\|_F = \sqrt{2}$. By Lemma \ref{lemma3}, for any $t>0$ we have
	$$\mathbbm{P}\big(|G| \geq t\big)\leq 2\exp\big(-C_2\min\big\{{t^2} ,{t} \big\}\big) \leq 2\exp\big(-{C_2t^2} \big)+2\exp\big(- {C_2t} \big).$$
	Thus, an integral calculation delivers $\mathbbm{E}[G^4] =   4\int_{0}^\infty t^3\mathbbm{P}(G\geq t) \mathrm{d}t    \leq C_3.$
	So we have 
	\begin{equation}
	    \label{519add2}
	    \mathbbm{E} \big<\alpha\alpha^*,X\big>^4 \leq 8C_3  + 8\big(\mathbbm{E}\mathcal{M}^2\big)^4\big(\sum_i \tilde{X}_{ii}\big)^4
	\end{equation}
	Putting (\ref{519add1}) and (\ref{519add2}) into (\ref{519add}), we have the lower bound
	$$\mathbbm{P}\big(\big<\alpha\alpha^*,X\big>^2\geq \frac{1}{2}C_1\big)\geq \frac{\big(C_1+\big(\mathbbm{E}\mathcal{M}^2\big)^2 \big(\sum_i \tilde{X}_{ii}\big)^2\big)^2}{8C_3 + 8 \big(\mathbbm{E}\mathcal{M}^2\big)^4 \big(\sum_i \tilde{X}_{ii}\big)^4}\geq \frac{1}{8}\min\big\{\frac{C_1^2}{C_3 },1\big\}:=C_4.$$
Since the estimation holds for any $X\in \mathcal{H}_{d,r}^s$, and the event implies $\big|\big<\alpha\alpha^*,X\big>\big| \geq \frac{\sqrt{C_1}}{\sqrt{2}}$, we have 
	\begin{equation}
	    \label{3.25}
	    Q_{\sqrt{C_1}/\sqrt{2}}(\mathcal{H}_{d,r}^s;D) \geq C_4.
	\end{equation}
We now invoke (\ref{2.10}) of Lemma \ref{lemma5} to $h$, since the $C\sqrt{d}$ dominates the right hand side of (\ref{2.10}) with high probability (e.g., apply Hoeffding's inequality to $|\sum_{k=1}^n \varepsilon_k|$), we have $\mathbbm{E}\|h\|\leq C_5 \sqrt{d}$ for some $C_5>0$. Thus, it immediately follows that
	\begin{equation}
	    \label{3.26}
	    W_n(\mathcal{H}_{d,r}^s;D) \leq \mathbbm{E} \|h\| \sup_{X\in \mathcal{H}_{d,r}^s} \|X\|_{nu}   \stackrel{(\ref{nunormbound})}{\leq }  C_5 \sqrt{rd}. 
	\end{equation}
	Now we plug (\ref{3.25}) and (\ref{3.26}) in (\ref{3.24}) and set $\xi = \frac{\sqrt{C_1}}{2\sqrt{2}}$, $t = \frac{C_4}{2}\sqrt{n}$, then it gives
	$$\mathbbm{P} \big( \inf_{X\in \mathcal{H}_{d,r}^s}\frac{1}{\sqrt{n}} \|\mathcal{D}(X)\|_1 \geq \frac{\sqrt{C_1}C_4}{4\sqrt{2}}\sqrt{n}-2C_5\sqrt{rd}\big)\geq 1-\exp\big(-\frac{C_4^2}{8}n\big).$$
	Immediately, $\inf_{X\in \mathcal{H}_{d,r}^s}\|\mathcal{D}(X)\|_1 \geq \frac{1}{2}C_6n$ follows when $n\geq C_6rd$ with sufficiently large $C_6$.
	It remains to show the upper bound for $\sup_{X\in\mathcal{H}_{d,r}^s}\| \mathcal{D}(X)\|_1$. Noting that $\|D\| \leq \|D^\mathcal{R}\| + \|D^\mathcal{I}\|$, by Theorem 4.4.5 in \cite{vershynin2018high} with probability at least $1-2\exp(-n)$ we have $\|D\| \leq C_7\sqrt{n}$. Under this condition, we note that any $X\in \mathcal{H}^s_{d,r}$ can be written as $X = \sum_{l = 1}^{r }s_l \beta_l\beta_l^* $ with $\sum_{l=1}^r s_l^2 = 1$, $\|\beta_l\|=1$. Then some algebra gives 
	\begin{equation}
	    \nonumber
	    \begin{aligned}
	        \| \mathcal{D}(X)\|_1 &= \sum_{k=1}^n\big|\big<\alpha_k\alpha_k^*, \sum_{l = 1}^{r } s_l  \beta_l\beta_l^*\big>\big|   \leq \sum_{k=1}^n \sum_{l=1}^{r } s_l |\alpha_k^*\beta_l|^2  \leq \sum_{l=1}^{r}s_l \|D\beta_l\|^2 \leq \sum_{l=1}^r s_l \|D\|^2 \leq C_7^2\sqrt{r}n.
	    \end{aligned}
	\end{equation}
	The result follows by taking supremum over $X \in \mathcal{H}_{d,r}^s$.	\hfill $\square$
	\begin{rem}
	\label{side1}
	The bound $\|\mathcal{D}(X)\|_1 \geq Cn$ itself is slightly stronger than Theorem 2.3 in \cite{huang2020performance}, and merits the extension from complex Gaussian measurement to sub-Gaussian measurement. This extension is due to a finer treatment of $Q_{2\xi}(\mathcal{H}_{d,r}^s;D)$ independent of the unitary invariance of complex Gaussian distribution, but we do not claim any novelty since similar argument has already been seen in \cite{krahmer2020complex} to analyse Phaselift under sub-Gaussian measurement. 
	However, their result is restricted to $r = 2$, so we still did the incremental extension to $r\in [d]$ for our intended extension to low-rank matrix sensing. 
	\end{rem}
	\vspace{2mm}


    With property of the rank-1 frame given in Theorem \ref{theorem6}, we are now at a position to derive general error bounds for arbitrary $\eta$. In (a) of the next Theorem, the bound $O\big(\frac{\|\eta\|}{\sqrt{n}}\big)$ is already known in \cite{soltanolkotabi2014algorithms,huang2020performance} for complex Gaussian measurement. Here, it is carried over to sub-Gaussian measurement by exactly the same proof.

    In part (b), we establish a new error bound by a different treatment to the right hand side of the optimal condition. A key ingredient is the upper bound for $\|\sum_{k=1}^n r_k\alpha_k\alpha_k^*\|$ established in Lemma \ref{lemma5}. The new bound could be essentially sharper in many interesting cases, thereby greatly complementing the existing one $O\big(\frac{\|\eta\|}{\sqrt{n}}\big)$.

    \begin{theorem}
        \label{theorem612}
        In the problem setting of NPR, for arbitrary $\eta = (\eta_1,...,\eta_n)^{\top}$ independent of $D = [\alpha_1,...,\alpha_n]^*$, there exist $C_i$ such that when $n\geq C_1d$, we have the following error bounds:
        \vspace{1mm}

        \noindent{{\rm (a)}} With probability at least $1-\exp(-C_2n)$, we have
     \begin{equation}
         \|\widehat{x}\widehat{x}^*-x_0x_0^*\|_F \leq \frac{C_3\|\eta\|}{\sqrt{n}}.
         \label{613.2}
     \end{equation}

        \noindent{{\rm (b)}} Let $\mathbf{1} = (1,1,...,1)^{\top}\in\mathbb{R}^n$. With probability at least $1-\exp(-C_4d)$, we have 
        \begin{equation}
            \label{612.8}
            \|\widehat{x}\widehat{x}^*-x_0x_0^*\|_F \leq C_5\Big( \|\eta\|_\infty \sqrt{\frac{d}{n}}+ \frac{|\mathbf{1}^{\top}\eta|}{{n}}\Big).
        \end{equation}
   
    \end{theorem}

 \noindent{\it Proof.} By the optimality of $\widehat{x}$ we have $\sum_{k=1}^n \big( y_k - |\alpha_k^*\widehat{x}|^2\big)^2\leq \sum_{k=1}^n \big(y_k - |\alpha_k^*x_0|^2\big)^2.$ We further use $y_k = |\alpha_k^*x_0|^2 +\eta_k$, then by some algebra it yields
 \begin{equation}
     \label{612.5}
     \sum_{k=1}^n \big<\alpha_k\alpha_k^*,  \widehat{x}\widehat{x}^*-x_0x_0^*\big>^2 \leq  2\sum_{k=1}^n \eta_k \big<\alpha_k\alpha_k^*, \widehat{x}\widehat{x}^*-x_0x_0^*\big>,
 \end{equation}
 or a more compact form $\|\mathcal{D}(\widehat{x}\widehat{x}^*-x_0x_0^*)\|^2 \leq 2\eta^\top \mathcal{D}(\widehat{x}\widehat{x}^*-x_0x_0^*)$. By Theorem \ref{theorem6} there exist $C_i$ such that when $n\geq C_1d$, with probability at least $1-\exp(-C_2n)$ we have $\inf_{X\in\mathcal{H}_{d,2}^s}\|\mathcal{D}(X)\|_1 \geq C_3n$, which implies $\inf_{X\in \mathcal{H}_{d,2}^s}\|\mathcal{D}(X)\|\geq C_3\sqrt{n}$ by Cauchy-Schwarz inequality. We assume $d(\widehat{x},x_0)\neq 0$ without losing any generality.
 
 \vspace{0.5mm}
 
 \noindent{(a)} Since $\|\mathcal{D}(\widehat{x}\widehat{x}^*-x_0x_0^*)\|^2 \leq 2\eta^\top \mathcal{D}(\widehat{x}\widehat{x}^*-x_0x_0^*) \leq 2\|\eta\| \|\mathcal{D}(\widehat{x}\widehat{x}^*-x_0x_0^*)\|$, and with high probability it holds that $\|\mathcal{D}(\widehat{x}\widehat{x}^*-x_0x_0^*)\| \geq d(\widehat{x},x_0) \inf_{X \in \mathcal{H}_{d,2}^s} \| \mathcal{D}(X)\|\geq C_3\sqrt{n}d(\widehat{x},x_0)>0$. Thus, we have $\|\mathcal{D}(\widehat{x}\widehat{x}^*-x_0x_0^*)\|\leq 2\|\eta\|$. Using $\|\mathcal{D}(\widehat{x}\widehat{x}^*-x_0x_0^*)\| \geq C_3\sqrt{n}d(\widehat{x},x_0)$ again gives the desired error bound.
 
 
 \vspace{0.5mm}
 
 \noindent
 (b) We give a different treatment to the right hand side of (\ref{612.5}). Specifically, we invoke Lemma \ref{lemma5} and know that for some $C_4,C_5$, with probability at least $1-2\exp (-C_4d)$,
    \begin{equation}
        \label{612.9}
            \Big\| \sum_{k=1}^n \frac{\eta_k}{\sqrt{n}}\alpha_k\alpha_k^*\Big\|\leq C_5\Big(\|\eta\|_\infty\sqrt{d}+\frac{|\mathbf{1}^{\top}\eta|}{\sqrt{n}}\Big).
    \end{equation}
    Thus, we can upper bound the right hand side of (\ref{612.5}) via
	\begin{equation}
	    \nonumber
	   \begin{aligned}
	        &\big<\sum_{k=1}^n\eta_k\alpha_k\alpha_k^*,\widehat{x}\widehat{x}^*-x_0x_0^*\big>\leq \Big\|\sum_{k=1}^n\eta_k\alpha_k\alpha_k^*\Big\|d(\widehat{x},x_0)\sup_{X\in \mathcal{H}_{d,2}^s}\|X\|_{nu}\\ 
	        &\stackrel{(\ref{nunormbound})}{\leq}\sqrt{2}d(\widehat{x},x_0)\Big\| \sum_{k=1}^n\eta_k\alpha_k\alpha_k^*\Big\|   \stackrel{(\ref{612.9})}{\leq} \sqrt{2}C_5d(\widehat{x},x_0) ({\|\eta\|_\infty\sqrt{nd}+|\mathbf{1}^{\top}\eta|}).
	   \end{aligned}
	\end{equation}
	Combining with the lower bound for the left hand side of (\ref{612.5}), $\|\mathcal{D}(\widehat{x}\widehat{x}^*-x_0x_0^*)\|^2\geq C_3^2n d(\widehat{x},x_0)^2$, we conclude the proof. \hfill $\square$
	
	\vspace{1mm}
	
	For two different error bounds obtained in Theorem \ref{theorem612}, it is of particular interest to compare which one is tighter, and we remark that this depends on the spikiness of $\eta$.
	
	\begin{rem}
	\label{remark613}
	  Cauchy-Schwarz inequality gives $|\mathbf{1}^{\top}\eta|/n \leq \|\eta\|/\sqrt{n}$. Thus, it suffices to compare (\ref{613.2}), $\mathscr{B}_1 = \|\eta\|/\sqrt{n}$, and the first term of (\ref{612.8}), $\mathscr{B}_2 = \|\eta\|_\infty\sqrt{d/n}$. Note that 
	\begin{equation}
	    \label{nonumber}
	    \frac{\mathscr{B}_2 }{\mathscr{B}_1} = \sqrt{\frac{d}{n}}\frac{\sqrt{n}\|\eta\|_\infty}{\|\eta\|} := \sqrt{\frac{d}{n}}\alpha^*(\eta),
	\end{equation}
    		where $\alpha^*(\eta) = (\sqrt{n}\|\eta\|_\infty)/\|\eta\|\in [1,\sqrt{n}]$ characterizes the "spikiness" of noise, i.e., to what level the energy of noise concentrates on a small amount of measurements. For low-spiky $\eta$ satisfying $\alpha^*(\eta)= \tilde{O}(1)$, $\mathscr{B}_2$ (hence (\ref{612.8})) would be tighter, especially when the sample size is sufficiently large. This is often the case of i.i.d. random noise. For highly spiky $\eta$ that may admit $\alpha^*(\eta) = \Omega(\sqrt{n})$, like sparse noise or impulsive noise, (\ref{613.2}) is the sharper bound. For clearer exposition, concrete examples will be presented in numerical experiments. Beyond that, it is worth pointing out that the spikiness, especially spikiness of the desired signal, is closely related to some other statistical estimation problems such as matrix completion \cite{negahban2012restricted,chen2022color}, phase retrieval \cite{krahmer2017phase,krahmer2020complex} and so forth.   
	\end{rem}
     For arbitrary noise, it shall be interesting to further investigate whether we can sharpen the two separate bounds, $O\big(\frac{\|\eta\|}{\sqrt{n}}\big)$ and $O\big(\|\eta\|_\infty\sqrt{\frac{d}{n}}+\frac{|\mathbf{1}^\top\eta|}{n}\big)$, to a unified and tighter one.

	 It is worth pointing out that   both error bounds simultaneously accommodate all $x_0\in \mathbb{C}^d$, since the proof deal with $x_0$ by taking supremum or infimum over $\mathcal{H}_{d,2}^s$ and thus does not rely on a specific $x_0$. Noting that letting $\eta\to 0$ in the bounds implies exact reconstruction in the noiseless case, thus implying a sample complexity of $O(d)$ suffices to reconstruct all $x_0$ without non-trivial ambiguity. From merely the perspective of sample complexity, the line of works on measurement number (e.g.,\cite{balan2006signal,bandeira2014saving,conca2015algebraic}) shows sharp constant, i.e., $n \approx 4d $ suffices for the complex case. On the other hand, the limitation of algebraic methods in these works is lack of guarantee on stability. 
	\subsection{NPR with Light-tailed Random Noise}	
	
  Recall that $O\big(\frac{\|\eta\|}{\sqrt{n}}\big)$, $O\big(\|\eta\|_\infty\sqrt{\frac{d}{n}} + \frac{|\mathbf{1}^{\top}\eta|}{n} \big)$ hold for arbitrary $\eta$ independent of $\alpha_k$, a natural idea is to invoke one of them to obtain reconstruction guarantee for random noise. For sub-Gaussian (or sub-exponential) noise, one may show the spikiness scales as $\sqrt{\log n}$ (or $\log n$), i.e., $\alpha^*(\eta) = \tilde{O}(1)$. Thus, it is suggested in Remark \ref{remark613} to use the new error bound obtained in this work.

	\begin{theorem}
	    \label{theorem7}
	   We consider {NPR} under zero-mean random noise $\eta$ independent of $D =[\alpha_1,...,\alpha_n]^*$. For some $C_i$, when $n\geq C_1d$: 
	   
	   \vspace{0.1mm}
	   
	   \noindent{\rm (a)} If $\eta$ is sub-Gaussian and $\|\eta\|_{\psi_2} = O(1)$, we have $ \|\widehat{x}\widehat{x}^* - x_0x_0^*\|_F \leq    C_2\sqrt{\frac{d\log n }{n}}$ with probability at least $1-C_3\exp(-C_4 d)-2n^{-49}.$
 	   
	   \noindent{\rm (b)} If $\eta_k$ are independent sub-exponential noise satisfying $\max_{k\in [n]}\|\eta_k\|_{\psi_1}=O(1)$, we have $\|\widehat{x}\widehat{x}^*-x_0x_0^*\|_F \leq C_4\sqrt{\frac{d(\log n)^2}{n}}$ with probability at least $1-C_3\exp\big(-C_4d\big) - 2n^{-9}$.
	\end{theorem}
	
	\noindent{\it Proof.} By (b) of Theorem \ref{theorem612} we first assume when $n\geq C_1d$, with probability at least $1-\exp(-C_2d)$, we have $\|\widehat{x}\widehat{x}^*-x_0x_0^*\| = O\big(\|\eta\|_\infty\sqrt{\frac{d}{n}}+\frac{|\mathbf{1}^\top\eta|}{n}\big)$. It remains to estimate $\|\eta\|_\infty,|\mathbf{1}^\top\eta|$ in each situation.

	\vspace{0.5mm}
	
	\noindent{(a)} Since $\max_k \|\eta_k\|_{\psi_2}\leq \|\eta\|_{\psi_2} = O(1)$, by Proposition \ref{pro4}, for some $C_3>0$, with probability at least $1-2n^{-49}$  we have $ \|\eta\|_{\infty}\leq C_3 \sqrt{\log n} $. Moreover, by noting $\|\mathbf{1}^{\top}\eta/\sqrt{n}\|_{\psi_2}\leq \|\eta\|_{\psi_2}= O(1)$ and $\mathbbm{E}(\mathbf{1}^{\top}\eta)=0$, (a) of Proposition \ref{pro1} gives $\mathbbm{P}\big(\frac{|\mathbf{1}^\top\eta|}{n}\geq t\big) \leq 2\exp \big(-C_4nt^2\big)$ for any $t>0.$ We set $t =  \sqrt{\frac{d}{n}}$ and obtain $\frac{|\mathbf{1}^{\top}\eta|}{n}\leq  \sqrt{\frac{d}{n}}$ with probability at least $1-2\exp(-C_4 d)$. By putting the upper bounds for $\|\eta\|_\infty$, $\frac{|\mathbf{1}^{\top}\eta|}{n}$ into $O\big(\|\eta\|_\infty\sqrt{\frac{d}{n}}+\frac{|\mathbf{1}^\top\eta|}{n}\big)$, the desired result follows. 
	 
	 \vspace{0.5mm}

	 \noindent{(b)} By Proposition \ref{pro4} with probability at least $1-2n^{-9}$ we have $ \|\eta\|_\infty \leq C_5\|\eta_k\|_{\psi_1}\log n$. Moreover, by Bernstein's inequality we obtain $\mathbbm{P}\big(\frac{|\mathbf{1}^\top\eta|}{n}\geq t\big) \leq 2\exp(-C_6n\min\{t,t^2\})$ for any $t>0$, see Theorem 2.8.1 in \cite{vershynin2018high}) for instance. We set $t =  \sqrt{\frac{d}{n}}$ in the inequality and obtain $\big|\frac{\mathbf{1}^{\top}\eta}{n}\big|\leq  \sqrt{\frac{d}{n}}$ with probability at least $1-2\exp(-C_6d)$. Now we can conclude the proof by putting pieces into $O\big(\|\eta\|_\infty\sqrt{\frac{d}{n}}+\frac{|\mathbf{1}^\top\eta|}{n}\big)$. \hfill $\square$

	  \vspace{1mm}

	  \cite{lecue2015minimax,eldar2014phase} obtained similar results for NPR in the real case. In particular, \cite{lecue2015minimax} proved  $O\big(\sqrt{\frac{d \log n}{n}}\big)$ for independent, sub-Gaussian, zero-mean $\eta_k$. However, the proofs in both works are not valid for the complex case. Here, Theorem \ref{theorem7} finds the missing complex counterpart of the error bound obtained in \cite{lecue2015minimax,eldar2014phase} for $x_0\in \mathbb{R}^n$, and the condition on noise is slightly weakened, i.e., (a) of Theorem \ref{theorem7} allows noise correlations. Indeed, such indirect approach based on the bound for arbitrary noise enables a very clean proof as well as easy recognition of unnecessary assumption. We will again follow similar route in NGPR.

	 As a side contribution, we also establish the error bound for sub-exponential noise. By contrast, it is unknown whether the proofs in \cite{lecue2015minimax,eldar2014phase} can yield error bound in such sub-exponential regime.

	 \subsection{NPR with Heavy-tailed Random Noise}
	 
	 	In this subsection, we consider NPR under heavy-tailed random noise with bounded $l$-th moment, i.e., $\mathbbm{E}|\eta_k|^l = O(1)$. Compared with bounded moment constraint of light-tailed distribution (see (b) of Proposition \ref{pro1}, \ref{pro3}), this bounded moment condition captures severer random corruption that may be uncontrollable. Obviously, due to lack of concentration property of $\eta$, two general bounds for arbitrary noise does not deliver any meaningful result directly.

	 The main issue is that the extreme values in heavy-tailed responses are overly influential to the loss function, making the ERM estimator unreliable. To robustify the $\ell_2$ loss, we simply shrink the responses to a pre-specified threshold $\tau$. Note that the effectiveness of the simple truncation has already been seen in high-dimensional regression via minimizing a regularized convex loss function \cite{fan2021shrinkage,zhu2021taming,chen2022high,wang2021robust}, but our problem setting here is quite different, i.e., the objective is non-convex and $x_0$ is unstructured. 

	  Let us start from the precise formulation. By truncating response $y_k$ to be $\widetilde{y}_k =  \sign(y_k)$ $\min\{|y_k|,\tau\}$ with suitable $\tau>0$, we propose the estimator which minimizes a robust $\ell_2$ loss constructed from the truncated responses:
	 \begin{equation}
	     \nonumber
	     \widehat{x}_{ht} = \mathop{\arg\min}_x \sum_{k=1}^n \big(\widetilde{y}_k - |\alpha_k^*x|^2\big)^2 .
	 \end{equation}
	 We assume $\eta_k$ is symmetric, i.e., $\eta_k$ and $-\eta_k$ have the same distribution.

	 For $\mathcal{K} \subset [n]$, we define $\mathcal{D}_{\mathcal{K}}(X)$ be the $|\mathcal{K}|$-dimensional sub-vector of $\mathcal{D}(X)$ constituted by entries with indices in $\mathcal{K}$. In the next Theorem, we establish error bound for the robust ERM estimator. A key estimation in the proof is the upper bound for $\sup_{X}\|\mathcal{D}_{\mathcal{K}}(X)\|_1$ where $\mathcal{K} = \{k\in [n]: |\eta_k| \geq c_0\tau\}$. Combining Markov's inequality and bounded $l$-th moment gives upper bound on $|\mathcal{K}|$, but the issue is that, the entries of $\mathcal{K}$ are still unknown, and there exist many possible $\mathcal{K}$. Therefore, we indeed need an upper bound of $\sup_{\mathcal{K},X}\|\mathcal{D}_{\mathcal{K}}(X)\|_1$. For this purpose, it should be clear that $\sup_{X}\|\mathcal{D}_{\mathcal{K}}(X)\|_1 = O(|\mathcal{K}|)$ shown in Theorem \ref{theorem6} is insufficient, hence we further establish $\sup_{\mathcal{K},X}\|\mathcal{D}_{\mathcal{K}}(X)\|_1 = O\big(|\mathcal{K}|\log n\big)$ in Lemma \ref{lemma8} to close the gap.

	 \begin{theorem}
	     \label{theoremht}
	     Consider NPR with i.i.d. symmetric random noise $\eta_k$ satisfying $\mathbbm{E}|\eta_k|^l = O(1)$. For a \textcolor{black}{specific} $x_0\in\mathbb{C}^d$ with $\|x_0\|=O(1)$, we choose $\tau= C_1\big(\frac{n}{d}\big)^{1/(2l)}\log n$ with sufficiently large $C_1$. Then we can find $C_i$, such that when $n\geq C_2d$, $$\|\widehat{x}\widehat{x}^*_{ht}-x_0x_0^*\|_F\leq C_3\big(\sqrt{\frac{d}{n}}\big)^{1-{1}/{l}}(\log n)^2$$ holds with probability at least $1-2n^{-9}-C_4\exp(-C_5d).$ 
	 \end{theorem}

	 \noindent{\it Proof.} By the optimality of $\widehat{x}_{ht}$ we have $\sum_{k=1}^n(\widetilde{y}_k -|\alpha_k^*\widehat{x}_{ht}|^2)^2 \leq \sum_{k=1}^n(\widetilde{y}_k-|\alpha_k^*x_0|^2)^2.$ Then we plug in $y_k = |\alpha_k^*x_0|^2+\eta_k$, some algebra gives
	 \begin{equation}
	     \label{613.6}
	     \sum_{k=1}^n \big<\alpha_k\alpha_k^*,x_0x_0^*-\widehat{x}_{ht}\widehat{x}_{ht}^*\big>^2 \leq 2\sum_{k=1}^n \big(\widetilde{y}_k-y_k+\eta_k\big)\big<\alpha_k\alpha_k^*,\widehat{x}_{ht}\widehat{x}_{ht}^*-x_0x_0^*\big>.
	 \end{equation} 
     For the left hand side of (\ref{613.6}), by Theorem \ref{theorem6} for some $C_i$ when $n\geq C_2d$, with probability at least $1-2\exp(-C_3d)$, we have \begin{equation}
         \begin{aligned}
             \nonumber
             &\sum_{k=1}^n \big<\alpha_k\alpha_k^*, x_0x_0^*-\widehat{x}_{ht}\widehat{x}_{ht}^*\big>^2 = \|\mathcal{D}(\widehat{x}_{ht}\widehat{x}_{ht}^*-x_0x_0^*)\|^2 \\ \geq &d(\widehat{x},x_0)^2 \inf_{X\in\mathcal{H}_{d,2}^s}  \|\mathcal{D}(X)\|^2 \geq d(\widehat{x},x_0)^2 \inf_{X\in\mathcal{H}_{d,2}^s} \frac{1}{n}\|\mathcal{D}(X)\|^2_1 \geq C_4nd(\widehat{x},x_0)^2
         \end{aligned}
     \end{equation} Thus, it remains to upper bound the right hand side of (\ref{613.6}), which can be decomposed as $ 2({T_1} + {T_2})$ where
     \begin{equation}
         \begin{cases}
         \nonumber
         {T_1}  = \sum_{k=1}^n (\widetilde{y}_k-y_k+\eta_k )\big<\alpha_k\alpha_k^*,\widehat{x}_{ht}\widehat{x}_{ht}^*-x_0x_0^*\big> \mathbbm{1} (|\eta_k|\geq \frac{1}{2}\tau)\\
         {T_2} = \sum_{k=1}^n  (\widetilde{y}_k-y_k+\eta_k )\big<\alpha_k\alpha_k^*,\widehat{x}_{ht}\widehat{x}_{ht}^*-x_0x_0^*\big>\mathbbm{1} (|\eta_k|< \frac{1}{2}\tau)
         \end{cases}
     \end{equation}
	 A similar calculation as (\ref{613add}) shows $\big\| |\alpha_k^*x_0|^2\big\|_{\psi_1} = O\big(\|x_0\|^2\big)= O(1)$. Therefore, by Proposition \ref{pro4} we can assume $\|\mathcal{D}(x_0x_0^*)\|_{\infty}=\max_{k}|\alpha_k^*x_0|^2  \leq C_5\ \log n\leq C_5\big(\frac{n}{d}\big)^{1/(2l)}\log n$ with probability at least $1-2n^{-9}$. On this event, we can set $\tau = C_1\big(\frac{n}{d}\big)^{1/(2l)}\log n$ with sufficiently large $C_1$ and assume $\frac{1}{2}\tau > \|\mathcal{D}(x_0x_0^*)\|_{\infty}$. Let us proceed the proof on this event.

	  We define $\mathcal{I}(\tau) = \{k\in [n]:|\eta_k|\geq \frac{1}{2}\tau\}$, i.e., the measurement indices involved in $T_1$. Apply Hoeffding's inequality (e.g., Theorem 2.2.6 in \cite{vershynin2018high}) to  $\mathbbm{1}\big(|\eta_k| \geq \frac{1}{2}\tau \big)$, we obtain
	\begin{equation}
	\label{add1}
	    \mathbbm{P}\Big(\Big|\frac{1}{n}\sum_{k=1}^n \mathbbm{1}(|\eta_k|\geq \frac{1}{2}\tau)-\mathbbm{P}\big(|\eta_k| \geq \frac{1}{2}\tau\big)\Big|\geq t\Big) \leq 2\exp\big(-2nt^2\big),~~\forall ~t>0.
	\end{equation}
	By Markov's inequality and then plug in $\tau$, we have 
\begin{equation}
\label{613.add2}
    \mathbbm{P}\big(|\eta_k|\geq \frac{1}{2}\tau \big)\leq \frac{2^l \mathbbm{E}|\eta_k|^l}{\tau^l} \leq \frac{1}{2}C_6\sqrt{\frac{d}{n}}.
\end{equation}
    Now we set $t = \frac{1}{2}C_6\sqrt{\frac{d}{n}}$ in (\ref{add1}), it gives $|\mathcal{I}(\tau)| \leq  C_6\sqrt{nd}\leq \zeta_0\sqrt{nd}$ with probability at least $1-2\exp(-\Omega(d)).$ Here, we select a fixed $\zeta_0> C_6$ such that $\zeta_0\sqrt{nd}$ is positive integer. On this event with high probability, we find $\mathcal{J}(\tau)$ such that $\mathcal{I}(\tau)\subset \mathcal{J}(\tau)\subset [n]$, $|\mathcal{J}(\tau)| = \zeta_0\sqrt{nd}$.  By Lemma \ref{lemma8}, with probability at least $1-2\exp(-\Omega(\sqrt{nd}))$ we have 
	 \begin{equation}
	     \label{613.9}
	     \begin{aligned}
	          &T_1 \leq \big(\max_k |\widetilde{y}_k|+ \|\mathcal{D}(x_0x_0^*)\|_\infty\big) d(\widehat{x}_{ht},x_0)\sup_{X \in \mathcal{H}_{d,2}^r}\sum_{k\in \mathcal{J}(\tau)} \big|\big<\alpha_k\alpha_k^*,X\big>\big| \\
	          &\leq \frac{3}{2}\tau d(\widehat{x}_{ht},x_0) \sup_{\mathcal{K}\in \mathcal{P}(\zeta_0)}\sup_{X \in \mathcal{H}_{d,2}^r}\sum_{k\in \mathcal{K}} \big|\big<\alpha_k\alpha_k^*,X\big>\big|\leq C_7\tau d(\widehat{x}_{ht},x_0) \sqrt{nd}\log \big(\frac{n}{d}\big)
	     \end{aligned}
	 \end{equation}
	 It remains to bound $T_2$ from above. When $|\eta_k|<\frac{1}{2}\tau$, we have $|y_k| \leq |\alpha_k^*x_0|^2+|\eta_k|\leq \tau$, so the definition of $\widetilde{y}_k$ gives $\widetilde{y}_k = y_k$. Therefore, under the same probability we can deal with $T_2$ as 
	 \begin{equation}
	     \nonumber
	     \begin{aligned}
	         &T_2  =  \sum_{k=1}^n \eta_k\big<\alpha_k\alpha_k^*,\widehat{x}_{ht}\widehat{x}_{ht}^*-x_0x_0^*\big>\mathbbm{1}\big(|\eta_k|< \frac{1}{2}\tau\big)  = \Big<\sum_{k=1}^n\big(\eta_k\mathbbm{1}(|\eta_k|<\frac{\tau}{2})\big)\alpha_k\alpha_k^*,\widehat{x}_{ht}\widehat{x}_{ht}^*-x_0x_0^*\Big> \\ &\stackrel{(\ref{nunormbound})}{\leq }  \sqrt{2}d(\widehat{x}_{ht},x_0)\Big\|{\sum_{k=1}^n\big(\eta_k\mathbbm{1}(|\eta_k|<\frac{\tau}{2})\big)\alpha_k\alpha_k^*}\Big\| \stackrel{(\ref{2.9})}{\leq} C_8 d(\widehat{x}_{ht},x_0)\Big(\sqrt{nd}\tau + |\sum_{k=1}^n \eta_k\mathbbm{1}(|\eta_k|<\frac{\tau}{2})|\Big),
	     \end{aligned}
	 \end{equation}
	 where we rule out probability $1-2\exp(-C_{9}d)$ and apply (\ref{2.9}) in Lemma \ref{lemma5} in the last inequality. Note that symmetric $\eta_k$ implies zero-mean $\eta_k\mathbbm{1}(|\eta_k|<\frac{\tau}{2})$, so Hoeffding's inequality (e.g., Theorem 2.2.6 in \cite{vershynin2018high}) yields $$\mathbbm{P}\Big(\big|\sum_{k=1}^n \eta_k \mathbbm{1}(|\eta_k|<\frac{\tau}{2})\big|\geq t\Big) \leq  2\exp\big(-\frac{2t^2}{n\tau^2}\big),~~\forall~t>0.$$
	 We set $t = \sqrt{nd}\tau$ and know that $|\sum_{k=1}^n \eta_k\mathbbm{1}(|\eta_k|<\frac{\tau}{2})| \leq \sqrt{nd}\tau$ with probability at least $1-2\exp(-2d)$. Putting things together, we conclude that  $$\mathbbm{P}\big(T_2 \leq 2C_9d(\widehat{x}_{ht},x_0)\sqrt{nd}\tau\big)\geq 1-2n^{-9}-C_{10}\exp(-C_{11}d).$$ Combining this with (\ref{613.9}), we can bound the right hand side of (\ref{613.6}) as $$2(T_1+T_2) \leq \big(4C_9+6C_8^2\big)\sqrt{nd}\tau d(\widehat{x}_{ht},x_0)\log\big(\frac{n}{d}\big).$$ We finally substitute this together with lower bound of left hand side into (\ref{613.6}), it gives
	 $$d(\widehat{x}_{ht},x_0)=O\Big(\big(\sqrt{\frac{d}{n}}\big)^{1-\frac{1}{l}}\log \big(\frac{n}{d}\big)\log n\Big),$$
	 which implies the desired result.
	 \hfill $\square$
	 	
	 \begin{rem}
	 \label{notaccom}
	 To confirm the truncation mainly shrinks the noise part, the analysis hinges on a crucial condition $\tau = \Omega(\|\mathcal{D}(x_0x_0^*)\|_\infty)$. However, the estimation $\|\mathcal{D}(x_0x_0^*)\|_\infty = O(\log n)$ only holds for a specific (i.e., fixed) $x_0$. Thus, unlike previous uniform results over $\mathbb{C}^d$, the bound in Theorem \ref{theoremht} is for a specific signal. Besides, the purpose of $\log n$ in $\tau$ is to guarantee $\tau = \Omega(\log n)$ when $n$ and $d$ are close, while one already has $\big(\frac{n}{d}\big)^{1/(2l)}= \Omega(\log n)$ if $\frac{n}{d}$ is relatively large. In this situation, we recommend the choice $\tau = \Theta\big(\big[\frac{n}{d}\big]^{1/(2l)}\big)$.
	 \end{rem}

	 The obtained error bound merits a near optimal sample complexity to achieve $o(1)$ error, but admittedly, the rate $\tilde{O}\big(\big[\sqrt{\frac{d}{n}}\big]^{1-1/l}\big)$ is essentially slower than $\tilde{O}\big(\sqrt{\frac{d }{n}}\big)$ obtained in sub-Gaussian noise. However, we are not aware of any comparable result, and in fact, this seems to be the first attempt of phase retrieval under heavy-tailed noise with bounded $l$-th moment. Therefore, it is still open whether faster rate can be achieved, or a uniform guarantee can be established.

	\subsection{NGPR with Arbitrary Noise}

	We begin with the measurement ensembles used in our problem set-up. In NGPR, $\{A_k\}$ are independent copies of a general Hermitian random matrix $\breve{A}$. Since Hermitian matrix contains $d^2$ real variables, first we need to sample $\breve{A}_0\in \mathbb{R}^{d\times d}$, which is assumed to satisfy Assumption \ref{assumption1}. Then we construct $\breve{A}=[\breve{a}_{ij}]$ from $\breve{A}_0 = [\breve{a}_{0,ii}]$ by letting $\breve{a}_{ii} = \breve{a}_{0,ii}$ when $i\in [d]$, $\breve{a}_{ij}= \frac{1}{\sqrt{2}}( \breve{a}_{0,ij} +\breve{a}_{0,ji} \ii)$ when $i<j$. Naturally, $\breve{a}_{ij} = \overline{\breve{a}_{ji}}$ when $i>j$. We point out that the mechanism for transforming $\breve{A}_0 \in \mathbb{R}^{d\times d}$ to $\breve{A} \in \mathcal{H}_d$ can be quite flexible, i.e., it can also be $\breve{A} = (\breve{A}_0 +\breve{A}_0^{\top}) + (\breve{A}_0 -\breve{A}_0^{\top}) \ii$, but it suffices to consider a specific one since the proof is similar.

\begin{assumption}
\label{assumption1}
The $d^2$-dimensional random vector $\mathrm{vec}(\breve{A}_0)$ is zero-mean and sub-Gaussian with $\| \mathrm{vec}(\breve{A}_0)\|_{\psi_2}=O(1)$. Besides, the covariance matrix $\breve{\Sigma} = \mathbbm{E}(\mathrm{vec}(\breve{A}_0)\mathrm{vec}(\breve{A}_0)^{\top})$ satisfies $\lambda_0\leq \lambda_{\min}(\breve{\Sigma}) \leq \lambda_{\max}(\breve{\Sigma}) \leq \lambda_1$ for some $\lambda_0,\lambda_1>0$ independent of other parameters.  
\end{assumption}

    In brief, just similar to many existing papers, entries of Hermitian $A_k$ are assumed to be sub-Gaussian and zero-mean. Recall that such $\{A_k\}$ is called a full-rank frame in this work.

	By defining the operator $\mathcal{A}(X) = \big(\big<A_1,X\big>,...,\big<A_n,X\big>\big)^{\top}$ for $X\in \mathbb{C}^{d\times d}$, NGPR model can be written in a more compact form $Y = \mathcal{A}(x_0x_0^*) + \eta$ where $Y = (y_1,...,y_n)^{\top},~\eta= (\eta_1,...,\eta_n)^{\top}$ are the response vector, noise vector, respectively. Similarly, $Y= \mathcal{A}(X_0) + \eta$ where $X_0 = U_0U_0^*$ for some $U_0\in \mathbb{C}^{d\times r}$ formulates the more general rank-$r$ PSD matrix sensing.

	Most analyses of low-rank matrix recovery problem rely on RIP condition on the operator $\mathcal{A}(.)$ \cite{recht2010guaranteed,candes2011tight}, i.e., $1-\delta \leq \|\mathcal{A}(X)\|^2 /n \leq 1+\delta$ for all low-rank matrix with unit Frobenius norm and some sufficiently small $\delta >0$. Intuitively, this requires $\frac{\mathcal{A}}{\sqrt{n}}(.)$ to operate on low-rank matrix nearly like an isometry. The well-known random measurement that leads to RIP (with high probability) is (sub)-Gaussian matrix with i.i.d. entries.

	To obtain a uniform bound for all $x\in \mathbb{C}^d$, it turns out that the relaxed condition $C_1\leq  \|\mathcal{A}(X)\|^2/n\leq C_2$ with some $C_1>0$ suffices. Here, $C_1,C_2$ do not need to be extremely close, thus allowing $\mathrm{vec}(\breve{A}_0)$ to have a rather general covariance matrix (see Assumption \ref{assumption1}). In Theorem \ref{theorem1}, the measurement ensemble $\mathcal{A}$ is shown to have this property for all $X\in \mathcal{H}_{d,r}^s$. The idea of the proof is from \cite{candes2011tight}, and key ingredient  is a covering argument of the set $\mathcal{H}_{d,r}^s$.

\begin{theorem}
    \label{theorem1}
    Under Assumption \ref{assumption1}, assume $\{A_k:k\in [n]\}$ are drawn as we stated before.
    Then for a specific positive integer $1\leq r\leq d$, there exist constants $C_i$ such that when $n \geq C_1 rd$, with probability at least $1-2\exp(-C_2n)$, we have $ C_3 \sqrt{n} \leq  \|\mathcal{A}(X)\| \leq C_4 \sqrt{n}$, $\forall~X \in \mathcal{H}_{d,r}^s$.
\end{theorem}

\noindent{\it Proof.} With regard to $\|.\|_F$, we first take a $\delta$-net of $\mathcal{H}_{d,r}^s$, denoted by $\mathcal{N} = \{X_1,...,X_{N_1}\}\subset \mathcal{H}_{d,r}^s$. By Lemma \ref{lemmaadd} we assume $N_1 \leq \big(\frac{13}{\delta}\big)^{3rd}$. Fix $X = [x_{ij}]\in \mathcal{H}_{d,r}^s$, we have $\|\mathcal{A}(X)\|^2 = \sum_{k=1}^n \big<A_k,X\big>^2$ where $A_1,...,A_n$ are i.i.d. copies of $\breve{A} = [\breve{a}_{ij}]$. To see the nature of $\big<\breve{A},X\big>$, some algebra gives 
$\big<\breve{A},X\big> = \sum_{i=1}^d x_{ii}\breve{a}_{ii}+\sum_{i<j}(\sqrt{2}x_{ij}^\mathcal{R})(\textcolor{black}{\sqrt{2}}\breve{a}_{ij}^\mathcal{R}) + \sum_{i<j}(\sqrt{2}x_{ij}^\mathcal{I})(\textcolor{black}{\sqrt{2}}\breve{a}_{ij}^\mathcal{I}).$
Recall that $\breve{A}\in\mathcal{H}_d$ are constructed from $\breve{A}_0 \in \mathbb{R}^{d\times d}$, if we define $\breve{X}= [\breve{x}_{ij}]$ by $\breve{x}_{ii} = x_{ii}$, $\breve{x}_{ij} = \sqrt{2}x_{ij}^\mathcal{R}$ when $i<j$, and $\breve{x}_{ij} = \sqrt{2}x_{ji}^\mathcal{I}$ when $i>j$, then by Assumption \ref{assumption1} and $\|\breve{X}\|_F =1$, we have
$$\mathbbm{E} \Big(\frac{1}{n}\|\mathcal{A}(X)\|^2\Big) =\mathbbm{E}\big<\breve{A},X\big>^2 = \mathbbm{E} \big<\breve{A}_0,\breve{X}\big>^2 = \mathrm{vec}(\breve{X})^{\top}\big(\mathbbm{E} \mathrm{vec}(\breve{A}_0)\mathrm{vec}(\breve{A}_0)^{\top}\big) \mathrm{vec}(\breve{X})\in [\lambda_0,\lambda_1].$$
Moreover, since $\big<\breve{A},X\big> = \big<\breve{A}_0,\breve{X}\big> = \mathrm{vec}(\breve{X})^{\top}\mathrm{vec}(\breve{A})$, so by (\ref{2.6}) we have $\|\big<\breve{A},X\big>^2 \|_{\psi_1} \leq \| \big<\breve{A},X\big>\|^2_{\psi_2}\leq \|\mathrm{vec}(\breve{A})\|^2_{\psi_2}= O(1)$. We now invoke Bernstein's inequality (Theorem 2.8.1 in \cite{vershynin2018high}) and obtain for any $t>0$, it holds that
\begin{equation}
    \nonumber
    \mathbbm{P}\Big( \big|\frac{1}{n}\|\mathcal{A}(X)\|^2 - \mathbbm{E}\big<\breve{A},X\big>^2 \big|\geq t\Big) \leq 2\exp\Big(-C_2n\min\{t^2,t\}\Big).
\end{equation}
Since it holds for any $X \in \mathcal{H}_{d,r}^s$, specifically, we can take $X =X_i \in \mathcal{N}$ and take union bound over $i\in [N_1]$, it yields that $$ \mathbbm{P}\Big(\sup_{i\in [N_1]}\Big|\frac{1}{n}\|\mathcal{A}(X_i)\|^2 - \mathbbm{E}\big<\breve{A},X_i\big>^2\Big|\geq t\Big)\leq 2\exp\Big(-C_2 n\min\{t^2,t\}+ 3rd\log\big(\frac{13}{\delta}\big)\Big).$$
We set $t =\min\{1, \frac{1}{2}\lambda_0\}$, then for sufficiently large $C_3$, when $n \geq C_3rd$, the right hand side of the above inequality is less than $2\exp(-C_4n)$ for some $C_4$. Combining with $\mathbbm{E}\big<\breve{A},X\big>^2\in [\lambda_0,\lambda_1]$, it gives \begin{equation}
    \label{68.1}
    \mathbbm{P}\Big(\frac{1}{2}\lambda_0\leq \frac{\|\mathcal{A}(X_i)\|^2}{n}\leq 2\lambda_1,~\forall ~i\in [N_1]\Big) \geq 1-2\exp(-C_4n).
\end{equation} 
Note that $\mathcal{H}_{d,r}^s$ is compact set, we can assume for some $X_{\max},X_{\min}\in \mathcal{H}_{d,r}^s$
$$ \sup_{X\in \mathcal{H}_{d,r}^s}\|\mathcal{A}(X)\| = \|\mathcal{A}(X_{\max})\| ~~\mathrm{and}~~\inf_{X\in\mathcal{H}^s_{d,r}}\| \mathcal{A}(X)\| = \| \mathcal{A}(X_{\min})\|.$$
Let $\hat{X} = X_{\max}$, there exists $i_0 \in [N_1]$ such that $\| \hat{X}-{X}_{i_0}\|_F \leq \delta$, and we have \begin{equation}
    \label{68.2}
    \begin{aligned}
        &\Big|\|\mathcal{A}(\hat{X})\| - \|\mathcal{A}(X_{i_0})\|\Big| \leq \| \mathcal{A}(\hat{X}-X_{i_0})\|\leq \delta \Big\| \mathcal{A}\big( \frac{\hat{X}-X_{i_0}}{\|\hat{X}-X_{i_0}\|_F}\big)\Big\| \\ 
         =& \delta \| \mathcal{A}(X_{(1)}+X_{(2)})\| \leq \delta \big(\| \mathcal{A}(X_{(1)})\| +\| \mathcal{A}(X_{(2)})\|\big)\leq 2\delta \|\mathcal{A}(X_{\max})\|,
    \end{aligned}
\end{equation}
where we note that $\frac{\hat{X}-X_{i_0}}{\|\hat{X}-X_{i_0} \|_F}\in \mathcal{H}_{d,2r}^s$ and hence can be decomposed as $X_{(1)}+X_{(2)}$ satisfying $X_{(1)},X_{(2)}\in \mathcal{H}_{d,r}$, $\|X_{(1)}\|_F,\|X_{(2)}\|_F\leq 1$. Combining with (\ref{68.1}), with probability at least $1-2\exp(-C_4n)$ we have \begin{equation}
    \label{68.3}
    \| \mathcal{A}(X_{\max})\| \leq \frac{1}{1-2\delta}\| \mathcal{A}(X_{i_0})\| \leq \frac{\sqrt{2\lambda_1}}{1-2\delta}  \sqrt{n}:=C_5\sqrt{n}.
\end{equation}
Evidently, when we let $\hat{X}=X_{\min}$, (\ref{68.2}) also holds for some different $i_0$. Thus, under the same probability, by (\ref{68.1}) and (\ref{68.3}) we have 
\begin{equation}
    \label{68.4}
    \|\mathcal{A}(X_{\min})\| \geq \|\mathcal{A}(X_{i_0})\| - 2\delta \|  \mathcal{A}(X_{\max})\| \geq \big(\sqrt{\frac{\lambda_0}{2}}-\frac{2\delta \sqrt{2\lambda_1}}{1-2\delta}\big) \sqrt{n}:=C_6\sqrt{n}.
\end{equation}
To guarantee $C_6>0$ we only need to set $\delta$ to be suitably small. Thus, (\ref{68.3}) and (\ref{68.4}) conclude the proof. \hfill $\square$

	\vspace{2mm}

	For this specific complex and Hermitian setting, in Theorem \ref{theorem1} of a recent paper \cite{thaker2020sample}, same condition  has been established and termed as $(C_3,C_4)$-stability, but only for $r=2$ and $A_k$ with i.i.d. complex Gaussian entries (which indeed satisfies more stringent RIP condition). We comment that the extension to $r\geq 2$ facilitates our intended extension to more general rank-$r$ setting, and our Assumption \ref{assumption1} on the random measurement ensemble is weaker. Besides, probably their analysis is not sharp since their probability term reads as $1-\xi$ where $\xi$ is a small constant.

With the established property of $A_k$ in place, we are now ready to establish the error bound for an arbitrary noise pattern.

	\begin{theorem}
	    \label{theorem3}
	    In the problem setting of NGPR, for arbitrary $\eta=(\eta_1,...,\eta_n)^{\top}$ independent of $A_k$, there exist constants $C_i$, such that when $n\geq C_1d$, the error bound 
	    \begin{equation}
	        \label{3.9}
	        \|\widehat{x}\widehat{x}^*-x_0x_0^*\|_F \leq C_2\|\eta\| \frac{\sqrt{d}}{n}
	    \end{equation}
	    holds with probability at least $1-5\exp(-C_3d)$. 
	\end{theorem}
	
	\noindent{\it Proof.} We start from the optimal condition $\sum_{k=1}^n\big(y_k - \widehat{x}^*A_k\widehat{x}\big)^2 \leq \sum_{k=1}^n \big(y_k - x_0^*A_k{x_0}\big)^2~.$	
	    Substitute $y_k = x_0^*A_kx_0+\eta_k$ in the optimal condition, some algebra yields  
	\begin{equation}
	    \label{3.10}
	    \sum_{k=1}^n\big<A_k,x_0x_0^* - \widehat{x}\widehat{x}^*\big>^2\leq2\big<\sum_{k=1}^n \eta_kA_k,\widehat{x}\widehat{x}^*-x_0x_0^*\big>.
	\end{equation}
   For the left hand side of (\ref{3.10}), by Theorem \ref{theorem1} we find $C_1,C_2,C_3$ such that when $n \geq C_1d$, with probability at least $1-\exp(-C_3n)$, we have $\sum_{k=1}^n\big<A_k,x_0x_0^* - \widehat{x}\widehat{x}^*\big>^2 \geq C_2 nd(\widehat{x},x_0)^2$.
   For the right hand side of (\ref{3.10}), we first find the upper bound as 
	\begin{equation}
	    \nonumber
	    \begin{aligned}
	        2\big<\sum_{k=1}^n \eta_kA_k,\widehat{x}\widehat{x}^*-x_0x_0^*\big>& \leq 2\|\eta\|d(\widehat{x},x_0) \Big\| \sum_{k=1}^n \frac{\eta_k}{\|\eta\|}A_k\Big\|\sup_{X\in \mathcal{H}_{d,2}^s}\|X\|_{nu}   \stackrel{(\ref{nunormbound})}{\leq }  2\sqrt{2}\|\eta\|d(\widehat{x},x_0) \|H_2\|,
	    \end{aligned}
	\end{equation}
	where we let $H_2= \sum_{k=1}^n \frac{\eta_k}{\|\eta\|}A_k$. Assume Hermitian $A_k$ is constructed from $\breve{A}_{k,0}\in\mathbb{R}^{d\times d}$ satisfying Assumption \ref{assumption1}, and thus $\|\mathrm{vec}(\breve{A}_{2,0})\|_{\psi_2} = O(1)$. Then, evidently, $H_2$ is constructed from $\breve{H}_{2,0}:=\sum_{k=1}^n \frac{\eta_k}{\|\eta\|}\breve{A}_{k,0}$ in the same way (stated in Lemma \ref{lemma4}). By Proposition \ref{pro2} and Definition \ref{def1}, one easily sees that $\|\mathrm{vec}(\breve{H}_{2,0})\|_{\psi_2} = O(1)$. Thus, we invoke Lemma \ref{lemma4} and obtain 	$\mathbbm{P}\big(\|H_2\|\geq C_5   \big(\sqrt{d}+t\big)\big) \leq 4\exp\big(-t^2\big)$ for any $t>0$. We set $t = \sqrt{d}$ in this inequality, and find that $\|H_2\| \leq 2C_5   \sqrt{d}$ with probability at least $1-4\exp(-d)$. This delivers the upper bound for the right hand side of (\ref{3.10}) 
	\begin{equation}
	    \label{3.12}
	   \mathbbm{P}\Big(  2\big<\sum_{k=1}^n \eta_kA_k,\widehat{x}\widehat{x}^*-x_0x_0^*\big> \leq 4\sqrt{2}C_5\|\eta\| \sqrt{d}d(\widehat{x},x_0)\Big) \geq 1-4\exp(-d).
	\end{equation}
	Therefore, with probability at least $1-5\exp\big(-\min \{C_3,1\}d\big)$ we can put the established estimations into both sides of (\ref{3.10}), the result follows.  \hfill $\square$

	\vspace{1mm}

   Similar to Theorem \ref{theorem612}, the bound $O\big(\|\eta\|\frac{\sqrt{d}}{n}\big)$ holds uniformly over $\mathbb{C}^d$. Thus, the implication of letting $\eta \to 0$ is that there exists a frame with measurement number $O(d)$ that allows exact reconstruction of all $x_0\in\mathbb{C}^d$. Note that the $O(d)$ itself is not as sharp as $4d-4$ given in \cite{wang2019generalized}, but in our result stability is guaranteed.

	\subsection{NGPR with Light-tailed Random Noise}
	
	We now go into the regime where $\eta_1,...,\eta_n$ are random noise with sub-Gaussian or sub-exponential tail. We follow a similar route as NPR and derive the error bounds from $O\big(\|\eta\|\frac{\sqrt{d}}{n}\big)$. Compared with a direct treatment on random noise, this approach allows us to nicely identify some inessential assumptions. As a result, the light-tailed random noise is allowed to be biased or correlated in NGPR. 
	
	\begin{theorem}
	    \label{theorem33}
	  We consider {NGPR} under random noise independent of $A_1,...,A_n$, assume $n\geq C_1 d$. For some $C_i$:

	  \noindent{(a)} If the sub-Gaussian vector $\eta$ satisfies $\|\eta-\mathbbm{E}\eta\|_{\psi_2} = O(1)$, and $\|\mathbbm{E}\eta\|_\infty = O(1)$, we have $ \|\widehat{x}\widehat{x}^*-x_0x_0^*\|_F  \leq C_2\sqrt{\frac{d}{n}}$ with probability at least $1-7\exp(-C_3d).$

	  \noindent{(b)} If $\max_k \|\eta_k\|_{\psi_2}= O(1)$, we have $\|\widehat{x}\widehat{x}^*-x_0x_0^*\|_F \leq C_4 \sqrt{\frac{d\log n}{n}}$ with probability at least $1-5\exp(-C_5d)-2n^{-24}$.

	  \noindent{(c)} If $\max_k\|\eta_k\|_{\psi_1}= O(1)$, we have $\|\widehat{x}\widehat{x}^*-x_0x_0^*\|_F \leq C_6\sqrt{\frac{d(\log n)^2}{n}}$ with probability at least $1-5\exp(-C_7d) -2n^{-9}$. 
	\end{theorem}
	
	\noindent{\it Proof.} We first invoke Theorem \ref{theorem3} and find some $C_i$ such that when $n\geq C_1d$, with probability at least $1-5\exp(-C_2d)$, $\|\widehat{x}\widehat{x}^*-x_0x_0^*\|_F = O\big(\|\eta\|\frac{\sqrt{d}}{n}\big)$ holds. Since $\|\eta\| \leq \sqrt{n}\|\eta\|_\infty$, on the same event we have $\|\widehat{x}\widehat{x}^*-x_0x_0^*\|_F = O\big(\|\eta\|_\infty\sqrt{\frac{d}{n}}\big)$.
	
	\vspace{0.1mm}

	\noindent{(a)} Based on $O\big(\|\eta\|\frac{\sqrt{d}}{n}\big)$, it suffices to show $\|\eta\| = O(\sqrt{n})$. Note that $\|\eta\|\leq \|\eta -\mathbbm{E}\eta\| + \|\mathbbm{E}\eta\|$, and it is obvious that $\|\mathbbm{E}\eta\| \leq \|\mathbbm{E}\eta\|_\infty\sqrt{n} = O(\sqrt{n})$. While the upper bound $\|\eta - \mathbbm{E}\eta\| = O(\sqrt{n})$ is standard, see Theorem 1.19 in \cite{rigollet2015high} for instance.
	
	\vspace{0.1mm}
	
	\noindent{(b) and (c)} Based on the bound $O\big(\|\eta\|_\infty\sqrt{\frac{d}{n}}\big)$, we immediately conclude the proof by invoking Proposition \ref{pro4}.
	\hfill $\square$
	
	\begin{rem}
	\label{remark3} We do not require $\eta_k$ to be independent or zero-mean, and all above error bounds guarantee consistency, more precisely, ERM can achieve $o(1)$ error with under near optimal sample size $n=\tilde{\Omega}(d)$. Note that situation (a) accommodates the conventionally assumed setting where $\eta_k$ are i.i.d. sub-Gaussian noise, and $O\big(\sqrt{\frac{d}{n}}\big)$ \textcolor{black}{is just the minimax rate in the real case of rank-1 matrix sensing} \cite{candes2011tight}. Situations (b), (c) make weaker assumptions on $\eta$'s probability tail, e.g., (b) encompasses sub-Gaussian noise with many repeated entries, while this is out of the range of (a). Accordingly, the bound is slightly weakened by logarithmic factor.
 	\end{rem}

	To our best knowledge, \cite{fan2022oracle} is the only work that studied exactly the same setting in the real case. We comment that Theorem \ref{theorem33} represents the all-round improvement of the statistical rate $O\big(\sqrt{\frac{d\log n}{n}}\big)$ obtained in \cite{fan2022oracle}: The proof handles $\mathbb{K}=\mathbb{R},\mathbb{C}$; The bound is sharper (with no logarithmic factor) for i.i.d. sub-Gaussian noise; The inessential zero-mean, independent assumptions are removed.

\subsection{NGPR with Heavy-tailed Random Noise}
	
We now turn to heavy-tailed random noise assumed to have bounded $l$-th moment. By truncating response $y_k$ to be $\widetilde{y}_k = \mathrm{sign}(y_k)\min\{|y_k|,\tau\}$ with $\tau>0$, we propose the estimator that minimizes a robust $\ell_2$ loss 
	\begin{equation}
	   \nonumber
	    \widehat{x}_{ht} = \mathop{\arg\min}_x \sum_{k=1}^n \big(\widetilde{y}_k - x^*A_kx\big)^2 .
	\end{equation}

	We establish estimation guarantee for $ \widehat{x}_{ht}$ in Theorem \ref{theorem5}. 
	
	For $\mathcal{K}\subset [n]$ we let $\mathcal{A}_\mathcal{K}(X)$ be the sub-vector of $\mathcal{A}(X)$ constituted by entries in $\mathcal{K}$.
	Similar to the technical challenge arising in Theorem \ref{theoremht}, we need to bound $\sup_{\mathcal{K},X}\|\mathcal{A}_{\mathcal{K}}(X)\|$ from above. Due to the additional supremum on $\mathcal{K}$, this is beyond the range of upper bound provided in Theorem \ref{theorem1} that reads as $\sup_{X}\|\mathcal{A}_{\mathcal{K}}(X)\| = O(\sqrt{|\mathcal{K}|})$. Instead, we strengthen it to be $\sup_{\mathcal{K},X}\|\mathcal{A}_{\mathcal{K}}(X)\| = O\big(\sqrt{|\mathcal{K}|\log n} \big)$ in Lemma \ref{lemma7}.


\begin{theorem}
    \label{theorem5}
    Consider {NGPR} with i.i.d. random noise satisfying $\mathbbm{E}|\eta_k|^l = O(1)$. For a \textcolor{black}{specific} $x_0\in \mathbb{C}^d$ with $\|x_0\| = O(1)$, we choose $\tau = C_1\big(\frac{n}{d}\big)^{1/(2l)}\sqrt{\log n}$ with sufficiently large $C_1$. Then we can find $C_i$, such that when $n\geq C_2d$, $$\|\widehat{x}_{ht}\widehat{x}_{ht}^* - x_0x_0^*\|_F \leq C_3 \big(\sqrt{\frac{d}{n}}\big)^{1-1/l}\log n$$
    holds with probability at least $1-2n^{-24} -8\exp \big(-C_4 d\big)$.
\end{theorem}

\noindent{\it Proof.} We start from the optimality $\sum_{k=1}^n \big(\widetilde{y}_k - \widehat{x}^*_{ht}A_k \widehat{x}^*_{ht}\big)^2 \leq \sum_{k=1}^n\big(\widetilde{y}_k - x_0^*A_kx_0\big)^2.$ Plug in $y_k = x_0^*A_kx_0+\eta_k$, some algebra gives
	\begin{equation}
	    \label{3.18}
	    \sum_{k=1}^n \big(x_0^*A_kx_0 -\widehat{x}^*_{ht}A_k \widehat{x}_{ht}\big)^2 \leq 2 \sum_{k=1}^n \big(\widetilde{y}_k - y_k + \eta_k\big) \big(\widehat{x}^*_{ht}A_k\widehat{x}_{ht} - x_0^*A_kx_0\big).
	\end{equation}
    By Theorem \ref{theorem1} for some $C_2,C_3,C_4$, when $n \geq C_2 d$, with probability at least $1-2\exp (-C_4n)$, we have $\sum_{k=1}^n \big(x_0^*A_kx_0 -\widehat{x}^*_{ht}A_k \widehat{x}_{ht}\big)^2\geq C_3 n d(\widehat{x}_{ht},x_0)^2$, thus the lower bound for the left hand side of (\ref{3.18}) is established. For the right hand side of (\ref{3.18}) we first decompose it as $ 2({T_1} + {T_2})$ where
\begin{equation}
    \begin{cases}
    \nonumber
{T_1} = \sum_{k=1}^n \big(\widetilde{y}_k - y_k + \eta_k\big) \big(\widehat{x}^*_{ht}A_k\widehat{x}_{ht} - x_0^*A_kx_0\big) \mathbbm{1}\big(|\eta_k|\geq \frac{1}{2}\tau\big); \\
{T_2} = \sum_{k=1}^n \big(\widetilde{y}_k - y_k + \eta_k\big) \big(\widehat{x}^*_{ht}A_k\widehat{x}_{ht} - x_0^*A_kx_0\big) \mathbbm{1}\big(|\eta_k|< \frac{1}{2}\tau\big) . 
  \end{cases}
\end{equation}
Note that $x_0^*A_kx_0 = \big<A_k,x_0x_0^*\big>$, $\|x_0x_0^*\|_F = \|x_0\|^2 =O(1)$, combining with Assumption \ref{assumption1} one easily verifies $ \|x_0^*A_kx_0\|_{\psi_2} \leq C_5$. Thus, we invoke (a) of Proposition \ref{pro4} and can assume $\|\mathcal{A}(x_0x_0^*)\|_\infty \leq C_6\sqrt{\log n}$ holds with probability at least $1-2n^{-24}$. Thus, by choosing $C_1$ sufficiently large, when $n\geq d$ we can assume $\frac{1}{2}\tau > \| \mathcal{A}(x_0x_0^*)\|_\infty$ and proceed the proof under this condition.

We define $\mathcal{I}(\tau) = \{k\in [n]: |\eta|_k\geq \frac{1}{2}\tau\}$, i.e., the measurement indices involved in $T_1$. A same argument as (\ref{add1}) and (\ref{613.add2}) gives $|\mathcal{I}(\tau)| \leq 2C_7\sqrt{nd}\leq \zeta_0\sqrt{nd}$ for some $C_7$ with probability at least $1-2\exp(-\Omega(d)).$ Here, we select a fixed $\zeta_0$ such that $\zeta_0\sqrt{nd}$ is positive integer. We further find ${\mathcal{J}}(\tau)$ such that $ \mathcal{I}(\tau)\subset {\mathcal{J}}(\tau)\subset[n]$ and $|{\mathcal{J}}(\tau)| =\zeta_0\sqrt{nd }$. Then we invoke Lemma \ref{lemma7}, with probability at least $1-2\exp(-\Omega(\sqrt{nd}))$ we have $$
    \sup_{X\in \mathcal{H}^s_{d,2}}\big(\sum_{k\in \mathcal{I}(\tau)} \big<A_k,X\big>^2\big)^{1/2} \leq \sup_{\mathcal{J}(\tau) = \mathcal{P}(\zeta_0)} \sup_{X\in \mathcal{H}^s_{d,2}}\| \mathcal{A}_{\mathcal{K}}(X)\|\leq C_{8}(nd)^{\frac{1}{4}}\sqrt{\log\big(\frac{n}{d}\big)}. 
$$
Now by (\ref{nunormbound}) and Cauchy-Schwarz inequality, we can upper bound $T_1$ as 
\begin{equation}
    \nonumber
    \begin{aligned}
       &{T_1} \leq \Big\{\sum_{k\in \mathcal{I}(\tau)} \big(\widetilde{y}_k -x_0^*A_kx_0\big)^2 \Big\}^{1/2}\sup_{X\in \mathcal{H}_{d,2}^s}\Big\{ \textcolor{black}{\sum_{k\in \mathcal{I}(\tau)} \big<A_k, X\big>^2}\Big\}^{1/2}d(\widehat{x}_{ht},x_0) \\
        &\leq  \Big(\sqrt{|\mathcal{J}(\tau)|(\tau + \frac{\tau}{2})^2}\Big)\Big(C_{8}(nd)^{\frac{1}{4}}\sqrt{\log\big(\frac{n}{d}\big)}\Big)d(\widehat{x}_{ht},x_0)
        \leq 3C_7C_8\tau \sqrt{nd}\sqrt{\log\big(\frac{n}{d}\big)}d(\widehat{x}_{ht},x_0).
    \end{aligned}
\end{equation}
	For the other term ${T_2}$, note that when $|\eta_k|<\frac{1}{2}\tau$, we have $|y_k| = |x_0^*A_kx_0 + \eta_k| \leq |x_0^*A_kx_0|+|\eta_k| \leq \tau$ and hence $y_k = \widetilde{y}_k$. Thus, we have  
	\begin{equation}
	    \nonumber
	   \begin{aligned}
	       { T_2} =& \sum_{k=1}^n \eta_k\mathbbm{1}\big(|\eta_k|<\frac{1}{2}\tau\big) \big<A_k,\widehat{x}_{ht}\widehat{x}_{ht}^* - x_0x_0^*\big>
	    =\Big<\sum_{k=1}^n \eta_k\mathbbm{1}(|\eta_k|<\frac{1}{2}\tau)A_k, \widehat{x}_{ht}\widehat{x}_{ht}^* - x_0x_0^*\Big> \\
	   \stackrel{(\ref{nunormbound})}{\leq }  &2\sqrt{n}d(\widehat{x}_{ht},x_0)\Big\|\sum_{k=1}^n\frac{\eta_k \mathbbm{1}\big(|\eta_k|<\frac{1}{2}\tau\big)}{\sqrt{n}}A_k\Big\| : = 2\sqrt{n}d(\widehat{x}_{ht},x_0)\|H_3\|,
	   \end{aligned} 
	\end{equation}
	where we let $ H_3 = \sum_{k=1}^n \frac{1}{\sqrt{n}}\eta_k\mathbbm{1}(|\eta_k|<\frac{\tau}{2})A_k$. Assume the Hermitian $A_k$ is constructed from $\breve{A}_{k,0}\in \mathbb{R}^{d\times d}$ satisfying $\|\mathrm{vec}(\breve{A}_{k,0})\|_{\psi_2}= O(1)$, then $H_3$ is constructed from $\breve{H}_{3,0} = \sum_{k=1}^n \frac{1}{\sqrt{n}}\eta_k\mathbbm{1}(|\eta_k|<\frac{\tau}{2})\breve{A}_{k,0}$ in the same way (stated in Lemma \ref{lemma4}). By Proposition \ref{pro2} and Definition \ref{def1}, one easily sees that  $\|\mathrm{vec}(\breve{H}_{3,0})\|_{\psi_2} = O(\tau)$. Thus, Lemma \ref{lemma4} gives $\mathbbm{P}(\|H_3\| \geq C_{9}\tau (\sqrt{d}+t)) \leq 4\exp(-t^2). $
	Setting $t= \sqrt{d}$, with probability at least $1-4\exp(-d)$ we have $\|H_3\| \leq 2C_{9}\tau  \sqrt{d}$. Putting it into the previous estimate on $T_2$, we obtain the upper bound for $T_2$ as $$T_2 \leq 4C_9\tau \sqrt{nd} d(\widehat{x}_{ht},x_0).$$

	Finally we put pieces together: Combining the upper bounds for $T_1$, $T_2$ we successfully bound right hand side of (\ref{3.18}); We further count the involved probability terms and plug the established estimations into both sides of (\ref{3.18}), it follows that 
	$$d(\widehat{x}_{ht},x_0)= O\Big(\big(\sqrt{\frac{d}{n}}\big)^{1-\frac{1}{l}}\sqrt{\log \big(\frac{n}{d}\big)}\sqrt{\log n}\Big),$$which implies the desired result. \hfill $\square$
	
	\vspace{1mm}
	 Notably, in Theorem \ref{theorem5} we still allow $\eta_k$ to be biased, but we do require the independent assumption, which is used when we apply Hoeffding's inequality to bound $|\mathcal{I}(\tau)|$ from above.

	Parallel to Remark \ref{notaccom}, similar comments are in order. Firstly, the result holds only for fixed $x_0$ due to a crucial scaling $\|\mathcal{A}(x_0x_0^*)\|_\infty = O(\sqrt{\log n})$. Besides, we recommend $\tau =\Theta  \big(\big[\frac{n}{d}\big]^{1/(2l)}\big)$ when $\frac{n}{d}$ is relatively large.

	 
	 \subsection{Noisy Rank-$r$ PSD Matrix Sensing}
\label{rank-r}	 
	 Due to the awareness of lack of theoretical guarantee for ERM estimation in NPR, NGPR, especially when $\mathbb{K}=\mathbb{C}$, the main aim of this paper is of course to establish new and sharper error bounds via proofs that are valid in both real and complex cases. Nevertheless, with Theorem \ref{theorem1}, Theorem \ref{theorem6}, Lemma \ref{lemma7} and Lemma \ref{lemma8} for general $r\in [d]$ in place, our proofs directly yield parallel results for more general low-rank matrix sensing. \textcolor{black}{We present this extension from $r = 1$ to $r\in[d]$ in the next two Theorems, and briefly give a sketch of the proof in Appendix \ref{appenc}. Details are left to avid readers.}
	 
	 	 \begin{theorem}
	     \label{theorem10}
	     Assume $X_0 = U_0U_0^*$ for some $U_0 \in \mathbb{C}^{d\times r}$ with known $r\in [d]$, $\{A_k=\alpha_k\alpha_k^*:k\in [n]\}$ are the rank-1 frame used in NPR. We consider the noisy  matrix sensing problem $y_k = \big<\alpha_k\alpha_k^*,U_0U_0^*\big>+\eta_k$ for $k\in [n]$, or a more compact form $Y = \mathcal{D}( U_0U_0^*)+\eta$. The ERM estimator $\widehat{X} = \widehat{U}\widehat{U}^*$ where $\widehat{U}$ is given by the first equation of (\ref{1.6}).
	     When $n = \Omega(rd)$ we have the following error bounds:

	     \noindent{{\rm (a)}} For arbitrary $\eta$, $\|\widehat{X}-X_0\|_F = O\big(\frac{\|\eta\|}{\sqrt{n}}\big)$ holds with probability at least $1-\exp(-\Omega(n))$.

	     \noindent{{\rm (b)}} For arbitrary $\eta$, $\|\widehat{X}-X_0\|_F = O\big(\|\eta\|_\infty\sqrt{\frac{rd}{n}}+\frac{\sqrt{r}|\mathbf{1}^\top\eta|}{n}\big)$ holds with probability at least $1-\exp(-\Omega(d))$.

	     \noindent{{\rm (c)}} If $\eta$ is zero-mean, sub-Gaussian with $\|\eta\|_{\psi_2} = O(1)$,  then it holds that $\|\widehat{X}-X_0\|_F = O\big(\sqrt{\frac{rd\log n}{n}}\big)$ with probability at least $1-\exp(-\Omega(d))-2n^{-49}$.

	      \noindent{{\rm (d)}} If $\eta_k$ are independent, sub-exponential with $\max_{k\in [n]}\|\eta_k\|_{\psi_1}=O(1)$, $\|\widehat{X}-X_0\|_F = O\big(\sqrt{\frac{rd(\log n)^2}{n}}\big)$ holds with probability at least $1-\exp(-\Omega(d))-2n^{-9}$.

	       \noindent{{\rm \textcolor{black}{(e)}}} If $\eta_k,~k\in[n]$ are i.i.d., symmetric, random noise satisfying $\mathbbm{E}|\eta_k|^l = O(1)$, for a specific $X_0 \in \mathbb{C}^{d\times d }$ with $\textcolor{black}{\|X_0\|_{nu} = O(1)}$, we choose $\tau  =C_1 \big(\frac{n}{d}\big)^{\frac{1}{2l}} \log n$ with sufficiently large $C_1$ and define the robust ERM estimator $\widehat{X}_{ht}=\widehat{U}_{ht}\widehat{U}_{ht}^*$ by substituting $y_k$ in original $\ell_2$ loss with $\widetilde{y}_k = \mathrm{sign}(y_k)\min\{|y_k|,\tau\}$. Then we have $\|\widehat{X}_{ht}-X_0\|_F = O\big(\big[\sqrt{\frac{rd}{n}}\big]^{1-\frac{1}{l}}r^{\frac{1}{2l}}(\log n)^2\big)$ with probability at least $1-2n^{-8}-C_2\exp(-\Omega(d))$.
	 \end{theorem}
	 \begin{theorem}
	     \label{theorem9}
	     Assume $X_0 = U_0U_0^*$ for some $U_0 \in \mathbb{C}^{d\times r}$ with known $r\in [d]$, $\{A_k:k\in [n]\}$ are the Hermitian frame used in NGPR. We consider the noisy matrix sensing problem $y_k = \big<A_k, U_0U_0^*\big>+\eta_k$ for $k\in [n]$, or a more compact form $Y = \mathcal{A}(U_0U_0^*)+\eta$. The ERM estimator $\widehat{X} = \widehat{U}\widehat{U}^*$ where $\widehat{U}$ is given by the second equation of (\ref{1.6}). When $n=\Omega(rd)$ we have the following error bound:
	     
	     \noindent{{\rm (a)}} For arbitrary $\eta$, $\|\widehat{X}-X_0\|_F = O\big(\|\eta\|\frac{\sqrt{rd}}{n}\big)$ holds with probability at least $1-5\exp(-\Omega(d))$.

	      \noindent{{\rm (b)}} If $\eta_k$ is sub-Gaussian and $\| \eta -\mathbbm{E}\eta\|_{\psi_2} = O(1)$, $\|\mathbbm{E}\eta\|_\infty= O(1)$, $\| \widehat{X}-X_0\|_F = O\big(\sqrt{\frac{rd}{n}}\big)$ holds with probability at least $1-7\exp(-\Omega(d))$.

	     \noindent{{\rm (c)}} If $\eta_k$ is sub-Gaussian and $\max_k\|\eta_k\|_{\psi_2} = O(1)$, $\|\widehat{X}-X_0\|_F = O\big(\sqrt{\frac{rd\log n}{n}}\big)$ holds with probability at least $1-5\exp(-\Omega(d))-2n^{-24}$.

	     \noindent{{\rm (d)}} If $\eta_k$ is sub-exponential and $\max_k\|\eta_k\|_{\psi_1}= O(1)$, $\|\widehat{X}-X_0\|_F = O\big(\sqrt{\frac{rd(\log n)^2}{n}}\big)$ holds with probability at least $1-5\exp(-\Omega(d))-2n^{-9}$.

	     \noindent{{\rm (e)}} If $\eta_k,~k\in [n]$ are i.i.d. random noise satisfying $\mathbbm{E}|\eta_k|^l = O(1)$, for a specific $X_0\in \mathbb{C}^{d\times d}$ with $\|X_0\|_F = O(1)$, we choose $\tau = C_1 \big(\frac{n}{rd}\big)^{\frac{1}{2l}}\sqrt{\log n}$ with sufficiently large $C_1$ and define the robust ERM estimator $\widehat{X}_{ht}=\widehat{U}_{ht}\widehat{U}_{ht}^*$ by substituting $y_k$ in original $\ell_2$ loss with $\widetilde{y}_k = \mathrm{sign}(y_k)\min\{|y_k|,\tau\}$. Then with probability at least $1-2n^{-24}-8\exp(-\Omega(d))$ we have $\|\widehat{X}_{ht}-X_0\|_F = O\big(\big[\sqrt{\frac{rd}{n}}\big]^{1-\frac{1}{l}}\log n\big)$.
	 \end{theorem}

      In Theorem \ref{theorem10}, \ref{theorem9}, the bounds in (a)-(d) simultaneously accommodate all rank-$r$ PSD matrix $X_0$, but the bound given in (e) is for a fixed $X_0$ only. Under heavy-tailed noise, in addition to the degradation of convergence rate, a new issue for the rank-$r$ matrix recovery via rank-1 frame is that the relative scaling between $n,d,r$ is $\frac{r^{\frac{l}{l-1}}d}{n}$ rather than the expected optimal one $\frac{rd}{n}$. Therefore, it needs further study for any possible improvement.

	\subsection{Rank-1 Frame v.s. Full-rank Frame}
	\label{comparison}
	
	To close this section, we compare the obtained error bounds  in NPR or NGPR. Note that this is in essence comparing the rank-1 frame $\{\alpha_k\alpha_k^*:k\in [n]\}$ and full-rank frame $\{A_k:k\in[n]\}$.

	Under arbitrary noise, NGPR possesses a bound $O\big(\|\eta\|\frac{\sqrt{d}}{n}\big)$. For NPR, we obtain two bounds $O\big(\frac{\|\eta\|}{\sqrt{n}}\big)$ and $O\big(\|\eta\|_\infty\sqrt{\frac{d}{n}}+\frac{|\mathbf{1}^\top\eta|}{n}\big)$, each of them can represent a tighter guarantee, see Remark \ref{remark613}. Mathematically, one can   verify 
	$$ \|\eta\|\frac{\sqrt{d}}{n} \leq \min\big\{ \frac{\|\eta\|}{\sqrt{n}},\|\eta\|_\infty \sqrt{\frac{d}{n}}+\frac{| \mathbf{1}^\top\eta|}{n}\big\}.$$
	Thus, the upper bound for NGPR error is at least as strong as that for NPR. Moreover, the NGPR bound becomes essentially tighter under $\eta$ that admits the scaling $\|\eta\| = \Theta(\sqrt{n})$, $\|\eta\|_\infty = \tilde{O}(1)$, $|\mathbf{1}^\top\eta| = \Omega(n)$, which are usually the case of biased noise such as $\eta = \mu_1 \mathbf{1}$, $\eta \sim \mathcal{N}(\mu_1,1)$ with $\mu_1 \neq 0$. More precisely, in this case, ERM in NGPR enjoys the guarantee $\tilde{O}\big(\sqrt{\frac{d}{n}}\big)$ that implies consistency, while we only have constant bound for NPR. Indeed, a lower bound in Theorem 1.2 in \cite{huang2020performance} confirms $\|\widehat{x}\widehat{x}^*-x_0x_0^*\|_F = \Theta(1)$ in NPR.

	The error bounds for sub-Gaussian random noise are immediate results from those for arbitrary noise. Therefore, although the bounds for both model are $\tilde{O}\big(\sqrt{\frac{d}{n}}\big)$, a significant difference is that in NPR $\eta_k$ is zero-mean, while in NGPR $\eta_k$ can has bounded deviation from $0$. Such difference also arises in results of heavy-tailed noise.

	Indeed, technically, this stems from the estimation of $\|\sum_{k=1}^n \eta_k A_k\|$ in NGPR, and $\| \sum_{k=1}^n $ $\frac{\eta_k}{\sqrt{n}}\alpha_k\alpha_k^*\|$ in NPR. More precisely,  $\|\sum_{k=1}^n \eta_k A_k\| = O(\sqrt{d})$ holds as long as $\|\eta\| = O(1)$. However, $\mathbbm{E}\eta_k = 0$ is essential for $\|\sum_{k=1}^n \frac{\eta_k}{\sqrt{n}}\alpha_k\alpha_k\|$ to have a scaling of $O(\sqrt{d})$. Otherwise, the upper bound would be significantly worsened. Here we give an intuitive interpretation: Unlike PSD $\alpha_k\alpha_k^*$ in NPR, the full-rank frame $A_k$ itself concentrates around zero, thereby removing noise biases automatically.

	Of course, using rank-1 frame significantly saves storage and computational cost, while our main results suggest that full-rank frame $\{A_k\}$ in NGPR is more robust to biased noise. Experimental results will be presented for clearer exposition.

	\section{Experimental Results}
	\label{section4}
	
	In this section, experimental results are presented to demonstrate and confirm 
	the obtained error bounds in Section \ref{section3}. We use Wirtinger flow to solve the ERM, and the details are given in Appendix \ref{append}.

	For a single simulation one has to set the the measurement ensemble, the underlying signal $x_0$, and the noise vector $\eta$. Unless otherwise specified, we would use complex Gaussian $\alpha_k$ (i.e., $\mathcal{M} = \mathcal{N}(0,\frac{1}{2})$ in Assumption \ref{assumption2}) for NPR, and $A_k$ with independent complex Gaussian entries (i.e., $\breve{A}_0$ with i.i.d. $\mathcal{N}(0,1)$ entries in Assumption \ref{assumption1}) for NGPR.

	 To ease our presentation, we specify several canonical settings of signal and noise. We let $x_\mathbf{1}[d]$ be the (normalized) $d$-dimensional signal with entries $\frac{1}{\sqrt{2d}}(1+\ii)$, i.e., $x_\mathbf{1}[d] = \frac{1}{\sqrt{2d}}(1+\ii)\mathbf{1}$. We would also use the random signal $x_{\mathcal{U}}[d]$ where the corresponding real vector $[\mathrm{Re}(x_{\mathcal{U}}[d]),$ $\mathrm{Im}(x_{\mathcal{U}}[d])]$ are uniformly distributed over the unit Euclidean sphere of $\mathbb{R}^{2d}$.

	For deterministic $\eta$, given a specific sample size $n$ we introduce the notation $ \mathbf{1} (s,\rho)$ to represent the $n$-dimensional $\eta$ with support set $[s]$, and entries in $[\rho s]$ are $1$, the entries in $[s] \setminus [\rho s] $ are $-1$. For instance, $ \mathbf{1}(n,1) = \mathbf{1}$, $ \mathbf{1}(n,0) =- \mathbf{1}$, $ \mathbf{1}(3,\frac{2}{3}) = (1,1,-1,0,...,0)^\top$.

	For random noise we first introduce (multivariate) Gaussian distribution, Laplace distribution, and Student's t-distribution, which would be used to mimic sub-Gaussian noise, sub-exponential noise, and heavy-tailed noise, respectively. We let $\mathcal{N}_n(\mu,\Sigma)$ be the $n$-dimensional Gaussian distribution with mean $\mu$, covariance matrix $\Sigma$. Besides, we use $Lap(\mu)$ to denote the Laplace distribution with mean $\mu$ and unit standard deviation. Moreover, $t(\nu)$ denotes Student's t-distribution with degrees of freedom $\nu$, while when $\nu>2$ we let $t^s(\nu) := \sqrt{1-\frac{2}{\nu}}t(\nu)$ be the rescaled distribution with unit variance. For any univariate distribution $\digamma$, $\eta_k\sim \digamma$ means entries of $\eta$ are independent copies of $\digamma$.

	Each data point in the figures is obtained from the average of 100 trials.
	\subsection{Deterministic Noise}
	
	The main aim of this part is to use the deterministic noise $\mathbf{1}(s,\rho)$ to demonstrate the obtained error bounds for arbitrary noise.

	For NPR, there exists only one bound $\mathcal{B}_1:=O\big(\frac{\|\eta\|}{\sqrt{n}}\big)$ in literature, while the new bound $\mathcal{B}_2 := O\big(\|\eta\|_\infty\sqrt{\frac{d}{n}}+\frac{|\mathbf{1}^\top \eta|}{n}\big)$ is one of our main contributions. In Remark \ref{remark613}, we point out that they are complementary to each other, and which bound is tighter depends on the spikiness of $\eta$.

	We first test NPR under the setting of the triple $(n,x_0,\eta)$ given by \begin{equation}
	\label{exp1}
	 \big(n, x_\mathbf{1}[\frac{n}{15}], \phi \cdot \mathbf{1}(10,1)\big),~\begin{cases}
	n \in 15\times  \{10,15,20,30,40,50,60,70\} \\
	\phi \in \{3.5,4.0\}
	\end{cases} .\end{equation}
    Under a fixed $\phi$, we can see that $\eta$ is a $10$-sparse noise with $\|\eta\|_\infty, \|\eta\|$ being constant, which gives a high spikiness $\alpha^*(\eta) = \Theta(\sqrt{n})$. Thus, $\mathcal{B}_1$ should be tighter for (\ref{exp1}). Indeed, by noting $\frac{n}{d} = 15$ in (\ref{exp1}), the bounds admit more explicit form as $\mathcal{B}_1 = O\big(\frac{1}{\sqrt{n}}\big)$, $\mathcal{B}_2 = O(1)$.

    The setting of our second simulation are specified by $(n,x_0,\eta)$ as
    \begin{equation}
	\label{exp2}
	 \big(n, x_\mathbf{\mathcal{U}}[30], \phi \cdot \mathbf{1}(n,0.5)\big),~\begin{cases}
	n \in 30\times  \{10,15,20,30,40,50,60,70\} \\
	\phi \in \{0.2,0.3,0.4\}
	\end{cases} .\end{equation}
	Here, the underlying signal with $d=30$ is fixed, and note that $\mathbf{1}(n,0.5)$ is just the all-ones vector with  half of the entries flipped to $-1$. Evidently, for each $\phi$ the noise vector achieves the lowest spikiness $\alpha^*(\eta) = 1$. And one can easily verify that $\mathcal{B}_1 = O(1)$, $\mathcal{B}_2 = O\big(\frac{1}{\sqrt{n}}\big)$, thus the roles of two bounds are swapped compared with   (\ref{exp1}).

\begin{figure}[ht]
    \centering
    \includegraphics[scale = 0.67]{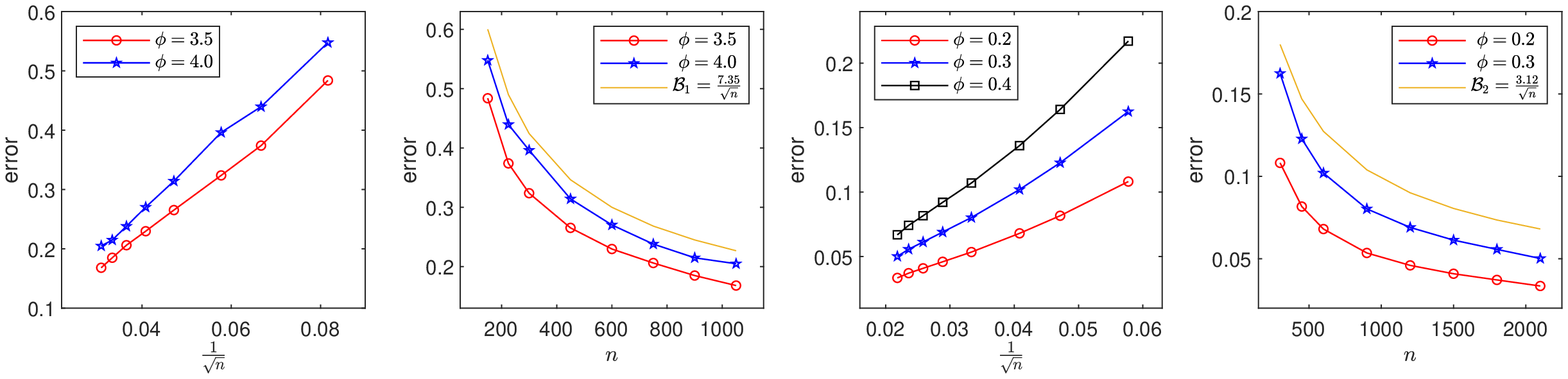}

~~~~~(a) \hspace{3.5cm} (b) \hspace{3.5cm} (c) \hspace{3.5cm} (d)

    \caption{
    {NPR}: (a) and (b) for (\ref{exp1}); (c) and (d) for (\ref{exp2}).}
    \label{nfig1}
\end{figure}

	The results of (\ref{exp1}), (\ref{exp2}) are respectively shown in Figures \ref{nfig1} (a)-(b) and Figures \ref{nfig1} (c)-(d).
	Given a specific noise magnitude $\phi$, in both simulations the curves of ``error v.s. $\frac{1}{\sqrt{n}}$'' are shaped like straight lines, thus confirming the statistical rate $O\big(\frac{1}{\sqrt{n}}\big)$. The curves of ``error v.s. $n$'' are also provided to display how the reconstruction error decreases under a bigger sample size. Moreover, we plot the theoretical curves $\mathcal{B}_1=\frac{7.35}{\sqrt{n}}$ for (\ref{exp1}), $\mathcal{B}_2=\frac{3.12}{\sqrt{n}}$ for (\ref{exp2}) to represent the tighter bound in each case, and one may see they are fairly parallel to the experimental curves. When $\phi$ varies, we have checked that the error are well proportional to the noise magnitude $\phi$. For instance, in  Figure \ref{nfig1} (d), the errors in the blue curve are about $\frac{3}{2}$ times as large as those shown in the red curve. This is consistent with our new bound $O\big(\| \eta\|_\infty\sqrt{\frac{d}{n}}\big)$.

\begin{figure}[ht]
    \centering
    \includegraphics[scale = 0.66]{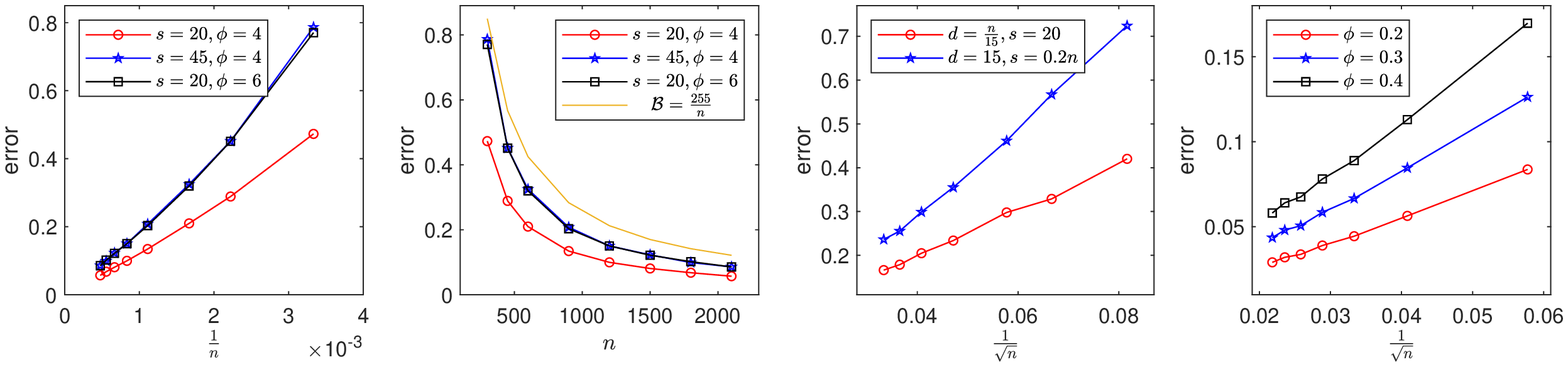}
    
 ~~~   (a) \hspace{3.5cm} (b) \hspace{3.5cm} (c) \hspace{3.5cm} (d)
    
    \caption{
    {NGPR}: (a) and (b) for (\ref{exp3}); (c) for (\ref{exp4}); (d) for 
    (\ref{exp5}).}
    \label{nfig2}
\end{figure}

		For NGPR, we similarly verify the general error bound $\mathcal{B}=O\big(\|\eta\|\frac{\sqrt{d}}{n}\big)$ under deterministic noise. We first test NGPR under sparse noise, specifically the two settings
		\begin{equation}
		    \label{exp3}
		    \big(n, x_\mathbf{\mathcal{U}}[30], \phi \cdot \mathbf{1}(s,1)\big),~\begin{cases}
	n \in 30\times  \{10,15,20,30,40,50,60,70\} \\
	(s,\phi) \in \{(20,4),(20,6),(45,4)\}
	\end{cases} 
		\end{equation}
	where  a fixed number of measurements are corrupted, and 
	\begin{equation}
		    \label{exp4}
		    \big(n, x_\mathbf{\mathbf{1}}[d], 4\cdot \mathbf{1}(s,1)\big),~\begin{cases}
	n \in 15 \times  \{10,15,20,30,40,50,60 \} \\
	(d,s) \in \{(\frac{n}{15},20),(15,0.2n)\}
	\end{cases} 
		\end{equation}
	where $(d,s) = \big(\frac{n}{15},20\big)$ means the problem dimension is proportional to $n$, $(d,s)= (15,0.2n)$ means a fixed percentage of measurements are corrupted. Besides, we also try a setting with all-ones noise give by
		\begin{equation}
		    \label{exp5}
		    \big(n,x_{\mathcal{U}}[30],\phi\cdot \mathbf{1}(n,1)\big),~n\in 30\times \{10,20,30,40,50,60,70\},~\phi \in (0.2,0.3,0.4).
		\end{equation}
		Evidently, for $d$-dimensional $x_0$ and $\eta = \phi\cdot \mathbf{1}(s,1)$, the error bound reads as $\mathcal{B} = O\big(\phi\frac{\sqrt{ds}}{n}\big)$. Thus, the obtained theoretical guarantee is $O\big(\frac{1}{n}\big)$ for (\ref{exp3}), and $O\big(\frac{1}{\sqrt{n}}\big)$ for (\ref{exp4}) and (\ref{exp5}). 
Figures \ref{nfig2} (a)-(b) display the results of (\ref{exp3}), and 
Figures \ref{nfig2} (c)-(d) show those of (\ref{exp4}), (\ref{exp5}). 
For the horizontal coordinate, we use $\frac{1}{n}$ in 
Figure \ref{nfig2} (a),
while we use $\frac{1}{\sqrt{n}}$ in 
Figure \ref{nfig2} (c)-(d).
Consistent with the implications of the error bound $\mathcal{B}$, the experimental data nearly display straight lines in these figures. Notably, in the first and second figure the curves of "$s=45,\phi=4$", "$s=20,\phi=6$" almost coincide, so the relation reflected in the bound that the error is proportional to $\|\eta\|$ is quite precise. In addition, the second figure plots the curves of (\ref{exp3}) with $n$ as the horizontal coordinate. A theoretical bound $\mathcal{B}=\frac{255}{n}$ is also provided for comparison.

	\subsection{Light-tailed Random Noise}
	
	 In this part, we will test the performance of ERM in NPR, NGPR under light-tailed random noise, with the purpose to demonstrate the guarantees presented in Theorem \ref{theorem7}, \ref{theorem33}.

	 Unlike existing works that impose independent assumption on $\eta_k$ (e.g., \cite{eldar2014phase,lecue2015minimax}), (a) of Theorem \ref{theorem7} removes this condition. To verify this, we first test complex NPR under the setting   
	 	\begin{equation}
		    \label{exp6}
		    \big(n, x_\mathbf{\mathcal{U}}[30], \eta\big),~\begin{cases}
	n \in 30 \times  \{10,15,20,30,40,50,60,70,80\} \\
	\eta \sim 0.2 \cdot \mathcal{N}_n(\mathbf{0},I_n)~~\mathrm{and}~~\eta \sim 0.2\cdot \mathcal{N}_n(\mathbf{0},\Sigma_n) 
	\end{cases} ,
		\end{equation}
	 where $\Sigma_n$ is a positive definite matrix with $\mathbf{1}$ as diagonal\footnote{Given $A\in\mathbb{C}^{n\times n}$, we let $\mathbf{D}(A)$ be the diagonal matrix sharing the same diagonal as $A$. Here, $\Sigma_n$ is randomly generated by $\big(\mathbf{D}(RR^\top)\big)^{-1/2}RR^\top\big(\mathbf{D}(RR^\top)\big)^{-1/2}$, where $R\in\mathbb{R}^{n\times n}$ has i.i.d., $\mathcal{N}(0,1)$ entries.}. Note that we still require $\mathbbm{E}\eta_k = 0$. To see this is necessary, we further test the following NPR setting 
	 \begin{equation}
	     \label{exp7}
	     \big(n, x_\mathbf{\mathcal{U}}[30], \eta\big),~n \in 30\cdot  20 \cdot [11]; ~\eta \sim \frac{\sqrt{2}}{5}\cdot\mathcal{N}_n(\mathbf{0},I_n)~\mathrm{and}~\eta \sim 0.2 \cdot \mathcal{N}_n(\mathbf{1},I_n).
	 \end{equation}
	 In (\ref{exp7}) the two choices of $\eta$ represent zero-mean noise and biased noise, while they are comparable in magnitude because of the same second moment.

\begin{figure}[ht]
    \centering
    \includegraphics[scale = 0.57]{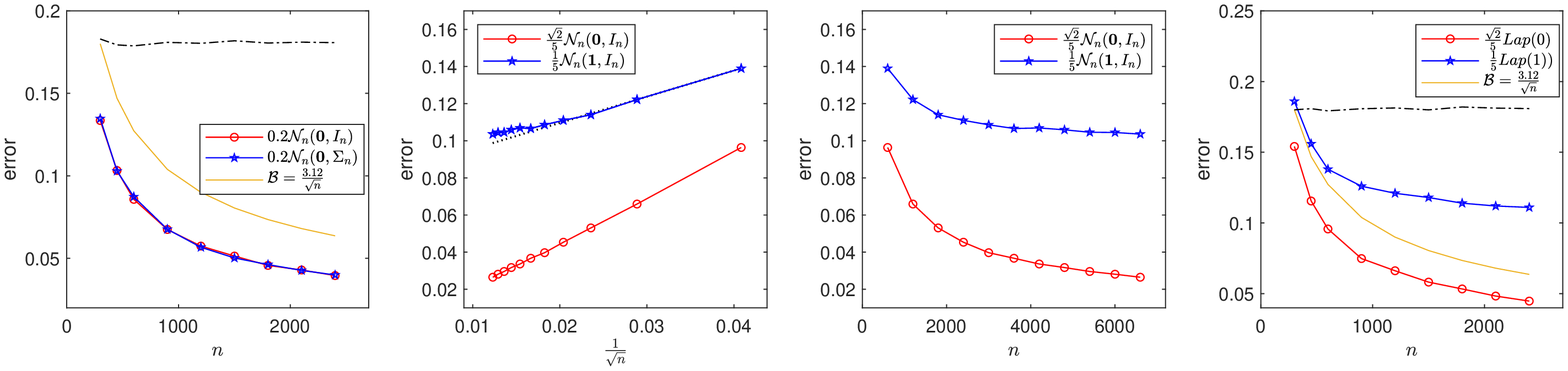}
    
~~~~~  (a) \hspace{3.4cm} (b) \hspace{3.4cm} (c) \hspace{3.4cm} (d)
    
    \caption{
    {NPR}: (a) for  
    (\ref{exp6}); (b)-(c) for (\ref{exp7}); (d) for (\ref{exp6}) under Laplace noise.}
    \label{nfig3}
\end{figure}

	 Figure \ref{nfig3} (a) displays the results of (\ref{exp6}). Clearly, the curves of independent noise (red) and correlated noise (blue) are almost coincident, thus confirming the independent assumption is inessential. Results of (\ref{exp7}) are shown in Figure \ref{nfig3} (b)-(c), with the horizontal coordinate being $\frac{1}{\sqrt{n}}$ or $n$. 
	 In Figure \ref{nfig3} (b), 
	 the data points obtained under the zero-mean noise nearly fall on a straight line. This verifies the statistical rate $\tilde{O}\big(\frac{1}{\sqrt{n}}\big)$. By contrast, the points on the blue curve near the left deviate from the direction of those near the right (compare with the black dotted line to see), thus the rate becomes slower than $O(n^{-1/2})$ under large $n$. Besides, one can observe that the "slope" of the blue curve are smaller than that of the red line, which implicitly implies that blue curve does not converge to $0$ even if $n\to \infty$. This is also clear in 
	 Figure \ref{nfig3} (c). Therefore, the zero-mean assumption is essential for the bound $\tilde{O}\big(\sqrt{\frac{d}{n}}\big)$ to hold. Indeed, ERM fails to achieve consistency under biased Gaussian noise has been proved in Theorem 1.2, \cite{huang2020performance}.

	 We also test (\ref{exp6}) with $\eta$ changed to be zero-mean Laplace noise $\eta_k \sim \frac{\sqrt{2}}{5}\cdot Lap(0)$ and biased Laplace noise $\eta_k\sim 0.2\cdot Lap(1)$ that have similar magnitude. The results shown in 
	 Figure \ref{nfig3} (d), particularly the red and blue experimental curves, are similar to
	 Figure \ref{nfig3} (c). 
	 Besides, in Figures \ref{nfig3} (a) and (d), 
	 we provide the theoretical curve $\mathcal{B} = \tilde{O}\big(\sqrt{\frac{d}{n}}\big) = \frac{3.12}{\sqrt{n}}$, which matches the red experimental curves pretty well. On the other hand, we also experimentally track the only previously known bound $O\big({\frac{\|\eta\|}{\sqrt{n}}}\big)$ and plot it as the black dash-dotted line. One shall evidently see it is a loose upper bound and far from the ERM performance.

\begin{figure}[ht]
    \centering
    \includegraphics[scale = 0.62]{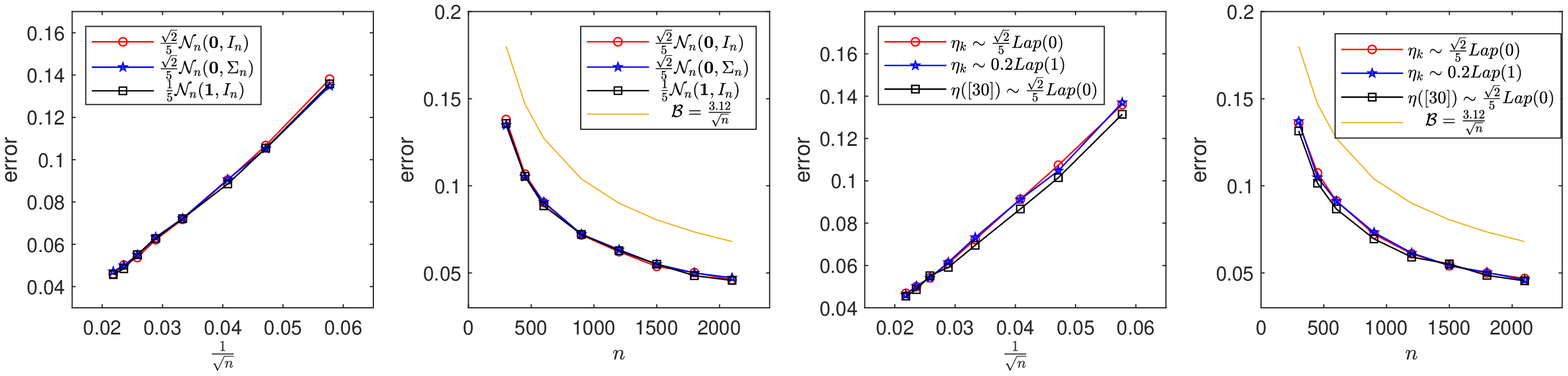}
    
    ~~~~~  (a) \hspace{3.4cm} (b) \hspace{3.4cm} (c) \hspace{3.4cm} (d)
    
    \caption{
    {NGPR}: (a) and (b)
    for (\ref{exp8}); (c) and (d) for (\ref{exp8})  
    under Laplace noise.}
    \label{nfig4}
\end{figure}

	  For NGPR, we similarly report numerical examples to demonstrate Theorem \ref{theorem33} from the perspective of the rate $\tilde{O}\big(\frac{1}{\sqrt{n}}\big)$, the dispensable independent or zero-mean assumption. We conduct NGPR under the setting 
	  \begin{equation}
		    \label{exp8}
		    \big(n, x_\mathbf{\mathcal{U}}[30], \eta\big),~\begin{cases}
	n \in 30 \times  \{10,15,20,30,40,50,60,70,80\} \\
	\eta \sim \big\{\frac{\sqrt{2}}{5} \cdot \mathcal{N}_n(\mathbf{0},I_n),\frac{\sqrt{2}}{5}\cdot \mathcal{N}_n(\mathbf{0},\Sigma_n), 0.2\cdot \mathcal{N}_n(\mathbf{1},I_n)  \big\} 
	\end{cases} ,
		\end{equation}
	  where three choices of $\eta$ with comparable noise magnitude represent zero-mean and independent, zero-mean and correlated, biased and independent Gaussian noise, respectively. To see the situations under sub-exponential noise, we merely change the choices of $\eta$ in (\ref{exp8}) to be 
	  \begin{equation}
	      \label{exp9}
	        \big\{ \eta_k \sim \frac{\sqrt{2}}{5}\cdot Lap(0),\eta_k\sim 0.2\cdot Lap(1), \eta([30])\sim  \frac{\sqrt{2}}{5}\cdot Lap(0) \big\}.
	  \end{equation}
    Here, the last choice means $\{\eta_k:k\in[30]\}$ are i.i.d. copies of $\frac{\sqrt{2}}{5}\cdot Lap(0)$, while we let $\eta_k = \eta_{l}$ with $l = k\mod 30 \in [30]$. This represents sub-exponential $\eta$ with correlated entries. In all these settings, theoretically, ERM in NGPR at least enjoys the guarantee $\tilde{O}\big(\sqrt{\frac{d}{n}}\big)= \tilde{O}\big(\frac{1}{\sqrt{n}}\big)$. And one shall see the experimental results shown in Figures \ref{nfig4} (a)-(d) are consistent with the theory. Notably, in NGPR, the experimental curves corresponding to zero-mean noise, biased noise almost coincide. This is in stark contrast to the situation in NPR, specifically 
Figures \ref{nfig3} (c)-(d) where the blue curve is significantly higher and flatter than the red one, thereby illustrating our conclusion that NGPR is more robust to biased noise. In other words, noise biases could not ruin the ERM estimation if the full-rank frame is used. 
    
    \subsection{Heavy-tailed Random Noise}
    
    A novelty of this work is to first consider the the ERM estimator for (generalized) phase retrieval problem where noise is only assumed to have bounded $l$-th moment (Thus can be heavy-tailed). Our proposal to enhance robustness is an additional truncation step applied to the responses, and the resulting ERM estimators are shown to possess a worst-case error bound $\tilde{O}\big(\big[\sqrt{\frac{d}{n}}\big]^{1-1/l}\big)$. In this part, numerical results will be presented to verify the truncation is not merely a technical trick to establish the bound, but indeed leads to significantly better estimator. We will also illustrate the truncation step is in essence a bias-variance trade-off.

    Recall that $t^s(\nu)$ denotes Student's t-distribution that is rescaled to have unit variance. We would i.i.d. draw heavy-tailed noise from $t^s(2.5) = \frac{1}{\sqrt{5}}t(2.5)$, which has bounded $l$-th moment if $l<2.5$. For illustrative purpose, we would use $l = 2.45$. Moreover, we further omit the logarithmic factors in the $\tau$ suggested in Theorem \ref{theoremht}, \ref{theorem5} (due to reasons stated in Remark \ref{notaccom}) and tune the truncation parameter as $\tau = c_0 \big(\frac{n}{d}\big)^{1/4.9}$ for some constant $c_0$. To be specific, we implement NPR under the setting
     \begin{equation}
		    \label{exp10}
		    \big(n, x_\mathbf{\mathcal{U}}[30], \eta,\tau\big),~\begin{cases}
	n \in 30 \times  \{10,15,20,25,30,35,40,45,50,60,70,80,90,100\} \\
\eta_k \sim 0.75\cdot t^s(2.5 );~\tau = 2.8 \big(\frac{n}{d}\big)^{1/4.9}~~\mathrm{or}~~ \tau = \infty 
	\end{cases} .
		\end{equation}
    Note that "$\tau = \infty$" corresponds to the original ERM estimator. Then, we do simulations for NGPR under exactly the same setting as (\ref{exp10}), but instead use $\tau = 2.4\big(\frac{n}{d}\big)^{1/4.9}$ for truncation.

\begin{figure}[ht]
    \centering
    \includegraphics[scale = 0.72]{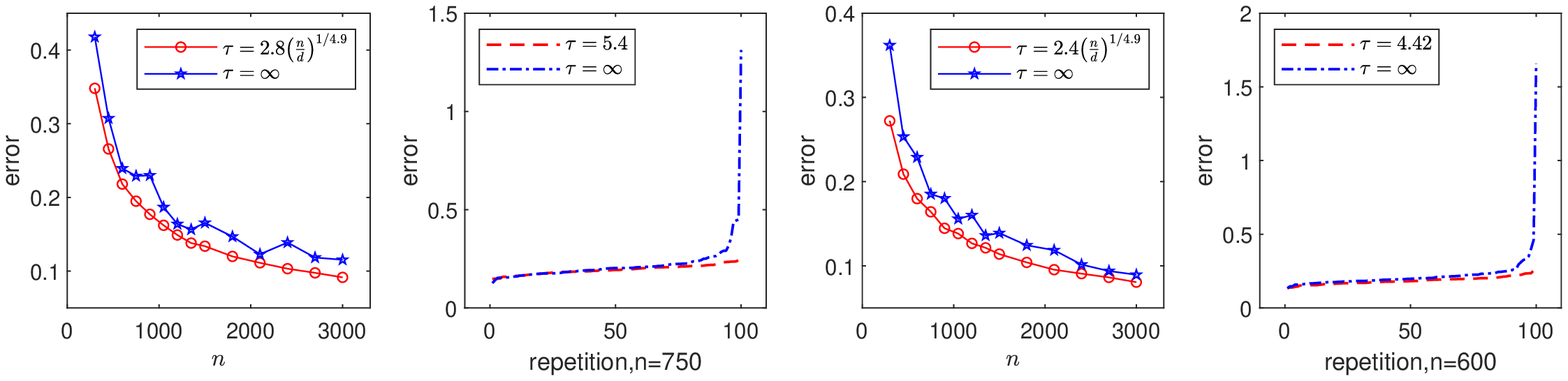}

 ~~~~~  (a) \hspace{3.4cm} (b) \hspace{3.4cm} (c) \hspace{3.4cm} (d)

    \caption{
    (a)-(b) for NPR under heavy-tailed noise; (c)-(d) for NGPR under heavy-tailed noise.}
    \label{nfig5}
\end{figure}

    The experimental results are shown in Figures \ref{nfig5} (a)-(b) for 
    NPR, and Figures \ref{nfig5} (c)-(d)
   for NGPR. One shall see the curves of original ERM (without truncation) is not smooth, but instead display some fluctuations. In contrast, the curves of robust ERM estimation   are monotonically decreasing and embrace obviously smaller recovery error.

    Recall that each data point is obtained as mean value of 100 trials, let us also inspect the errors of each single simulation for some understandings. Specifically, for NPR under $n= 750$, or NGPR under $n=600$, we plot the errors in 100 repetitions, which is rearranged to render monotonically increasing curves. As it turns out, under good circumstances (e.g., $\eta$ does not contain any outliers) the original ERM is comparable to the proposed robust estimator. However, without truncation ERM suffers from large variance, and the worst cases can lead to ridiculous estimations, for instance, the largest errors in Figure \ref{nfig5} (b) and (d) exceed the desired signal norm $1$. Evidently, the truncation effectively avoid these unacceptable results. The insight is that, the truncation introduces some biases to render variance reduction, thereby achieving a bias-variance trade-off.

    \subsection{Other Measurement Ensembles}
    
    While previous experiments are conducted under complex Gaussian $\alpha_k$ and $A_k$, our results apply to some other measurement ensembles. To be specific, their entries can be drawn from rather general sub-Gaussian distributions, and in NGPR entries of $A_k$ can even be correlated (see Assumption \ref{assumption2}, \ref{assumption1}). In this part, we test two other measurements.

\begin{figure}[ht]
    \centering
    \includegraphics[scale = 0.69]{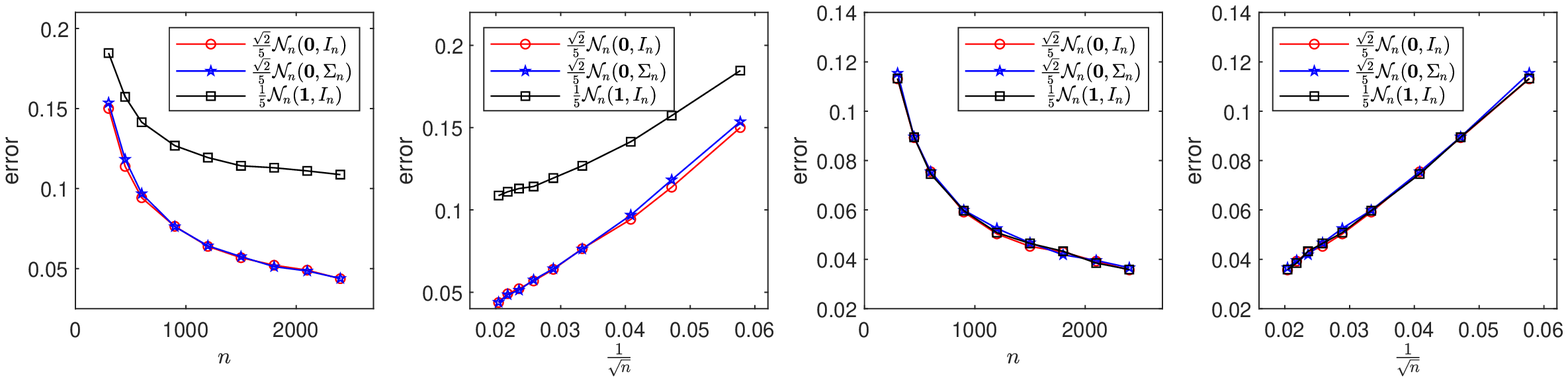}
    
    ~~~~~  (a) \hspace{3.4cm} (b) \hspace{3.4cm} (c) \hspace{3.4cm} (d)
    
    \caption{
(a)-(b) for NPR with $\mathcal{M}=\frac{16}{17}\cdot\mathcal{N}(0,0.5)+\frac{1}{17}\cdot\big(3\varepsilon\big)$; (c)-(d) for NGPR with $\mathrm{vec}(\breve{A}_0)\sim \mathcal{N}_{d^2}(\mathbf{0},\Sigma_{d^2})$.}
    \label{nfig6}
\end{figure}

    We implement NPR, NGPR under the setting given by (\ref{exp8}). In NPR, we use the measurement ensembles with $\mathcal{M}$ in Assumption \ref{assumption2} being a mixture of two distributions, i.e., $\mathcal{N}(0,0.5)$ with weight $\frac{16}{17}$, $3\cdot \varepsilon$ with weight $\frac{1}{17}$ (Here $\varepsilon$ denotes Rademacher variable). This $\mathcal{M}$ satisfies our fourth moment condition, but the resulting random vector violates the small-ball probability assumption required in some early works (e.g., \cite{eldar2014phase}). In NGPR, 
    we draw $\breve{A}_0$ in Assumption \ref{assumption1} by $\mathrm{vec}(\breve{A}_0)\sim \mathcal{N}_{d^2}(\mathbf{0},\Sigma_{d^2})$, where $\Sigma_{d^2}$ is no longer the previously used $I_{d^2}$ but drawn by a random mechanism: We first draw $R_0$ as a $d^2 \times d^2$ random matrix with i.i.d. standard Gaussian entries, then use the Matlab function "orth(.)" to find the corresponding orthogonal matrix $R_1$, finally we let $\Sigma_{d^2} = R_1^\top R_2R_1$ where $R_2$ is a diagonal matrix with i.i.d. entries   uniformly distributed over $[\lambda_0,\lambda_1]= [0.4,2.5] $. Experimental results with similar implications are observed, see Figure \ref{nfig6}.

	\section{Concluding Remarks}
	\label{section5}
	Minimizing an empirical $\ell_2$ loss is the goal of many existing non-convex algorithms in (generalized) phase retrieval or low-rank matrix recovery, to name a few, wirtinger flow or its variants \cite{candes2015phase,kolte2016phase,huang2020solving}, procrustes flow \cite{tu2016low}. There exist several works studying stability guarantee of ERM in NPR \cite{soltanolkotabi2014algorithms,huang2020performance,eldar2014phase,lecue2015minimax}. Nevertheless, the guarantee $O\big(\frac{\|\eta\|}{\sqrt{n}}\big)$ in \cite{soltanolkotabi2014algorithms,huang2020performance} is not tight for some noise patterns, while $O\big(\sqrt{\frac{d\log n}{n}}\big)$ in \cite{lecue2015minimax} is restricted to the real case and independent Gaussian noise. For ERM in NGPR, a recent preprint \cite{fan2022oracle} showed $O\big(\sqrt{\frac{d\log n}{n}}\big)$, but also, for the real case and independent sub-Gaussian noise only. Therefore, ERM performance in a noisy setting of these signal recovery models, a more theoretical problem, has not yet been well studied, especially in the complex case $\mathbb{K}=\mathbb{C}$. For instance, in complex NPR under i.i.d. standard Gaussian noise,   existing results cannot tell whether ERM achieves consistency in the estimation. The lack of theoretical understanding of ERM motivates this work.

    To close the gap, we establish new error bounds that are tighter or more general. The proofs are presented in complex case but can be easily adapted to real case. In addition, all our theoretical bounds are extended to the problem of noisy low-rank matrix recovery.

    For NPR under arbitrary noise, we establish a new bound $O\big(\|\eta\|_\infty\sqrt{\frac{d}{n}}+\frac{| \mathbf{1}^\top\eta|}{n}\big)$ to complement the known one $O\big(\frac{\|\eta\|}{\sqrt{n}}\big)$. Both bounds may represent a stronger guarantee. In particular, the known one is sharper for highly "spiky" noise like sparse noise, while the new bound is tighter for noise with low spikiness, which is often the case of i.i.d. random noise. In fact, this new bound immediately gives $\tilde{O}\big(\sqrt{\frac{d}{n}}\big)$ under sub-Gaussian noise, thus the missing complex counterpart of \cite{lecue2015minimax,eldar2014phase} is found. For NGPR under arbitrary $\eta$, we obtain the error bound $O\big(\|\eta\|\frac{\sqrt{d}}{n}\big)$. This is experimentally confirmed, for instance, the rate in (\ref{exp3}) is of $O\big(\frac{1}{n}\big)$ when $\|\eta\|,d= O(1)$, while the rate precisely shrinks to $O\big(\frac{1}{\sqrt{n}}\big)$ in (\ref{exp4}) when either $d = \Theta(n)$ or $\|\eta\| = \Theta(\sqrt{n})$ occurs (Figure \ref{nfig2}). With this general bound in place, we directly show $\tilde{O}\big(\sqrt{\frac{d}{n}}\big)$ for sub-Gaussian noise. Unlike most existing works assuming independent, zero-mean, sub-Gaussian noise to capture moderate random noise, our result $\tilde{O}\big(\sqrt{\frac{d}{n}}\big)$ are obtained under weaker assumption. To be specific, $\{\eta_k\}$ can be sub-exponential and correlated in both NPR, NGPR, and $\eta_k$ can be biased in NGPR. Although the rank-1 frame used in NPR is cheaper in storage and computation, our results indicate that the full-rank frame used in NGPR would be essentially more robust to biased noise, \textcolor{black}{one may compare Figure \ref{nfig3} (c) and Figure \ref{nfig4} (b) to see}.

	For heavy-tailed random noise with bounded $l$-th moment, the original ERM estimator is not robust due to the possibility that the $\ell_2$ loss is dominated by several response outliers. To resolve the issue, we propose a robust ERM estimator that minimizes the robust loss constructed from the truncated responses. The robust ERM estimator is shown to possess the error bound $\tilde{O}\big(\big[\sqrt{\frac{d}{n}}\big]^{1-\frac{1}{l}}\big)$ in both NPR, NGPR. Numerical examples are given to corroborate that the proposed robust ERM estimator outperforms the original ERM, and the additional truncation step achieves a bias-variance trade-off, see Figure \ref{nfig5}. For phase retrieval under heavy-tailed noise, our work seems to be a first study and poses several open problems, for example, whether faster rate is achievable, or a uniform recovery result can be established. We leave it further investigation for any improvement of the current results.

	\bibliographystyle{plain}
	\bibliography{library}
	
	\appendix
\section{Mendelson's Small Ball Method}
\label{appendixa}
    Lemma \ref{lemma1} gives Mendelson's small ball method, an effective tool developed in \cite{mendelson2014learning,mendelson2018learning} to lower bound a non-negative empirical process. Here, we adopt the version given in Lemma 1, \cite{dirksen2016gap} with $p=1$.
     \begin{lem}
    \label{lemma1} Fix a set $E\in \mathbb{C}^d$, let $\varphi$ be a random vector in $\mathbb{C}^d$, and let $\varphi_1,...,\varphi_n$ be independent copies of $\varphi$. Define $ Q_{\xi}(E;\varphi) := \inf_{u \in E} ~\mathbbm{P}\big(| \big<\varphi, u\big>| \geq \xi\big).$
    Let $\varepsilon_1,...,\varepsilon_n$ be independent Rademacher random variables (i.e., $\mathbbm{P}(\varepsilon_i=1) = \mathbbm{P}(\varepsilon_i = -1)=1/2$) independent of everything else, and define the mean empirical width of the set $E$ as 
   $W_n(E;\varphi) := \mathbbm{E}~\sup_{u\in E}\big<h,u\big>$, where $h:=\frac{1}{\sqrt{n}}\sum_{i=1}^n \varepsilon_i \varphi_i$.
    Then, for any $\xi>0$, $t>0$, with probability at least $1-\exp(-\frac{t^2}{2})$ we have
    \begin{equation}
        \nonumber
        \inf_{u\in E}  \frac{1}{\sqrt{n}}\sum_{i=1}^n |\big<\varphi_i,u\big>|   \geq \xi \sqrt{n}Q_{2\xi}(E;\varphi) -2W_n(E;\varphi) - \xi t.
    \end{equation}
    \end{lem}
     
     We apply the framework of Mendelson's small ball method in NPR, and evidently, a lower bound for $Q_{2\xi}(E;\varphi)$ would be part of the work. For this purpose, Paley-Zygmund inequality presented in Lemma \ref{lemma2} is the standard tool, while we still need to deal with $"Z^2"$, which turns out to be power of a quadratic chaos, by the Hanson-Wright inequality in Lemma \ref{lemma3}.

    \begin{lem}
    \label{lemma2}
    {\rm(Lemma 7.16 in \cite{foucart2013invitation})} If a nonnegative random variable $Z$ has finite second moment, then $\mathbbm{P}(Z>t) \geq {(\mathbbm{E}Z-t)^2}/{\mathbbm{E}(Z^2)}$ for any $0\leq t \leq \mathbbm{E}Z$.
    \end{lem}
    
    	\begin{lem}
	\label{lemma3}
	{\rm(Theorem 6.2.1 in \cite{vershynin2018high})} Let $X = (X_1,...,X_N)\in\mathbb{R}^N$ be a random vector with independent, zero-mean, sub-Gaussian entries, and $A\in\mathbb{R}^{N\times N}$. We define $K = \max_i \|X_i\|_{\psi_2}$, then there exists a constant $C$ such that for any $t>0$, we have 
	\begin{equation}
	    \nonumber
	    \mathbbm{P}\left(|X^{\top}AX-\mathbbm{E}X^{\top}AX|\geq t\right) \leq 2\exp \Big[-C\min\big\{\frac{t^2}{K^4 \|A\|_F^2},\frac{t}{K^2 \|A\|}\big\}\Big].
	\end{equation}
 
	\end{lem}
	
	\section{Technical Lemmas}

     \label{appendixb}
 
 In this work, we frequently apply the following two basic concentration inequalities, namely Berninstein's inequality and Hoeffding's inequality. 
 
 \begin{lem}
 \label{bern}
 {\rm (Theorem 2.8.1 in \cite{vershynin2018high})} Let $X_1,...,X_N$ be independent, sub-exponential random variable. Then for any $t>0$, for some constant $C$ we have \begin{equation}
     \nonumber
     \mathbbm{P}\Big(| \sum_{i=1}^N (X_i-\mathbbm{E} X_i)|\geq t\Big)\leq 2\exp\Big(-C\min\big\{\frac{t^2}{\sum_{i=1}^N \| X_i\|^2_{\psi_1}},\frac{t}{\max_{i\in [N]}\| X_i\|_{\psi_1}} \big\}\Big)
 \end{equation}
 \end{lem}
 
 \begin{lem}
 \label{hoef}
 {\rm (Theorem 2.2.6 in \cite{vershynin2018high})} Let $X_1,...,X_N$ be independent, bounded, random variables satisfying $X_i \in [m_i,M_i]$ for each $i$. Then for any $t>0$, we have \begin{equation}
     \nonumber
     \mathbbm{P}\Big(| \sum_{i=1}^N (X_i-\mathbbm{E} X_i)|\geq t\Big) \leq 2\exp\Big(-\frac{2t^2}{\sum_{i=1}^N (M_i-m_i)^2}\Big).
 \end{equation}
 \end{lem}
 
In the following we derive the bound of $\|\sum_{k=1}^n r_k\alpha_k\alpha_k^*\|$, which is a key estimation used in NPR. The key idea of proof is to apply covering argument to upper bound matrix norm. Similar but more restrictive results can be seen in literature (e.g., \cite{candes2015phase,huang2020performance}), but do not suffice for our purpose. For example, Lemma 2.7 in \cite{huang2020performance} is restricted to i.i.d. Rademacher variables $r_k$ and complex Gaussian $\alpha_k$.

	\begin{lem}
	\label{lemma5}
	Suppose the zero-mean random vector $v = (v_1,...,v_{2d})^{\top}\in \mathbb{R}^{2d}$ has i.i.d. sub-Gaussian entries satisfying $\|v_1\|_{\psi_2} = O(1)$. Assume the random vector $\alpha \in \mathbb{C}^d$ is given by $\alpha_i = v_i+v_{i+d}\ii$ for $i\in[d]$, and $\alpha_1,...,\alpha_n$ are i.i.d. copies of $\alpha$. When $n\geq d$ for any $r = (r_1,...,r_n)^{\top}$ independent of $\alpha_k$, there exist $C_i$ such that
	\begin{equation}
	    \label{2.9}
	    \mathbbm{P}\Big(\Big\|\sum_{k=1}^n \frac{r_k}{\sqrt{n}}\alpha_k\alpha_k^*\Big\|\leq C_1\sqrt{d}\|r\|_\infty + C_2\frac{|\sum_{k=1}^n r_k|}{\sqrt{n}}\Big)\geq 1- 2\exp\big(-C_3d\big)
	\end{equation}
	Moreover, we have the upper bound for the expectation (with regard to $\alpha_k$) as
	\begin{equation}
	    \label{2.10}
	    \mathbbm{E}\Big\|\sum_{k=1}^n \frac{r_k}{\sqrt{n}}\alpha_k\alpha_k^*\Big\| \leq C_4\sqrt{d}\|r\|_\infty + C_5\frac{|\sum_{k=1}^nr_k|}{\sqrt{n}}.
	\end{equation}
	\end{lem}
	
	\noindent{\it Proof.} In this proof we let $H_1 = \sum_{k=1}^n \frac{r_k}{\sqrt{n}}\alpha_k\alpha_k^*$ to lighten the notation. We construct a $\frac{1}{4}$-net $w_1,...,w_{N_1}$ of the unit Euclidean sphere $\mathcal{S}^{d-1}_\mathbb{C}$ of $\mathbb{C}^d$, then by Corollary 4.2.13 in \cite{vershynin2018high} we can assume $N_1 \leq 9^{2d}$. For any $w \in \mathcal{S}_\mathbb{C}^{d-1}$, there exists $i_0$ such that $\|w - w_{i_0}\|\leq \frac{1}{4}$, and hence 
	\begin{equation}
	    \nonumber
	    w^*H_1w = (w-w_{i_0})^*H_1w + w_{i_0}^*H_1(w-w_{i_0})+w_{i_0}^*H_1w_{i_0} \leq \frac{1}{2}\|H_1\|+ \sup_{1\leq i\leq N_1}w_i^*H_1w_i .
	\end{equation}
	Take supremum of $w$ over $\mathcal{S}^{d-1}_\mathbb{C}$, we obtain \begin{equation}
	\label{612.1}
	    \|H_1\| \leq 2\sup_{1\leq i \leq N_1}w_i^*H_1w_i=2\sup_{i\in[N_1]} \sum_{k=1}^n \frac{r_k}{\sqrt{n}} |\alpha_k^*w_i|^2.
	\end{equation}
    We let $\alpha_k^\mathcal{R} = \beta_k$, $\alpha_k^\mathcal{I} =  \gamma_k$ and estimate the sub-exponential norm of $\frac{r_k}{\sqrt{n}}|\alpha_k^*w_i|^2$:
    \begin{equation}
        \label{613add}
        \begin{aligned}
        & \left\lVert\frac{r_k}{\sqrt{n}}|\alpha_k^*w_i|^2\right\rVert_{\psi_1} \leq  \frac{|r_k|}{\sqrt{n}}\left( \left\lVert \big(\beta_k^{\top}w_i^\mathcal{R}+\gamma_k^{\top} w_i^\mathcal{I}\big)^2 \right\rVert_{\psi_1}+\left\lVert \big(\beta_k^{\top}w_i^\mathcal{I} - \gamma_k^{\top}w_i^\mathcal{R}\big)^2 \right\rVert_{\psi_1}\right)\\  &\stackrel{(\ref{2.6})}{\leq }\frac{|r_k|}{\sqrt{n}} \left(\left\lVert \beta_k^{\top}w_i^\mathcal{R}+\gamma_k^{\top} w_i^\mathcal{I} \right\rVert_{\psi_2}^2+\left\lVert \beta_k^{\top}w_i^\mathcal{I} - \gamma_k^{\top}w_i^\mathcal{R} \right\rVert_{\psi_2}^2\right) \stackrel{\mathrm{Prop.} \ref{pro2}}{\leq }  C_1\frac{|r_k|}{\sqrt{n}}.
        \end{aligned}
    \end{equation}
   Note that $\mathbbm{E} \sum_{k=1}^n \frac{r_k}{\sqrt{n}}|\alpha_k^*w_i|^2 = \frac{2\|\beta_1\|_{L_2}^2}{\sqrt{n}}\sum_{k=1}^n r_k$, so we can use Theorem 2.8.1 in \cite{vershynin2018high} for a specific $v_i$ and then take the union bound of $1\leq i \leq N_1$, for any $t>0$ it yields 
   \begin{equation}
       \nonumber
       \begin{aligned}
           \mathbbm{P} \Big(\sup_{  i\in [ N_1]}\Big| \sum_{k=1}^n \frac{r_k|\alpha_k^*v_i|^2}{\sqrt{n}} - \frac{2\|\beta_1\|_{L_2}^2}{\sqrt{n}}\big(\sum_{k=1}^n r_k\big)\Big| \geq t\Big)  \leq 2\exp \Big((2\log 9)d-C_2\min\Big\{\frac{nt^2}{\|r\|^2  }, \frac{\sqrt{n}t}{\|r\|_\infty  }\Big\}\Big)
       \end{aligned}
   \end{equation}
    We let $4\|\beta_1\|_{L_2}^2 = C_3$, some basic algebra gives 
    \begin{equation}
        \nonumber
        \begin{aligned}
            \sup_{ i\in [ N_1]}\Big| \sum_{k=1}^n \frac{r_k|\alpha_k^*v_i|^2}{\sqrt{n}} - \frac{2\| \beta_1\|_{L_2}^2}{\sqrt{n}}\big(\sum_{k=1}^n r_k\big)\Big| & \geq \sup_{i\in [ N_1]}  \Big|\sum_{k=1}^n \frac{r_k}{\sqrt{n}}|\alpha_k^*v_i|^2\Big| -\frac{1}{2}C_3\Big|\frac{\sum_{k=1}^n r_k}{\sqrt{n}}\Big|  \\ & \stackrel{(\ref{612.1})}{\geq } \frac{1}{2}\|H_1\| -  \frac{1}{2}C_3\Big|\frac{\sum_{k=1}^n r_k}{\sqrt{n}}\Big|.
        \end{aligned}
    \end{equation}
    Thus, for any $t>0$ we have the probability inequality 
    \begin{equation}
        \begin{aligned}
            \label{612.2}
            \mathbbm{P}\Big(\|H_1\| \geq t + C_3\Big|\frac{\sum_{k=1}^n r_k}{\sqrt{n}}\Big|\Big)\leq 2\exp \Big((2\log 9)d-C_2\min\Big\{\frac{nt^2}{\|r\|^2 }, \frac{\sqrt{n}t}{\|r\|_\infty }\Big\}\Big)
        \end{aligned}
    \end{equation}
    Set $t =u \sqrt{d}\|r\|_\infty$ with $u\geq 1$, let $\xi_0 = C_3\big|\frac{\sum_{k=1}^n r_k}{\sqrt{n}}\big|$, since $n\geq d$, $n\|r\|_\infty^2 \geq \|r\|^2$, (\ref{612.2}) implies 
    \begin{equation}
        \label{612.3}
        \mathbbm{P}\big(\|H_1\| \geq \sqrt{d}\|r\|_\infty u + \xi_0\big) \leq 2\exp \big([2\log 9 -C_2u]d\big),~~\forall u\geq 1.
    \end{equation}
    Evidently, (\ref{2.9}) follows by choosing sufficiently large $u$ in (\ref{612.3}), for instance, $u = u_{0}: = 4C_2^{-1}\log 9$. Moreover, we can let $u = u_0 + b,~b>0$ in (\ref{612.3}), it yields \begin{equation}
        \label{612.4}
        \mathbbm{P}\big(\|H_1\| \geq  (\sqrt{d}\|r\|_\infty)b+ \xi_1\big) \leq   \exp(-C_2b),~~\forall ~b>0,
    \end{equation}
    where we define $\xi_1 : = u_0\sqrt{d}\|r\|_\infty + \xi_0 $. Now by a standard integral trick, we have 
    \begin{equation}
        \begin{aligned}
            \nonumber
           & \mathbbm{E}\|H_1\| = \int_{0}^\infty \mathbbm{P}\big(\|H_1\| \geq t\big)~\mathrm{dt} \leq \xi_1 +  \int_{\xi_1}^\infty \mathbbm{P}\big(\|H_1\| \geq t\big)~\mathrm{d}t \\ 
           &~\leq ~ \xi_1 + \Big( \int_{0}^\infty \mathbbm{P}\big(\|H_1\| \geq \sqrt{d}\|r\|_\infty b + \xi_1\big)~\mathrm{d}b\Big) \|r\|_\infty\sqrt{d}~~~~~~~~~~(\text{let}~ t =  \sqrt{d}\|r\|_\infty b + \xi_1) \\
           &\stackrel{(\ref{612.4})}{\leq }  \xi_1 + \Big(\int_{0}^\infty \exp(-C_2b)~\mathrm{d}b\Big)\|r\|_\infty\sqrt{d}.
        \end{aligned}
    \end{equation}
    Note that the integral is finite, so we finish the proof. \hfill $\square$

	In NGPR, we will use the covering number of $\mathcal{H}_{d,r}^s$ to establish a relaxed version of restricted isometry property.
	   
	\begin{lem}
	\label{lemmaadd}
	{\rm (Covering number for $\mathcal{H}_{d,r}^s$)} For any $\delta > 0$, there exists a $\delta$-net for $\mathcal{H}_{d,r}^s$ with no more than $\big(\frac{12}{\delta}+1\big)^{r(2d+1)}$ elements, i.e., we can find $\{B_1,...,B_{N_1}\}\subset \mathcal{H}_{d,r}^s$ with $|N_1|\leq \big(\frac{12}{\delta}+1\big)^{r(2d+1)}$, such that for any $B\in \mathcal{H}_{d,r}^s$ there exists $1\leq i_0 \leq N_1$ satisfying $\| B_{i_0}-B\|_F \leq \delta$.
	\end{lem}
	\noindent{\it Proof.} The proof follows the idea of Lemma 3.1 in \cite{candes2011tight}. Under the norm $\|.\|_F$, let $\mathcal{N}_1 = \{\Sigma_{(1)},...,\Sigma_{(K_1)}\}$ be a $\frac{\delta}{3}$-net of the set of real diagonal matrix with noralized Frobenius norm, i.e., $\mathcal{S}_1 = \{\Sigma = [\sigma_{ij}]\in\mathbb{R}^{d\times d}:\sigma_{ij} = 0,i\neq j;\sum_i \sigma_{ii}^2 = 1\}$. Under $\ell_2$ norm, let $\mathcal{N}_2 = \{u_{(1)},...,u_{(K_2)}\}$ be a $\frac{\delta}{3}$-net of $\{u\in \mathbb{C}^d:\|u\|=1\}$, and consider $\mathcal{N}_3 = \{(u_1,...,u_r) \in \mathbb{C}^{d\times r}:u_i \in \mathcal{N}_2,~\forall i\in [r]\}$. Then by Corollary 4.2.13 in \cite{vershynin2018high}, we can assume $|\mathcal{N}_1| \leq \big(\frac{6}{\delta}+ 1\big)^{r}$, $|\mathcal{N}_2| \leq \big(\frac{6}{\delta}+1\big)^{2d}$, and hence $|\mathcal{N}_3|\leq \big(\frac{6}{\delta}+1\big)^{2dr}$.

	Define $\mathcal{S}_2 = \{U \in \mathbb{C}^{d\times r}: U^*U = I_r\}$, and consider $\mathbb{C}^{d\times r}$ with norm $\|U\|_{2,\infty} =  \max_{j\in [r]} \|u_j\|$ where $u_j$'s are columns of $U$. Evidently, $\mathcal{N}_3$ is an "external $\frac{\delta}{3}$-covering" of $\mathcal{S}_2$. By the relation between "external covering number" and "internal covering number"\footnote{\url{https://terrytao.wordpress.com/2014/03/19/metric-entropy-analogues-of-sum-set-theory/}}, there exists $\mathcal{N}_4 \subset \mathcal{S}_2$ is $\frac{\delta}{3}$-net of $\mathcal{S}_2$, and $|\mathcal{N}_4| \leq \big(\frac{12}{\delta}+1\big)^{2dr}$. We construct a set  $$\mathcal{N}_5 = \{U\Sigma U^*: \Sigma \in \mathcal{N}_{1},U\in \mathcal{N}_4\}\subset \mathcal{H}_{d,r}^s.$$ It remains to verify that for any $B\in \mathcal{H}_{d,r}^s$, there exists $\hat{B} \in \mathcal{N}_5$ such that $\|B - \hat{B}\|_F \leq \delta$.

	Assume $B\in\mathcal{H}_{d,r}^s$ admits the compact SVD $B = U_0\Sigma_0U_0^*$, where $U_0  \in \mathcal{S}_2$, $\Sigma_0 \in \mathcal{S}_1$. Thus, there exists $\hat{B} = \hat{U}\hat{\Sigma}\hat{U}^* \subset \mathcal{N}_5$ with $\hat{U}\subset \mathcal{N}_4$ satisfying $\|\hat{U}-U_0 \|_{2,\infty} \leq \frac{\delta}{3}$, and $\hat{\Sigma}\in\mathcal{N}_1$ satisfying $\|\hat{\Sigma} - \Sigma_0 \|_F \leq \frac{\delta}{3}$. Now we estimate the approximation error $\|\hat{B} - B\|_F$ as
	\begin{equation}
	    \begin{aligned}
	        \nonumber
	        &\| \hat{B} - B\|_F  = \|\hat{U}\hat{\Sigma}\hat{U}^* - U_0\Sigma_0U_0^* \|_F \\\leq  & \| \hat{U} (\hat{\Sigma}-\Sigma_0)\hat{U}^*\|_F + \| (\hat{U}-U_0)\Sigma_0\hat{U}^*\|_F +\|U_0\Sigma_0(\hat{U} - U_0)^*\|_F \\
	         = &  \|\hat{\Sigma} - \Sigma_0\|_F + 2\|(\hat{U}-U_0)\Sigma_0\|_F \leq \frac{\delta }{3}+ 2 \|\Sigma_0\|_F \| \hat{U}-U_0\|_{2,\infty}\leq \delta.
	    \end{aligned}
	\end{equation}
	Note that $|\mathcal{N}_5|\leq |\mathcal{N}_1\|\mathcal{N}_4|\leq \big(\frac{12}{\delta}+1\big)^{r(2d+1)}$, the result follows.
	\hfill $\square$
	
	\vspace{1.5mm}
	
	Based on a well-known bound for sub-Gaussian matrix, it is not hard to control the operator norm of an Hermitian matrix constructed by the way in NGPR. We present it as a Lemma.

    \begin{lem}
    \label{lemma4}
    Suppose the zero-mean random matrix $B\in \mathbb{R}^{d\times d}$ satisfies $\| \mathrm{vec}(B)\|_{\psi_2} = O(1)$. Let $B=[b_{ij}]$, we construct Hermitian matrix $A= [a_{ij}] $ by letting $a_{ii} = b_{ii}$, $a_{ij}= \frac{1}{\sqrt{2}}(b_{ij} + b_{ji}\ii)$ when $i<j$, and $a_{ij} = \frac{1}{\sqrt{2}}(b_{ij}-b_{ji}\ii)$ when $i>j$. Then $\mathbbm{P} \big( \|A\| \leq C_1 \big(\sqrt{d}+t \big)\big) \geq 1-4\exp (-t^2)$ for some $C_1$.

    \end{lem}
    \noindent{\it Proof.} We let $A^{(1)} = [a^{(1)}_{ij}]$ be the real upper triangular matrix given by $a^{(1)}_{ii} = \frac{1}{2}b_{ii}$, $a^{(1)}_{ij} = \frac{1}{\sqrt{2}}b_{ij}$ when $i<j$, also let $A^{(2)} = [a^{(2)}_{ij}]$ be the real strictly upper triangular matrix defined by $a^{(2)}_{ij} = \frac{1}{\sqrt{2}}b_{ji}$ when $i<j$. Obviously, $A = A^{(1)} + \big[A^{(1)}\big]^{\top} + \ii A^{(2)}- \ii \big[ A^{(2)} \big]^{\top} $. Hence, we obtain $\|A\| \leq 2\|A^{(1)}\| + 2 \| A^{(2)} \|\leq 4 \max\big\{\|A^{(1)}\|,\|A^{(2)}\|\big\}.$ By definition of sub-Gaussian norm of a random vector, it is obvious that $\max\{\|\mathrm{vec}(A^{(1)})\|_{\psi_2},\|\mathrm{vec}(A^{(2)})\|_{\psi_2}\}\leq \| \mathrm{vec}(B)\|_{\psi_2} = O(1)$. Thus, Lemma 5.3 in \cite{rigollet2015high} followed by a union bound gives for some $C_1$ and any $t>0$, with probability at least $1-4\exp(-t^2)$, $\max\{\|A^{(1)}\|,\|A^{(2)}\|\} \leq C_1(\sqrt{d}+t)$ holds. This concludes the proof.       \hfill $\square$

	\vspace{1mm}

    To establish the error bounds for NPR, NGPR under heavy-tailed noise, the upper bound given in Theorem \ref{theorem6}, \ref{theorem1} are insufficient, the next two Lemmas are thus provided to resolve the technical issues. We first define $\mathcal{P}(r,\zeta) = \{\mathcal{K}\subset [n]: |\mathcal{K}| = \zeta \sqrt{rnd}\}$. Here, $\zeta>0$, $r\in [d]$, and $\zeta\sqrt{rnd}$ is a \textcolor{black}{positive integer.} Specifically, we let $\mathcal{P}(\zeta) := \mathcal{P}(1,\zeta)$.

    In NPR, recall the notation $\mathcal{D}(X) = (\big<\alpha_1\alpha_1^*,X\big>,...,\big<\alpha_n\alpha_n^*,X\big>)^\top$. For any $\mathcal{K}\subset [n]$ we define $\mathcal{D}_\mathcal{K}(X)$ be the $|\mathcal{K}|$-dimensional sub-vector of $\mathcal{D}(X)$ constituted by entries in $\mathcal{K}$. Moreover, for $D = (\alpha_1,...,\alpha_n)^*$ we let $D_{\mathcal{K}}\in \mathbb{C}^{|\mathcal{K}|\times d}$ be the sub-matrix of $D$ constituted by rows in $\mathcal{K}$. Note that $\|\mathcal{D}_{\mathcal{K}}(\beta\beta^*)\|_1 = \| D_\mathcal{K}\beta\|^2$ holds.

\begin{lem}
\label{lemma8}
    Assume $\{\alpha_k\alpha_k^*:k\in [n]\}$ are drawn as the setting of NPR under Assumption \ref{assumption2}. Consider a fixed $\zeta$ and the set $\mathcal{P}(\zeta)$ given above. When $n\geq d$, for some $C_i$ with probability at least $1-2\exp(-C_1\sqrt{nd})$ we have 
    $$\sup_{\mathcal{K}\in \mathcal{P}(\zeta)}\sup_{X\in\mathcal{H}_{d,r}^s}\|\mathcal{D}_{\mathcal{K}}(X)\|_1\leq C_2 \sqrt{rnd}\log \big(\frac{n}{d}\big).$$
\end{lem}

\noindent{\it Proof.} Any $X\in \mathcal{H}_{d,r}^s$ can be written as $X = \sum_{l=1}^{r }s_l\beta_l\beta_l^* $ with $\sum_l s_l^2 = 1$, $\|\beta_l\|=1$. Thus, we have $\|\mathcal{D}_{\mathcal{K}}(X)\|_1\leq \sum_{l=1}^r |s_l|\| \mathcal{D}_{\mathcal{K}}(\beta_l\beta_l^*)\|_1 = \sum_{l=1}^r|s_l| \| {D}_{\mathcal{K}}\beta_l\|^2 \leq \sum_{l=1}^r|s_l|\| {D}_{\mathcal{K}}\|^2\leq \sqrt{r}\| {D}_{\mathcal{K}}\|^2$. This gives \begin{equation}
    \label{620}
    \sup_{\mathcal{K}\in \mathcal{P}(\zeta)}\sup_{X\in\mathcal{H}_{d,r}^s}\|\mathcal{D}_{\mathcal{K}}(X)\|_1 \leq\sqrt{r}\sup_{\mathcal{K}\in \mathcal{P}(\zeta)} \| D_{\mathcal{K}}\|^2.
\end{equation}
For a fixed $\mathcal{K}\in\mathcal{P}(\zeta)$, since $n\geq d$ and $|\mathcal{K}| = \zeta\sqrt{nd}$, by Theorem 4.4.5 in \cite{vershynin2018high}, for any $t>0$ and some $C_1$ we have $\mathbbm{P}\big(\|D_\mathcal{K}\|\geq C_1((nd)^{1/4}+t)\big)\leq 2\exp(-t^2).$ Note that $|\mathcal{P}(\zeta)|= \binom{n}{\zeta\sqrt{nd}}\leq \big(\frac{3}{\zeta}\sqrt{\frac{n}{d}}\big)^{\zeta\sqrt{nd}}$, so a union bound gives for any $t>0$, 
$$ \mathbbm{P}\big(\sup_{\mathcal{K}\in\mathcal{P}(\zeta)}\| D_\mathcal{K}\|\geq C_1((nd)^{\frac{1}{4}}+t)\big)\leq 2\exp\big(-t^2+\zeta \sqrt{nd}\log\big(\frac{3}{\zeta}\sqrt{\frac{n}{d}}\big)\big).$$
Setting $t = C_2(nd)^{1/4}\sqrt{\log \big(\frac{n}{d}\big)}$ for some sufficiently large $C_2$ delivers that with probability at least $1-2\exp(-C_3\sqrt{nd})$, $\sup_{\mathcal{K}\in \mathcal{P}(\zeta)}\|D_\mathcal{K}\| \leq C_3(nd)^{1/4}\sqrt{\log \big(\frac{n}{d}\big)}$ holds. Putting it into (\ref{620}), the result follows. \hfill $\square$

\vspace{1mm}

    In NGPR, recall the notation $\mathcal{A}(X) = (\big<A_1,X\big>,...,\big<A_n,X\big>)^\top$ for any $X\in\mathbb{C}^{d\times d}$. Given $\mathcal{K} \subset [n]$, we define $\mathcal{A}_{\mathcal{K}}(X)\in \mathbb{C}^{|\mathcal{K}|}$ to be the sub-vector of $\mathcal{A}(X)$ constituted by entries in $\mathcal{K}.$
     \begin{lem}
     \label{lemma7}
      Assume $\{A_k:k\in [n]\}$ are drawn as the setting of NGPR under Assumption \ref{assumption1}. Consider  fixed $\zeta >0$, $r\in [d]$ and the set $\mathcal{P}(r,\zeta)$ given above. When $ n\geq C_1r d$ for some sufficiently large $C_1$,  with probability at least $1-\exp(-C_3\sqrt{nd})$ we have 
      $$\sup_{\mathcal{K} \in \mathcal{P}(r,\zeta)}\sup_{X\in\mathcal{H}_{d,r}^s} \|\mathcal{A}_{\mathcal{K}}(X)\|\leq C_2 (rnd)^{\frac{1}{4}}\sqrt{ \log \big(\frac{n}{rd}\big)}.$$
     \end{lem}

\noindent{\it Proof.} We follow the beginning of the proof of Theorem \ref{theorem1}. With regard to $\|.\|_F$, we first take a $\frac{1}{4}$-net of $\mathcal{H}_{d,r}^s$, denoted by $\mathcal{N} = \{X_1,...,X_{N_1}\}\subset \mathcal{H}_{d,r}^s$ with $N_1\leq 52^{3rd}$ by Lemma \ref{lemmaadd}. Then for specific $X \in \mathcal{H}_{d,r}^s$ and $\mathcal{K}\in  \mathcal{P}(r,\zeta)$, for any $t>0$ we have 
	\begin{equation}
    \nonumber
    \mathbbm{P}\Big( \big|\frac{1}{\zeta\sqrt{rnd}}\|\mathcal{A}_{\mathcal{K}}(X)\|^2 - \mathbbm{E}\big<\breve{A},X\big>^2 \big|\geq t\Big) \leq 2\exp\big(-C_2\sqrt{rnd}\min\{t^2,t\}\big).
\end{equation}
As already shown in the Proof of Theorem \ref{theorem1}, $\mathbbm{E}\big<\breve{A},X\big>^2\in [\lambda_0,\lambda_1]$. Thus, let us further note that $|\mathcal{P}(r,\zeta)|= \binom{n}{\zeta\sqrt{rnd}}\leq \big(\frac{3}{\zeta}\sqrt{\frac{n}{rd}}\big)^{\zeta\sqrt{rnd}}$ and take union bound over $X \in \mathcal{N}$, $\mathcal{K}\in \mathcal{P}(r,\zeta)$, for any $t>0$ it yields 
\begin{equation}
    \begin{aligned}
        \nonumber
        \mathbbm{P}\Big(\sup_{\mathcal{K}\in\mathcal{P}(\zeta)}\sup_{i\in [N_1]}\big|&\frac{1}{\zeta\sqrt{rnd}} \|\mathcal{A}_{\mathcal{K}}(X_i)\|^2 -\mathbbm{E}\big<\breve{A},X_i\big>^2 \big| \geq t\Big) \\ 
        &\leq 2\exp\Big(-C_2\sqrt{rnd}\min\{t^2,t\}+3\log(52)rd + \zeta\sqrt{rnd}\log \big(\frac{3}{\zeta}\sqrt{\frac{n}{rd}}\big)\Big). 
    \end{aligned}
\end{equation}
	  When $n\geq  C_1rd$ with sufficiently large $C_1$, we have $\sqrt{rnd}= \Omega (rd)$, hence we set $t = C_3 \log \big(\frac{n}{rd}\big)$ with sufficiently large $C_3$ and obtain with probability at least $1-2\exp(-C_4\sqrt{nd})$ we have $$ \sup_{\mathcal{K}\in\mathcal{P}(r,\zeta)}\sup_{i\in [N_1]}\big| \frac{1}{\zeta\sqrt{rnd}} \|\mathcal{A}_{\mathcal{K}}(X_i)\|^2 -\mathbbm{E}\big<\breve{A},X_i\big>^2 \big| = O\Big(\log \big(\frac{n}{rd}\big)\Big).$$
	  Since $\mathbbm{E}\big<\breve{A},X_i\big>^2 = O(1)$, it further gives 
	  $ \sup_{\mathcal{K}\in\mathcal{P}(r,\zeta)}\sup_{i\in [N_1]}\|\mathcal{A}_{\mathcal{K}}(X_i)\|  = O\Big((rnd)^{\frac{1}{4}}\sqrt{\log \big(\frac{n}{rd}\big)}\Big) . $
	  Since $\mathcal{H}_{d,r}^s$ is compact, we can find $\mathcal{K}_0\in \mathcal{P}(r,\zeta)$, $X_0\in \mathcal{H}_{d,r}^s $ such that 
	  $ \|\mathcal{A}_{\mathcal{K}_0}(X_0)\|=\sup_{\mathcal{K} \in \mathcal{P}(r,\zeta)}$ $\sup_{X\in\mathcal{H}_{d,r}^s} \|\mathcal{A}_{\mathcal{K}}(X)\|. $
	  There exists $i_0\in [N_1]$ such that $\|X_0 - X_{i_0}\|_F\leq \frac{1}{4}$. Noting that one can write $\frac{X_0-X_{i_0}}{\|X_0-X_{i_0}\|}= X^{(1)}+X^{(2)}$ with $X^{(1)},X^{(2)}\in \mathcal{H}_{d,r}$ satisfying $\|X^{(1)}\|_F,\|X^{(2)}\|_F\leq 1$, which gives 
	  \begin{equation}
	      \begin{aligned}
	          \nonumber
	          &\|\mathcal{A}_{\mathcal{K}_0}(X_0)\|\leq \|\mathcal{A}_{\mathcal{K}_0}(X_0-X_{i_0})\|+\|\mathcal{A}_{\mathcal{K}_0}(X_{i_0})\|  \leq \sup_{\mathcal{K}\in\mathcal{P}(\zeta)}\sup_{i\in [N_1]}\|\mathcal{A}_{\mathcal{K}}(X_i)\|\\ & + \frac{1}{4}{\big(\|\mathcal{A}_{\mathcal{K}_0}(X^{(1)})\|+\|\mathcal{A}_{\mathcal{K}_0}(X^{(2)})\| } \big) \leq  \sup_{\mathcal{K}\in\mathcal{P}(\zeta)}\sup_{i\in [N_1]}\|\mathcal{A}_{\mathcal{K}}(X_i)\| + \frac{1}{2}\|\mathcal{A}_{\mathcal{K}_0}(X_0)\|.
	      \end{aligned}
	  \end{equation}
Thus, we obtain $\|\mathcal{A}_{\mathcal{K}_0}(X_0)\| \leq 2\sup_{\mathcal{K}\in\mathcal{P}(r,\zeta)}\sup_{i\in [N_1]}\|\mathcal{A}_{\mathcal{K}}(X_i)\| = O \big((rnd)^{\frac{1}{4}}\sqrt{\log \big(\frac{n}{rd}\big)}\big)$. \hfill $\square$  
 
 \section{Proof Sketch for Matrix Sensing}
\label{appenc}

Alternatively, In the proof sketch we may use $T_1 \lesssim T_2$ to denote $T_1 = O(T_2)$, $T_1 \gtrsim T_2$ for $T_1 = \Omega(T_2)$, and $T_1 \asymp T_2$ to represent $T_1 = \Theta(T_2)$.

\vspace{0.3mm}

\noindent
\textbf{A Proof Sketch of Theorem \ref{theorem10}.} \textbf{(a).} The optimality gives $\|Y - \mathcal{D}(\widehat{U}\widehat{U}^*)\| \leq \| Y -\mathcal{D}(U_0U_0^*)\|$. Use $Y = \mathcal{D}(U_0U_0^*)+\eta$, we have $$ \|\mathcal{D}(U_0U_0^*-\widehat{U}\widehat{U}^*) +\eta\|\leq \|\eta\|\Longrightarrow \|\mathcal{D}(\widehat{X} - X_0 )\| \lesssim \|\eta\|.$$
Since when $n = \Omega(rd)$ we have \begin{equation}
    \label{C.1}
   \begin{aligned}
       & \|\mathcal{D}(\widehat{X} - X_0 )\|  \geq \|\widehat{X}-X_0\|_F \inf_{X\in\mathcal{H}_{d,2r}^s}\|\mathcal{D}(X)\|\\ &\geq   \|\widehat{X}-X_0\|_F\inf_{X\in\mathcal{H}_{d,2r}^s}\frac{\|\mathcal{D}(X)\|_1}{\sqrt{n}}
   \stackrel{\mathrm{Thm.}  \ref{theorem6}}{\gtrsim} \sqrt{n}\|\widehat{X}-X_0\|_F ,
   \end{aligned}
\end{equation}
which gives $\|\widehat{X}-X_0\|= O\big(\frac{\|\eta\|}{\sqrt{n}}\big)$. 

\noindent
\textbf{(b).} We start from $\sum_{k=1}^n(y_k - \big<\alpha_k\alpha_k^*,\widehat{U}\widehat{U}^*\big>)^2\leq \sum_{k=1}^n(y_k - \big<\alpha_k\alpha_k^*,{U}_0{U_0}^*\big>)^2$. Plug in $y_k = \big<\alpha_k\alpha_k^*,{U}_0{U_0}^*\big> +\eta_k$, some algebra gives 
\begin{equation}
    \label{C.2}
    \|\mathcal{D}(\widehat{X}-X_0)\|^2 \lesssim \big<\sum_{k=1}^n \eta_k\alpha_k\alpha_k^*,\widehat{X}-X_0\big> .
\end{equation}
 For right hand side of (\ref{C.2}), 
\begin{equation}
    \label{C.3}
    \begin{aligned}
    &\big<\sum_{k=1}^n \eta_k\alpha_k\alpha_k^*,\widehat{X}-X_0\big> \leq \|\widehat{X}-X_0\|_F \sup_{X\in\mathcal{H}_{d,2r}^s}\big<\sum_{k=1}^n\eta_k\alpha_k\alpha_k^*, X \big>\\
    \leq &\|\widehat{X}-X_0\|_F   \Big\|\sum_{k=1}^n\eta_k\alpha_k\alpha_k^*\Big\| \sup_{X\in\mathcal{H}_{d,2r}^s} \|X\|_{nu}  \stackrel{\scriptscriptstyle\mathrm{Lem.}\ref{lemma5},(\ref{nunormbound})}{\lesssim }
   \sqrt{r} \|\widehat{X}-X_0\|_F \big(\|\eta\|_\infty\sqrt{nd}+|\mathbf{1}^\top\eta|\big).
    \end{aligned}
\end{equation}
Putting (\ref{C.1}), (\ref{C.3}) into (\ref{C.2}) concludes the proof.

\noindent
    \textbf{(c) and (d).} Based on the obtained bound $O\big(\|\eta\|_\infty\sqrt{\frac{rd}{n}}+\frac{\sqrt{r}|\mathbf{1}^\top\eta|}{n}\big)$, the proofs are exactly the same as Theorem \ref{theorem7}.

\noindent
    \textbf{(e).} $\sum_{k=1}^n\big( \widetilde{y}_k - \big<\alpha_k\alpha_k^*,\widehat{X}_{ht}\big>\big)^2\leq \sum_{k=1}^n \big(\widetilde{y}_k - \big<\alpha_k\alpha_k^*,{X_0}\big> \big)^2$ and $y_k = \big<\alpha_k\alpha_k^*,X_0\big>+\eta_k$ give
    \begin{equation}
        \label{C.7}
        \big\|\mathcal{D}(X_0-\widehat{X}_{ht})\big\|^2 \lesssim \big(\sum_{k\in\mathcal{I}(\tau)} + \sum_{k\in\mathcal{I}(\tau)^c}\big)\big(\widetilde{y}_k-y_k+\eta_k\big)\big<\alpha_k\alpha_k^*, \widehat{X}_{ht}-X_0\big>.
    \end{equation}
    Here, $\mathcal{I}(\tau) = \{k\in[n]:|\eta_k| \geq \frac{\tau}{2}\}$, $\mathcal{I}(\tau)^c = [n]\setminus \mathcal{I}(\tau)$. A same argument in the proof of Theorem \ref{theoremht} delivers $| \mathcal{I}(\tau)|\leq \zeta_0\sqrt{ {nd}} $ with high probability, where $\zeta_0>0$, $\zeta_0\sqrt{nd}$ is positive integer. For left hand side of (\ref{C.7}), same as (\ref{C.1}), we have $\|\mathcal{D}(X_0-\widehat{X}_{ht})\|^2\gtrsim n \|X_0-\widehat{X}_{ht} \|_F^2$.

    We assume $X_0 = \sum_{l=1}^r s_l\beta_l\beta_l^*$ with $\sum_l |s_l| = O(1)$, $\|\beta_l\|=1$, and for a fixed $X_0$, $\beta_l,l\in[r]$ are also fixed. Since $\big\| \alpha_k^*\beta_l \big\|_{\psi_1} = O(1)$ with high probability 
    \begin{equation}
        \nonumber
        \|\mathcal{D}(X_0)\|_\infty \leq \sum_{l=1}^r |s_l| \big\|\mathcal{D}(\beta_l\beta_l^*)\big\|_\infty \lesssim \max_{l\in[r]}\max_{k\in[n]} |\alpha_k^*\beta_l|^2 \stackrel{ \mathrm{Prop.}\ref{pro4}}{\lesssim } \log n 
        \end{equation}
    On this event, choosing $C_1$ sufficiently large guarantees $ \frac{1}{2}\tau \geq  \|\mathcal{D}(X_0)\|_\infty$. Now, for the first summation in right hand side of (\ref{C.7}), with high probability we have  
    \begin{equation}
        \label{C.8}
        \begin{aligned}
       & \sum_{k\in\mathcal{I}(\tau)} \big(\widetilde{y}_k-y_k+\eta_k\big)\big<\alpha_k\alpha_k^*, \widehat{X}_{ht}-X_0\big> \lesssim \tau \|\widehat{X}_{ht}-X_0\|_F \sup_{X\in\mathcal{H}_{d,2r}^s}\|\mathcal{D}_{\mathcal{I}(\tau)}(X)\|_1 \\
        &\leq  \tau  \|\widehat{X}_{ht}-X_0\|_F \sup_{\mathcal{K}\in\mathcal{P}(\zeta_0)}\sup_{X\in\mathcal{H}_{d,2r}^s}\|\mathcal{D}_{\mathcal{K} }(X)\|_1 \stackrel{ \mathrm{Lem.}\ref{lemma8}}{\lesssim } \tau \sqrt{rnd}\log\big(\frac{n}{d}\big) \|\widehat{X}_{ht}-X_0\|_F .
        \end{aligned}
    \end{equation}
    For the second summation, the treatment is almost the same as that of Theorem \ref{theoremht},
    \begin{equation}
        \label{C.9}
        \begin{aligned}
              &\sum_{k\in\mathcal{I}(\tau)^c} \big(\widetilde{y}_k-y_k+\eta_k\big)\big<\alpha_k\alpha_k^*, \widehat{X}_{ht}-X_0\big> = \big< \sum_{k=1}^n \eta_k\mathbbm{1}(|\eta_k|\leq  \frac{\tau}{2})\alpha_k\alpha_k^*,\widehat{X}_{ht}-X_0\big> \\ &\lesssim \| \widehat{X}_{ht}-X_0\|_F\Big\|\sum_{k=1}^n \eta_k\mathbbm{1}(|\eta_k|\leq  \frac{\tau}{2})\alpha_k\alpha_k^*\Big\|\sup_{X\in\mathcal{H}_{d,2r}^s}\|X\|_{nu} \lesssim \tau \sqrt{rnd}  \|\widehat{X}_{ht}-X_0\|_F .
        \end{aligned}
    \end{equation}
By putting things together we conclude the proof. \hfill $\square$

\vspace{1mm}

\noindent
\textbf{A Proof Sketch of Theorem \ref{theorem9}.} \textbf{(a).} We combine $\sum_{k=1}^n (y_k - \big<A_k,\widehat{U}\widehat{U}^*\big>)^2 \leq \sum_{k=1}^n (y_k - \big<A_k,U_0U_0^*\big>)^2$ and $y_k = \big<A_k,U_0U_0^*\big>+\eta_k$ and obtain 
\begin{equation}
    \label{C.4}
    \sum_{k=1}^n\big<A_k,\widehat{X}-X_0\big>^2 \lesssim \big<\sum_{k=1}^n\eta_kA_k,\widehat{X}-X_0\big>.
\end{equation}
For the left hand side when $n = \Omega(rd)$ we have
\begin{equation}
    \label{C.5}
    \sum_{k=1}^n \big<A_k,\widehat{X}-X_0\big>^2 \geq \|\widehat{X}-X_0\|_F^2 \inf_{X\in\mathcal{H}_{d,2r}^s}\|\mathcal{A}(X)\|^2  \stackrel{\mathrm{Thm.}\ref{theorem1}}{\gtrsim} n  \|\widehat{X}-X_0\|_F^2.
\end{equation}
For the right hand side, it delivers that 
\begin{equation}
    \label{C.6}
    \big<\sum_{k=1}^n\eta_kA_k,\widehat{X}-X_0\big> \leq \|\widehat{X}-X_0\|_F\big\|\sum_{k=1}^n\eta_kA_k\big\| \sup_{X\in \mathcal{H}_{d,2r}^s}\|X\|_{nu} \stackrel{\scriptscriptstyle \mathrm{Lem}.\ref{lemma4},(\ref{nunormbound})}{\lesssim } \|\eta\|\sqrt{rd}\|\widehat{X}-X_0\|_F.
\end{equation}
Putting (\ref{C.5}), (\ref{C.6}) into (\ref{C.4}) gives the desired bound.

\vspace{0.3mm}

\noindent
\textbf{(b), (c) and (d).} Using $O\big(\|\eta\|\frac{\sqrt{rd}}{n}\big)$, the proofs are exactly the same as Theorem \ref{theorem33}.

\vspace{0.3mm}

\noindent
\textbf{(e).} $\sum_{k=1}^n\big( \widetilde{y}_k - \big<A_k,\widehat{X}_{ht}\big>\big)^2\leq \sum_{k=1}^n \big(\widetilde{y}_k - \big<A_k,{X_0}\big> \big)^2$ and $y_k = \big<A_k,X_0\big>+\eta_k$ give
    \begin{equation}
        \label{C.10}
        \big\|\mathcal{A}(X_0-\widehat{X}_{ht})\big\|^2 \lesssim \big(\sum_{k\in\mathcal{I}(\tau)} + \sum_{k\in\mathcal{I}(\tau)^c}\big)\big(\widetilde{y}_k-y_k+\eta_k\big)\big<A_k, \widehat{X}_{ht}-X_0\big>.
    \end{equation}
    Here, $\mathcal{I}(\tau) = \{k\in[n]:|\eta_k| \geq \frac{\tau}{2}\}$, $\mathcal{I}(\tau)^c = [n]\setminus \mathcal{I}(\tau)$. A same argument in the proof of Theorem \ref{theoremht} delivers $| \mathcal{I}(\tau)|\leq \zeta_0\sqrt{rnd} $ with high probability, where $\zeta_0>0$, $\zeta_0\sqrt{rnd}$ is positive integer. For left hand side of (\ref{C.10}), same as (\ref{C.5}), we have $\|\mathcal{A}(X_0-\widehat{X}_{ht})\|^2\gtrsim n \|X_0-\widehat{X}_{ht} \|_F^2$.

    Since $\|X_0\|_F = O(1)$, we have $\|\big<A_k,X_0\big>\|_{\psi_2} = O(1)$. By Proposition \ref{pro4}, $\|\mathcal{A}(X_0)\|_\infty = O(\sqrt{\log n})$ holds with high probability. Thus, we choose $C_1$ sufficiently large such that $\frac{1}{2}\tau \geq  \|\mathcal{A}(X_0)\|_\infty$. On these events, similar to bounding $T_1$ in Theorem \ref{theorem5} we have 
    \begin{equation}
        \label{C.11}
        \begin{aligned}
              &\sum_{k\in\mathcal{I}(\tau)} \big(\widetilde{y}_k-y_k+\eta_k\big)\big<A_k, \widehat{X}_{ht}-X_0\big> \lesssim \big(|\mathcal{I}(\tau)| |\tau|^2\big)^{1/2}\|\widehat{X}_{ht}-X_0\|_F\sup_{X\in\mathcal{H}_{d,2r}^s} \|\mathcal{A}_{\mathcal{I}(\tau)}(X)\| \\
              &\lesssim \tau \sqrt{|\mathcal{I}(\tau)|}\|\widehat{X}_{ht}-X_0\|_F\sup_{\scriptscriptstyle\mathcal{K}\in\mathcal{P}(r,\zeta_0)}  \sup_{\scriptscriptstyle X\in\mathcal{H}_{d,2r}^s} \|\mathcal{A}_{\mathcal{K}}(X)\|\stackrel{\scriptscriptstyle \mathrm{Lem}.\ref{lemma7}}{\lesssim } \tau\sqrt{rnd\log\big(\frac{n}{rd}\big)} \|\widehat{X}_{ht}-X_0\|_F   
        \end{aligned}
    \end{equation}
Similar to bounding $T_2$, for the second summation in right hand side of (\ref{C.10}), we have
\begin{equation}
    \begin{aligned}
          \label{C.12}
         & \sum_{k\in\mathcal{I}(\tau)^c} \big(\widetilde{y}_k-y_k+\eta_k\big)\big<A_k,\widehat{X}_{ht}-X_0\big> =  \big<\sum_{k=1}^n\eta_k\mathbbm{1}(|\eta_k|\leq \frac{\tau}{2})A_k,\widehat{X}_{ht}-X_0\big> \\
         & \stackrel{(\ref{nunormbound})}{\leq}  \sqrt{nr }\tau\|\widehat{X}_{ht}-X_0\|_F \Big\| \sum_{k=1}^n\frac{\eta_k\mathbbm{1}(|\eta_k|\leq {\tau}/{2})}{\sqrt{n }\tau}A_k\Big\|  \stackrel{\mathrm{Lem}. \ref{lemma4}}{\lesssim}\tau\sqrt{rnd}\|\widehat{X}_{ht}-X_0\|_F
    \end{aligned}
\end{equation}
Putting (\ref{C.11}), (\ref{C.12}) into (\ref{C.10}), the result follows. \hfill $\square$

\section{Wirtinger Flow}
\label{append}
We use Wirtinger flow  with spectral initialization to solve ERM in NPR and NGPR. For NPR, we use the code available in Soltanolkotabi's website\footnote{\url{https://viterbi-web.usc.edu/~soltanol/WFcode.html}} with the maximum iteration $2500$ changed to $2000$. One may check the implementation details in their paper \cite{candes2015phase}. For NGPR, we let $S = \frac{1}{n}\sum_{k=1}^n y_kA_k$ and then use 50 power iterations to approximately find its leading eigenvector $\nu_0$. Recall $Y = (y_1,...,y_n)^{\top}$, we use $z_0 = \big(\frac{\|Y\|^2}{n}\big)^{1/4}\frac{\nu_0}{\|\nu_0\|}$ as the initialization. This is the spectral initialization suggested in \cite{huang2020solving}. 
	 Based on the Wirtinger derivative $$\nabla_zf(x) = \big[\frac{\partial}{\partial z}\frac{1}{n}{\sum_{k=1}^n(y_k-z^*A_kz)^2} \big]^* = {\frac{2}{n}\sum_{k=1}^n(z^*A_kz - y_k)A_kz} $$ the update formula is thus given by 
	$$z_{t+1} = z_{t} - s \cdot\frac{2}{n}\sum_{k=1}^n(z_t^*A_kz_t - y_k)A_kz_t.$$
	Following the step size scaling in \cite{huang2020solving} we use $s = 0.01\sqrt{\frac{n}{\|Y\|^2}}$. After $2000$ iterations, the algorithm is stopped and returns $z_{2000}$.

	Due to lack of guarantee of Wirtinger flow for our experimental settings where different noise patterns occur and non-Gaussian measurement is used, we point out that the algorithm above are not theoretically guaranteed to find a global minimizer. 
    Our strategy to address the issue is an additional inspection of the return. It is easy to see that $z_{2000}$ satisfying 
    \begin{equation}
        \begin{cases}
        \label{4.111}
         \sum_{k=1}^n\big(y_k - |\alpha_k^*z_t|^2\big)^2 \leq \sum_{k=1}^n \big(y_k - |\alpha_k^*x_0|^2\big)^2~~~~(\text{in NPR}) \\
     \sum_{k=1}^n \big(y_k - z_t^*A_kz_t\big)^2 \leq \sum_{k=1}^n \big(y_k - x_0^*A_kx_0\big)^2~~~(\text{in NGPR}) 
        \end{cases}
    \end{equation}

\noindent
possesses the same error bounds as the global minimizer of ERM. Therefore, we only count in the simulation whose return satisfies (\ref{4.111}). This is unrealistic for a real-world problem (due to the unknown $x_0$), but suffices for the purpose of our numerical simulations, i.e., to verify the obtained theoretical error bounds. In our simulations, we report that Wirtinger flow is quite stable  and always returned a point satisfying (\ref{4.111}).

\end{document}